\newcommand{\ZZauthor}{Guangchen Lan}

\newcommand{\ZZcampus}{West Lafayette}

\newcommand{\ZZdegree}{Doctor of Philosophy}

\newcommand{\ZZdocument}{A Dissertation}

\newcommand{\ZZgraduation}{May 2026}

\newcommand{\ZZinstitution}{Purdue University}

\newcommand{\ZZprogram}{Electrical and Computer Engineering}

\newcommand{\ZZtitle}{%
  Reinforcement Learning for Scalable and Trustworthy Intelligent Systems
}

\newcommand{\ZZshowcolophon}{false}

\newcommand{\ZZshowdiagonalline}{false}

\newcommand{\ZZshowgridlines}{false}

\newcommand{\ZZshowmarginlines}{false}

\newcommand{\ZZshowtimestamp}{false}

\newcommand{\ZZshowtodonotes}{false}

\InputIfFileExists{optional-debugging-code.tex}{}{}

\documentclass{PurdueThesis}

\makeatletter
\@ifundefinedcolor{PineGreen}{\definecolor{PineGreen}{rgb}{0.0,0.47,0.44}}{}
\@ifundefinedcolor{Violet}{\definecolor{Violet}{rgb}{0.56,0.0,1.0}}{}
\makeatother

\newcommand{\ZZatinformation}{}

\graphicspath{{graphics/}}

\makeatletter
  \def\input@path{{misc}{packages}}
\makeatother

\ConfigureBibliography

\usepackage{bm}

\usepackage{bold-extra}

\usepackage{chemfig}

\usepackage
[
  n,            
  advantage,
  operators,
  sets,
  adversary,
  landau,
  probability,
  notions,
  logic,
  ff,
  mm,
  primitives,
  events,
  complexity,
  oracles,
  asymptotics,
  keys,
]
{cryptocode}

\usepackage{fancyvrb}
  \DefineShortVerb{\|}

\usepackage{filehook}

\usepackage{listings}

\usepackage[most]{tcolorbox}

\usepackage[version=4]{mhchem}
  
\usepackage{cancel}

\usepackage{microtype}
\usepackage{graphicx}
\usepackage{booktabs} 
\usepackage{multirow} 
\usepackage{array}
\usepackage{etoc}
\usepackage{colortbl}
\usepackage{xcolor}
\definecolor{RoyalBlue}{rgb}{0.25, 0.41, 0.88}
\definecolor{Tan}{rgb}{0.95, 0.52, 0.0}

\usepackage{amsmath}
\usepackage{amssymb,bm}
\usepackage{mathtools}
\usepackage{enumitem}

\usepackage{hyperref}
\usepackage{url}
\usepackage{algorithm}
\usepackage{algorithmic}

\usepackage[capitalize,noabbrev]{cleveref}

\usepackage{hyperxmp}
  \hypersetup{
    pdfauthor    = {Mark Senn},
    pdfcopyright = {Copyright \copyright\ 2024 by Mark Senn.  All rights reserved.},
    pdfdate      = {2025-05},
    pdfkeywords  = {LaTeX; Purdue University; PurdueThesis},
    pdflang      = {en},
    pdfpublisher = {Purdue University},
    pdfsubject   = {%
                     PurdueThesis is a LaTeX document class used for
                     master's bypass reports,
                     master's theses,
                     PhD dissertations,
                     and PhD preliminary reports.
                     This template demonstrates how to use PurdueThesis.%
                   },
    pdftitle     = {This is the Title},
  }

\usepackage{latexsym}

\ifdefined\directlua
  \usepackage{luacode}
\fi

\usepackage{makeidx}
  \makeindex

\usepackage{mathtools}

\usepackage{media9}

\usepackage{multicol}

\usepackage{pa-ditto}

\usepackage{pa-figure-dash}

\usepackage{pa-logos}

\usepackage{pdfpages}

\usepackage{pa-mismath}
  \MathUp{e}  \MathUp{i}  \MathUp{j}  \pinumber
\def\pa{\rotatebox[origin=c]{14}{\partial}}

\def\Fourier{\mathcal{F}}

\def\Laplace{\mathcal{L}}

\usepackage{pa-repeat}

\usepackage{placeins}

\usepackage{soul}

\usepackage{textcomp}

\usepackage{tikz}
  \usetikzlibrary{arrows.meta}
  \usepackage{circuitikz}
  \usepackage{menukeys}
  \usetikzlibrary{3d}
  \usetikzlibrary{calc,shadows,shapes.misc,shapes.symbols}
  \usetikzlibrary{decorations.text}
  \usetikzlibrary{decorations.pathmorphing} 
  \usetikzlibrary{fit}
  \usetikzlibrary{backgrounds}	

\usepackage[compat=1.1.0]{tikz-feynman}

\usepackage{xfrac}

\newcommand{\I}[1]{\MyRepeat{\indent}{#1}}

\long\def\MyI#1%
  {%
    {%
      \fontsize{8}{10}\tt
      \VerbatimInput
        [
          firstnumber = 1,
          numbers     = left,
          xleftmargin = 0.33in,
        ]%
        {#1}
    }%
  }

\newcommand{\MyIO}
  {%
    \input{z.out}

    {%
      \fontsize{8}{10}\tt
      \VerbatimInput
        [
          firstnumber = 1,
          numbers     = left,
          xleftmargin = 0.33in,
        ]
        {z.out}
    }
    \FloatBarrier
  }

\usepackage{pa-typographic-conventions}

\usepackage{covington}
\usepackage{slgloss}

\begin{document}

\makeatletter
\renewcommand{\ZZAppendixName}{APPENDIX}
\def\@@makechapterhead#1{\uppercase{\@chapapp~\thechapter. #1}}

\setcounter{tocdepth}{3}

\maketitle

\ProvidesFile{ch-front.tex}[2024-09-12 front matter chapter]
%
%
%
%
%
%

%
%
%
\begin{statement}
  \entry{Dr.~Christopher G. Brinton, Chair}{Elmore Family School of Electrical and Computer Engineering}
  \entry{Dr.~Vaneet Aggarwal}{Edwardson School of Industrial Engineering}
  \entry{Dr.~Abolfazl Hashemi}{Elmore Family School of Electrical and Computer Engineering}
  \entry{Dr.~David I. Inouye}{Elmore Family School of Electrical and Computer Engineering}
  \approvedby{Dr.~Milind Kulkarni}
\end{statement}




\pdfbookmark{TABLE OF CONTENTS}{Contents}
\tableofcontents

\listoftables

\listoffigures

\begin{abstract}
Reinforcement learning has become a powerful paradigm for improving the capability of intelligent systems, but its practical deployment faces two central challenges. First, reinforcement learning must scale efficiently in distributed environments where communication bandwidth is limited and computation is heterogeneous across agents. Second, as reinforcement learning is increasingly used in post-training large language models and autonomous agents, the optimized policies must also be aligned with human preferences and satisfy safety requirements such as privacy-aware information disclosure. This dissertation addresses both challenges through four complementary contributions spanning federated optimization, preference alignment, and contextual safety.

The first part of the dissertation studies scalable reinforcement learning in federated settings. We propose \textit{FedNPG-ADMM}, a communication-efficient framework for synchronous federated reinforcement learning that integrates the alternating direction method of multipliers (ADMM) with natural policy gradient optimization. FedNPG-ADMM reduces the per-iteration communication complexity from $O(d^2)$ to $O(d)$, where $d$ is the number of model parameters, while retaining the stationary convergence guarantees of standard federated natural policy gradient methods. We further propose \textit{AFedPG}, an asynchronous federated reinforcement learning framework that improves efficiency under heterogeneous computing speeds. To address stale updates in asynchronous settings, AFedPG employs a delay-adaptive lookahead technique that accounts for the staleness of received gradients. We establish that AFedPG achieves an $O(\epsilon^{-2.5}/N)$ sample complexity, yielding a linear speedup with respect to the number of agents, and improves the global time complexity from $O(t_{\max}/N)$ to $O((\sum_{i=1}^{N} 1/t_i)^{-1})$, where $t_i$ denotes the computation time of agent $i$. Extensive experiments on MuJoCo benchmarks validate these theoretical findings: FedNPG-ADMM substantially reduces communication overhead while maintaining reward performance comparable to standard FedNPG, and AFedPG consistently outperforms synchronous approaches in convergence speed and scalability across heterogeneous environments.

The second part of the dissertation studies trustworthy reinforcement learning for large language models. We introduce \textit{Maximum a Posteriori Preference Optimization (MaPPO)}, a principled preference optimization framework that incorporates prior reward knowledge into a Maximum a Posteriori objective. Building on the Direct Preference Optimization paradigm, MaPPO generalizes DPO and its variants by moving beyond the oversimplified maximum-likelihood treatment of preference pairs as pure binary classification. MaPPO supports both offline and online settings, requires no additional hyperparameters, and can be used as a plugin for DPO variants such as SimPO, IPO, and CPO. Empirical evaluations across standard alignment benchmarks, including MT-Bench, AlpacaEval 2.0, and Arena-Hard, demonstrate consistent improvements in alignment performance across multiple model families and sizes without sacrificing computational efficiency.

We further study \textit{contextual integrity} in large language models, a safety property concerning what information is appropriate to disclose in a given context. We first show that explicit reasoning about contextual integrity can improve a model's ability to distinguish between information that should and should not be shared. Building on this insight, we develop a reinforcement-learning-based post-training framework that instills contextual-integrity reasoning directly into the model policy. Using a synthetic dataset of approximately 700 automatically generated examples spanning diverse domains and disclosure norms, we show that the proposed method substantially reduces inappropriate information disclosure while maintaining task utility across multiple model families and sizes. Importantly, these gains transfer to the human-annotated \textit{PrivacyLens} benchmark, where the method achieves substantial reductions in privacy leakage.

Together, these contributions advance reinforcement learning along two complementary dimensions. On the one hand, they make reinforcement learning more scalable through communication-efficient and asynchronous federated optimization. On the other hand, they make reinforcement learning more trustworthy by improving alignment with human preferences and by reducing contextually inappropriate information disclosure in language-based intelligent systems. As a whole, this dissertation argues that the next generation of intelligent systems will require both efficient optimization and trustworthy behavior, and that reinforcement learning provides a unifying framework for addressing both goals.
\end{abstract}

%
%

\ProvidesFile{ch-fedrl.tex}[2026-04-07 introduction chapter]

\chapter{Introduction and Thesis Overview}
\label{ch:introduction}

\section{Motivation}

Reinforcement learning (RL) provides a general framework for learning sequential decision-making policies through interaction with an environment \cite{williams1992simple}. Over the past decade, RL has become a central methodology for improving the capability of intelligent systems across a wide range of domains, including transportation, networking, resource allocation, robotics, and large-scale control \cite{al2019deeppool, geng2023reinforcement, chen2023two, gonzalez2023asap,mao2025liquid,zhang2025invertible}. In modern machine learning systems, RL has also played a major role in aligning large language models (LLMs) with human intent, as illustrated by reinforcement learning from human feedback (RLHF) in systems such as InstructGPT, GPT-4, and Gemma \cite{ouyang2022training, openai2023gpt4, Gemma2024}.

Despite its strong empirical success, the practical deployment of RL continues to face two fundamental challenges. The first challenge is \emph{scalability}. Many RL applications operate in distributed settings, require large amounts of online data, and must function under limited communication bandwidth and heterogeneous computation resources \cite{provost2013data, Predd2006, niknam2020federated}. In such settings, naive centralized training may be prohibitively expensive, slow, or even infeasible. The second challenge is \emph{trustworthiness}. As RL is increasingly used to optimize the behavior of human-facing language models and autonomous agents, the learned policies must not only be effective, but also aligned with human preferences and capable of respecting safety-sensitive norms, such as appropriate information disclosure and privacy preservation \cite{christiano2017deep, ouyang2022training, nissenbaum2009privacy}.

These two challenges are often studied separately. On the one hand, a large body of work focuses on improving the efficiency of RL systems through better optimization, parallelism, and distributed learning \cite{kakade2001natural, schulman2017proximal, schulman2017trust, mcmahan2017communication}. On the other hand, recent work in LLM post-training emphasizes preference alignment, reasoning quality, and safety \cite{dpo_neurips2023, mireshghallah2024can, shao2024privacylens}. This dissertation argues that these two directions should be viewed as complementary aspects of a unified research agenda: the next generation of intelligent systems must be both \emph{scalable} and \emph{trustworthy}, and reinforcement learning offers a common framework for addressing both goals.

\section{Reinforcement Learning for Scalable Intelligent Systems}

The first theme of this dissertation is the scalability of RL in distributed environments. Classical RL algorithms often assume that data can be collected and processed centrally. In many practical scenarios, however, data is generated by multiple distributed agents, each interacting with its own local environment. Transferring raw trajectories to a central server may introduce severe communication overhead, latency, privacy risks, and systems bottlenecks \cite{Predd2006, niknam2020federated, lan2023fl}. Federated learning (FL) provides a natural alternative by allowing agents to share model updates rather than raw data \cite{mcmahan2017communication}. Extending this idea to RL yields \emph{federated reinforcement learning} (FedRL), in which multiple agents collaboratively learn a common policy while keeping local experience decentralized.

FedRL is attractive because it preserves the parallelism inherent in distributed data collection while reducing the need for raw data sharing. At the same time, it introduces new optimization and systems challenges. Policy-gradient methods with strong convergence properties, such as natural policy gradient (NPG) and trust-region methods, are often second-order or curvature-aware, which makes their communication cost high in federated settings \cite{kakade2001natural, schulman2017trust}. While first-order methods such as policy gradient and Proximal Policy Optimization (PPO) are often favored for their simplicity \cite{schulman2017proximal}, second-order methods are known to exhibit superior convergence behavior in challenging continuous-control problems. In addition, conventional synchronous federated optimization suffers from the \emph{straggler problem}: the overall speed of learning is limited by the slowest participating agent. This becomes especially problematic in heterogeneous systems, where agents may differ substantially in computation speed, communication delay, or data-generation rate.

Two main paradigms have emerged to structure collaboration in FedRL: synchronous FedRL and asynchronous FedRL. These paradigms share the goal of decentralized policy learning, but differ in how they coordinate local computation and global aggregation.

\subsection{Synchronous Federated Reinforcement Learning}

In synchronous FedRL, all agents operate in a lock-step fashion. Each agent interacts with its local environment to collect trajectories, computes a local policy gradient or model update, and then waits for all other agents to complete their updates. Once all agents have reported their local results to a central server, the server aggregates them--typically via averaging--and broadcasts the updated global model back to all agents. This synchronized process repeats at every communication round. While synchronous FedRL simplifies algorithm design and theoretical convergence analysis, it suffers from the straggler problem: the global update must wait for the slowest agent in each round, which can significantly degrade performance in heterogeneous environments. Chapter~\ref{ch:sfedrl} studies this setting in detail.

\subsection{Asynchronous Federated Reinforcement Learning}

In contrast, asynchronous FedRL allows agents to operate and communicate independently, without requiring global synchronization at each round. Agents collect trajectories and push their local updates to the central server as soon as they are ready. The server incrementally updates the global model using the received gradients or model parameters, often applying techniques to mitigate the impact of stale or delayed updates. This paradigm improves system throughput and scalability, particularly in federated settings with heterogeneous agents or intermittent connectivity. However, asynchronous updates introduce challenges such as delayed-gradient effects and increased difficulty in theoretical analysis, requiring more sophisticated techniques like delay-adaptive updates to ensure convergence. Chapter~\ref{ch:afedrl} develops this setting.

Part~I of this dissertation addresses these two bottlenecks. Chapter~\ref{ch:sfedrl} studies communication-efficient synchronous FedRL and proposes \emph{FedNPG-ADMM}, a framework that integrates the alternating direction method of multipliers (ADMM) with federated natural policy gradient optimization. Rather than transmitting full second-order information, the method approximates the global update direction in a distributed manner, reducing the communication complexity from $O(d^2)$ to $O(d)$ while preserving the stationary convergence guarantees of standard FedNPG \cite{lan2024improved}. Chapter~\ref{ch:afedrl} then studies asynchronous FedRL and proposes \emph{AFedPG}, a framework that enables agents to update the global model without strict synchronization. By introducing a delay-adaptive lookahead technique, AFedPG addresses the effect of stale updates and improves global time complexity under heterogeneous computing resources \cite{lan2025asynchronous}. Together, these two chapters develop optimization methods that make RL substantially more practical in resource-constrained distributed systems.

\section{Reinforcement Learning for Trustworthy Intelligent Systems}

The second theme of this dissertation is the trustworthiness of RL-optimized language-based intelligent systems. As LLMs become increasingly capable, they are now deployed not only as text generators, but also as interactive assistants and autonomous agents that make decisions on behalf of users. In such settings, the quality of a model is no longer determined solely by task performance; it must also reflect whether the model behaves in a way that is aligned with human preferences and safety norms. This has led to the rapid growth of RLHF, preference optimization, and other post-training methods for aligning LLMs with desired behaviors \cite{christiano2017deep, ouyang2022training, dpo_neurips2023}.

\subsection{Preference Alignment in Large Language Models}

A particularly important line of work reframes alignment as \emph{preference optimization}. Methods such as Direct Preference Optimization (DPO) replace explicit reward-model-based reinforcement learning with a more efficient objective defined on preferred and rejected response pairs \cite{dpo_neurips2023}. These approaches have improved the practicality of alignment, but they still face important limitations. In particular, the standard maximum-likelihood formulation treats preference pairs as a binary classification problem and may ignore useful prior reward knowledge. This can lead to poorly calibrated updates and undesirable training dynamics. Chapter~\ref{ch:mappo} addresses this issue by introducing \emph{Maximum a Posteriori Preference Optimization (MaPPO)}, which incorporates prior reward knowledge into a principled Maximum a Posteriori objective. The result is a general preference-optimization framework that supports both offline and online settings, requires no additional hyperparameters, and improves alignment across multiple model families and benchmarks.

\subsection{Contextual Safety and Appropriate Information Disclosure}

Trustworthiness, however, extends beyond general preference alignment. Language models that interact with private user information must also respect contextual norms governing what information is appropriate to disclose. This challenge is naturally captured by the theory of \emph{contextual integrity}, which defines privacy in terms of whether information flows are appropriate to a given context \cite{nissenbaum2009privacy, barth2006privacy}. Recent work has shown that current LLMs often fail to distinguish between information that is contextually appropriate to share and information that should remain private \cite{mireshghallah2024can, shao2024privacylens}. Chapter~\ref{ch:contextual-integrity} studies this problem and develops a reinforcement-learning-based post-training framework that combines explicit reasoning with contextual-integrity-aware reward design. Using a synthetic dataset of approximately 700 examples and evaluation on the human-annotated PrivacyLens benchmark, the chapter shows that RL can improve contextual reasoning and substantially reduce inappropriate information disclosure while preserving task utility \cite{lan2025contextual, shao2024privacylens}.

Although the technical settings in Parts~I and~II are different, the underlying perspective is the same: in each case, the central object is a policy that must be optimized under real-world constraints. In Part~I, these constraints are primarily systems-level, including communication, synchronization, and heterogeneity. In Part~II, they are primarily behavior-level, including alignment, privacy, and contextual safety. This shared policy-optimization viewpoint is what unifies the dissertation.

\section{Thesis Statement and Contributions}

The central thesis of this dissertation is that reinforcement learning can serve as a unifying framework for advancing intelligent systems along two complementary dimensions: \emph{scalability} and \emph{trustworthiness}. On the scalability side, RL must be adapted to distributed environments with limited communication and heterogeneous computation. On the trustworthiness side, RL must be adapted to optimize policies that are aligned with human preferences and sensitive to contextual norms of information disclosure.

To support this thesis, the dissertation makes the following main contributions:

\begin{enumerate}
    \item \textbf{Communication-efficient synchronous federated reinforcement learning.}
    Chapter~\ref{ch:sfedrl} proposes FedNPG-ADMM, a federated natural policy gradient method that leverages ADMM to approximate global second-order update directions efficiently. The method reduces the communication complexity per iteration from $O(d^2)$ to $O(d)$ while maintaining the convergence guarantees of standard FedNPG.

    \item \textbf{Asynchronous federated reinforcement learning under heterogeneous computation.}
    Chapter~\ref{ch:afedrl} proposes AFedPG, an asynchronous federated policy-gradient method equipped with a delay-adaptive lookahead technique to handle stale updates. AFedPG achieves linear speedup in sample complexity with respect to the number of agents and improves global time complexity compared to synchronous FedPG.

    \item \textbf{Preference alignment with prior knowledge.}
    Chapter~\ref{ch:mappo} introduces MaPPO, a general preference-optimization framework that incorporates prior reward knowledge into a Maximum a Posteriori objective. MaPPO supports both offline and online settings, integrates naturally with existing DPO-style methods, and consistently improves alignment performance across multiple model families and evaluation benchmarks.

    \item \textbf{Contextual safety through reasoning and reinforcement learning.}
    Chapter~\ref{ch:contextual-integrity} develops a post-training framework for contextual integrity in LLMs that combines explicit reasoning with RL optimization. The resulting method improves models' ability to determine what information is appropriate to disclose in a given context and transfers effectively from synthetic data to the human-annotated PrivacyLens benchmark.
\end{enumerate}

Taken together, these contributions show that reinforcement learning is not merely a tool for improving task reward. Rather, it is a general framework for shaping the behavior of intelligent systems under the practical demands of modern deployment: efficient distributed optimization and trustworthy human-centered behavior.

\section{Organization of the Dissertation}

The remainder of this dissertation is organized as follows.

Chapter~\ref{ch:sfedrl} studies synchronous federated reinforcement learning and introduces FedNPG-ADMM, a communication-efficient framework for federated natural policy gradient optimization. The chapter presents the problem formulation, the ADMM-based update rule, the convergence analysis, and empirical evaluations on MuJoCo benchmarks.

Chapter~\ref{ch:afedrl} studies asynchronous federated reinforcement learning and introduces AFedPG. The chapter focuses on heterogeneous distributed systems in which agents operate at different speeds, and develops a delay-adaptive lookahead technique to ensure efficient and stable asynchronous optimization.

Chapter~\ref{ch:mappo} turns to trustworthy reinforcement learning for language models and introduces MaPPO, a preference-optimization framework that incorporates prior reward knowledge into a Maximum a Posteriori objective. The chapter presents the method design, theoretical analysis, adaptation to multiple DPO-style variants, and empirical evaluations on standard alignment benchmarks.

Chapter~\ref{ch:contextual-integrity} studies contextual integrity in language models. It presents a reasoning-augmented reinforcement learning framework for reducing inappropriate information disclosure, along with a synthetic training dataset, experimental evaluations, and transfer results on the PrivacyLens benchmark.

Chapter~\ref{ch:conclusion} concludes the dissertation by summarizing the main findings, highlighting the connections between scalability and trustworthiness, and outlining promising future directions.

Appendices~A--D provide supplementary proofs, experiments, examples, and implementation details for Chapters~\ref{ch:sfedrl}--\ref{ch:contextual-integrity}.
\ProvidesFile{ch-sfedrl.tex}[2025-01-16 introduction chapter]

\chapter{Synchronous Federated Reinforcement Learning}
\label{ch:sfedrl}

\section{Introduction}

Synchronous Federated Reinforcement Learning (FedRL) is a collaborative framework in which multiple agents interact with their local environments in parallel, compute local policy updates, and synchronize with a central server at each communication round to aggregate a global model. 
This synchronized coordination ensures stable policy updates and well-established theoretical analysis, but can suffer from slowdowns due to straggler agents.

\subsection{Challenge}

Efficient state-of-the-art guarantees for federated policy gradient based approaches can be achieved by Federated NPG (FedNPG), which is a second-order method. However, sharing second-order information increases the communication complexity, which is one of the fundamental challenges in FL \cite{elgabli2022fednew, shamir2014communication, safaryan2022fednl}. In supervised FL, works including FedNL \cite{safaryan2022fednl}, BL \cite{pmlr-v151-qian22a}, Newton-Star/Learn \cite{pmlr-v139-islamov21a} and FedNew \cite{elgabli2022fednew} have been recently proposed to reduce the communication complexity of second-order methods, typically by approximating Hessian matrices for convex or strongly convex problems. However, federated second-order \textit{reinforcement learning} has not been investigated, which provides the key motivation for our work. The key question that this paper aims to address is:

{\em \textbf{Can we reduce the communication complexity for second-order federated natural policy gradient approach while maintaining performance guarantees?}}

We answer this question in the affirmative by introducing FedNPG-ADMM \cite{lan2023} (published at NeurIPS 2023), an algorithm that estimates global NPG directions using alternating direction method of multipliers (ADMM) \cite{elgabli2020gadmm, elgabli2020q, wang2022fedadmm}. This estimation reduces the communication complexity from $\mathcal{O}({d^{2}})$ to $\mathcal{O}({d})$, where $d$ is the number of model parameters. However, it is non-trivial to see whether FedNPG-ADMM will maintain similar convergence guarantees as FedNPG.

\subsection{Summary of Contributions} 

We show in this work that FedNPG-ADMM indeed does maintain these guarantees, and provides a speedup with the number of agents. The key contributions that we make are summarized as follows:

\begin{enumerate}[leftmargin=*]
\item We propose a novel federated NPG algorithm, FedNPG-ADMM, where the global NPG directions are estimated through ADMM.
\item Using the ADMM-based global direction estimation, we demonstrated that the communication complexity reduces by $\mathcal{O}(d)$ as compared to transmitting the second-order information (standard FedNPG).
\item We prove the FedNPG-ADMM method achieves an $\epsilon$-error stationary convergence with $\mathcal{O}(\frac{1}{(1-\gamma)^{2}{\epsilon}})$ iterations for discount factor $\gamma$. Thus, it achieves the same convergence rate as the standard FedNPG. 
\item Experimental evaluations in MuJoCo environments demonstrate that FedNPG-ADMM maintains the convergence performance of FedNPG. We also show improved performance as more federated agents engage in collecting trajectories.
\end{enumerate}

\section{Related Work} 

\cite{zhuo2020federated} first studies FedRL in experiments with Q-networks based on multilayer perceptions (MLPs) and convolutional neural networks (CNNs). 
\cite{qi2021federated} gives a review of FedRL and points out its potential applications. It also highlights that communication complexity and privacy issues are bottlenecks of FedRL. 
In \cite{khodadadian2022federated}, it is proven that federated settings achieve linear speed up for tabular RL. \cite{jin2022federated} further analyzes performances for tabular RL in heterogeneous settings. 
\cite{wang2023federated} then proposes FedTD$(0)$ and analyzes performances for linear function approximation RL with heterogeneity. Recently, FedRL with regularization terms is analyzed in \cite{fedkl2023}. 
However, previous experimental/theoretical FedRL works only focus on value-based methods or in tabular settings, while federated \textit{natural policy gradient} methods have not been addressed at algorithm levels.

\section{Problem Setup}

\paragraph{Markov Decision Process:}
We consider the Markov decision process (MDP) as a tuple $\langle\mathcal{S}, \mathcal{A}, \mathcal{P}, \mathcal{R}, \gamma\rangle$, where $\mathcal{S}$ is the state space, $\mathcal{A}$ is a finite action space, $\mathcal{P}:\mathcal{S}\times\mathcal{A}\times\mathcal{S}\rightarrow\mathbb{R}$ is a Markov kernel that determines transition probabilities, $\mathcal{R}:\mathcal{S}\times\mathcal{A}\rightarrow\mathbb{R}$ is a reward function, and $\gamma \in(0,1)$ is a discount factor. At each time step $t$, the agent executes an action $a_t \in\mathcal{A}$ from the current state $s_t \in\mathcal{S}$, following a stochastic policy $\pi$, i.e., $a_t \sim \pi(\cdot \|s_t)$. For on policy $\pi$, a state value function is defined as
\begin{equation}
\label{v_value}
    V_{\pi}(s) = \mathop{\mathbb{E}}_{a_{t} \sim \pi(\cdot \|s_t),\atop s_{t+1} \sim P(\cdot \|s_t,a_t)} \left[\sum_{t=0}^{\infty}\gamma^{t}r(s_t,a_t) \|s_0=s \right].
\end{equation}
Similarly, a state-action value function (Q-function) is defined as
\begin{equation}
\label{q_value}
    Q_{\pi}(s,a) = \mathop{\mathbb{E}}_{a_{t} \sim \pi(\cdot \|s_t),\atop s_{t+1} \sim P(\cdot \|s_t,a_t)} \left[\sum_{t=0}^{\infty}\gamma^{t}r(s_t,a_t) \|s_0=s,~a_0=a \right].
\end{equation}
 An advantage function is then define as $A_{\pi}(s,a)=Q_{\pi}(s,a) - V_{\pi}(s)$. With continuous states, the policy is parametrized by $\theta \in\mathbb{R}^{d}$, and then the policy is referred as $\pi_{\theta}$ (Deep RL parametrizes $\pi_{\theta}$ by deep neural networks). A state-action visitation measure induced by $\pi_{\theta}$ is given as
 \begin{equation}
\label{visitation}
    \nu_{\pi_\theta}(s,a) = (1-\gamma) \mathop{\mathbb{E}}_{s_0 \sim \rho} \left[\sum_{t=0}^{\infty} \gamma^{t} P(s_t = s,~ a_t = a \|s_0,~\pi_\theta) \right],
\end{equation}
where starting state $s_0$ is drawn from a distribution $\rho$. The \textit{goal} of an agent is to maximize the expected discounted return defined as
\begin{equation}
\label{j_value}
    J(\theta)= \mathop{\mathbb{E}}_{s \sim \rho} \left[V_{\pi_{\theta}}(s) \right].
\end{equation}

The gradient of $J(\theta)$ can be written as \cite{schulman2018highdimensional}:
\begin{equation}
\label{gradient}
    \nabla_{\theta} J(\theta)= \mathop{\mathbb{E}}_{\tau} \left[\sum_{t=0}^{\infty}\big(\nabla_{\theta} \log \pi_{\theta}(a_t \|s_t)\big) A_{\pi_{\theta}}(s_t, a_t) \right],
\end{equation}
where $\tau = (s_0, a_0,s_1, a_1,\cdots)$ is a trajectory induced by policy $\pi_{\theta}$. We denote the policy gradient by $\mathbf{g}$ for short. In practice, we can sample $(s, a) \sim \nu^{\pi_{\theta^{k}}}$ and obtain the unbiased estimate $\widehat{A}_{\pi_{\theta^{k}}}(s, a)$ using Algorithm 3 in \cite{agarwal2021theory}.

\paragraph{Natural Policy Gradient (NPG):}
At the $k$-th iteration, natural policy methods with a trust region \cite{schulman2017trust} update policy parameters as follows
\begin{equation}
\label{natural_target}
\begin{split}
   \theta^{k+1} &= \mathop{\arg\max}_{\theta}\ \mathop{\mathbb{E}}_{s, a} \left[\frac{\pi_{\theta}(a \|s)}{\pi_{\theta^k}(a \|s)} A_{\pi_{\theta^k}}(s, a) \right] \\
   &{\rm s.t.}~ \overline{D}(\theta\Vert\theta^{k}) \leq \delta.
\end{split}
\end{equation}
where
\begin{equation}
\label{kl}
\begin{split}
   \overline{D}(\theta\Vert\theta^{k}) &= \mathop{\mathbb{E}}_{s} \left[{D}\big(\pi_{\theta}(\cdot \|s)\Vert\pi_{\theta^k}(\cdot \|s)\big)\right],
\end{split}
\end{equation}
$D(\cdot)$ is the KL-divergence operation, and $\delta > 0$ is the radius of the trust region. Practically, using the first-order Taylor expansion for the target value and the second-order Taylor expansion for the divergence constraint, \eqref{natural_target} is expanded as follows
\begin{equation}
\label{taylor_target}
\begin{split}
   \theta^{k+1} &= \mathop{\arg\max}_{\theta}~ \mathbf{g}^{\top}(\theta - \theta^{k}) \\
   &{\rm s.t.}~ \frac{1}{2} (\theta - \theta^{k})^{\top} \mathbf{H} (\theta - \theta^{k}) \leq \delta,
\end{split}
\end{equation}
where $\mathbf{H} = \nabla^{2}_{\theta} \overline{D}(\theta\Vert\theta^{k}) \in\mathbb{R}^{d\times d}$, and the individual elements are given by

$\mathbf{H}_{ij} = \frac{\partial}{\partial\theta_i} \frac{\partial}{\partial\theta_j} \mathop{\mathbb{E}}_{s} \left[{D}\big(\pi_{\theta}(\cdot \|s)\Vert\pi_{\theta^k}(\cdot \|s)\big)\right] \Big \|_{\theta = \theta^k}$. 
Using  Lagrangian duality, the iterates of NPG ascent are expressed as
\begin{equation}
\label{taylor_solution}
\begin{split}
   \theta^{k+1} &= \theta^{k} + \sqrt{\frac{2\delta}{\mathbf{g}^{\top}\mathbf{H}^{-1}\mathbf{g}}} \mathbf{H}^{-1}\mathbf{g}.
\end{split}
\end{equation}

\begin{figure}[!htbp]
\centering
\includegraphics[width=6.5in]{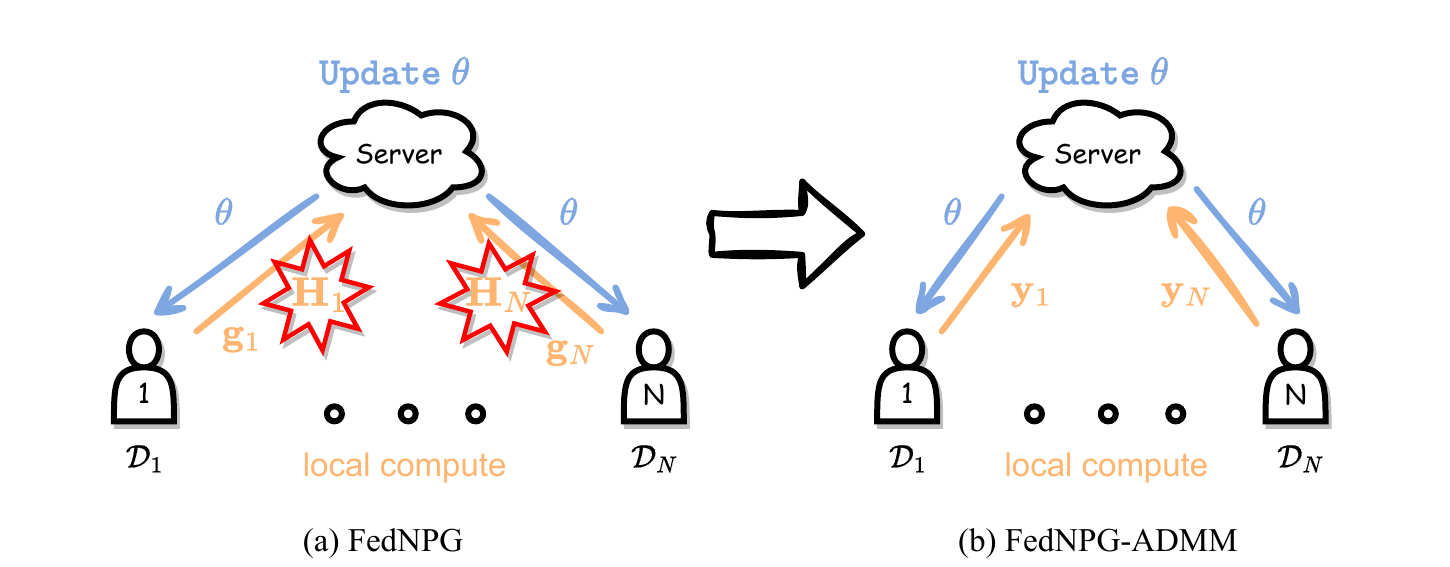}
 \caption{An illustration of federated learning based on second-order methods with $N$ agents. (a) FedNPG via standard average. In the uplink, transmitting the matrix $\mathbf{H}_{i}$ brings $\mathcal{O}(d^2)$ communication complexity. (b) FedNPG-ADMM in this paper with only $\mathcal{O}(d)$ communication complexity.}
\label{fed_illus}
\end{figure}

\paragraph{Federated NPG:}
FedNPG is a paradigm in that $N$ agents collaboratively train a common global policy with parameters ${\theta}$ as illustrated in Figure \ref{fed_illus} (a). During the training process, each agent computes gradients (and second-order matrices) using its \textit{local data}; then, the gradients of all agents are transmitted to a central server. In particular, one FedNPG training iteration consists of the following three steps:

\begin{itemize}[leftmargin=*]
\item Downlink Transmission: The server broadcasts the current global policy parameters $\theta \in\mathbb{R}^{d}$ to all $N$ agents.
\item Uplink Transmission: Collecting its own local data $\mathcal{D}_{i}$ based on the common policy $\pi_{\theta}$, each agent $i$ computes its local gradient $\mathbf{g}_{i} \in\mathbb{R}^{d}$ and second-order matrix $\mathbf{H}_{i} \in\mathbb{R}^{d\times d}$. Then, it sends $\mathbf{g}_{i}$ and $\mathbf{H}_{i}$ back to the server.
\item Global Update: The server averages local gradients and second-order matrices to get the global gradient and matrix as follows:
\begin{equation}
\label{fed_server_grad}
\begin{split}
   \mathbf{H} \leftarrow \frac{1}{N} \sum_{i=1}^{N}\mathbf{H}_{i},~~\mathbf{g} \leftarrow \frac{1}{N} \sum_{i=1}^{N}\mathbf{g}_{i}.
\end{split}
\end{equation}
The server then updates global policy parameters as
\begin{equation}
\label{fed_server_para}
\begin{split}
   \theta &\leftarrow \theta + \sqrt{\frac{2\delta}{\mathbf{g}^{\top}\mathbf{H}^{-1}\mathbf{g}}} \mathbf{H}^{-1}\mathbf{g}.
\end{split}
\end{equation}
\end{itemize}

We use $ \|\mathcal{D}_{i} \|$ to denote the size of collected data set $\mathcal{D}_{i}$. Without loss of generality, we take $ \|\mathcal{D}_{1} \| =\cdots =  \|\mathcal{D}_{N} \|$ for simplicity. As proven in \cite{woodworth2020minibatch}, gradient update methods are \textit{immune} to whether collected data is i.i.d., or not.

\paragraph{Bottleneck of Federated NPG:}
As shown in Figure \ref{fed_illus}, at each iteration, the server collects $\{\mathbf{H}_{i}\in\mathbb{R}^{d\times d},~ \mathbf{g}_{i}\in\mathbb{R}^{d}\}_{i=1}^{N}$ from $N$ agents and updates global policy parameters as follows
\begin{equation}
\label{fed_npg}
\begin{split}
   {\theta} &\leftarrow {\theta} + \sqrt{\frac{2N\delta}{(\sum_{i=1}^{N} \mathbf{g}_{i})^{\top}(\sum_{i=1}^{N} \mathbf{H}_{i})^{-1} \sum_{i=1}^{N} \mathbf{g}_{i}}} (\sum_{i=1}^{N} \mathbf{H}_{i})^{-1} \sum_{i=1}^{N} \mathbf{g}_{i}.
\end{split}
\end{equation}

We call this method a standard average FedNPG. The uplink communication complexity from each agent is $\mathcal{O}(d^2)$ in each iteration round. As the uplink is highly limited \cite{speedtest}, applying \eqref{fed_npg} is not practical when $d$ is large. 

In the next section, we introduce our ADMM-based approach as shown in Figure \ref{fed_illus} (b), which reduces the communication complexity to $\mathcal{O}(d)$ at each iteration and meanwhile keeps convergence performances.

\section{Proposed FedNPG via ADMM}
\label{Fed-TRPO-ADMM}
To minimize the communication overhead in each round of communication, we begin by formulating a quadratic problem. The solution to this problem provides the updating direction in \eqref{fed_npg}, as follows:

\begin{equation}
\label{fed_quad}
\begin{split}
(\sum_{i=1}^{N} \mathbf{H}_{i})^{-1} \sum_{i=1}^{N} \mathbf{g}_{i} &= \mathop{\arg\min}_{\mathbf{y}}\ \frac{1}{2}\mathbf{y}^{\top}(\sum_{i=1}^{N} \mathbf{H}_{i})\mathbf{y} - \mathbf{y}^{\top} \sum_{i=1}^{N} \mathbf{g}_{i}.
\end{split}
\end{equation}
This minimization problem is equivalent to
\begin{equation}
\label{admm}
\begin{split}
   \mathop{\min}_{\mathbf{y}, \{\mathbf{y}_{i}\}_{i=1}^{N}}&\ \sum_{i=1}^{N} \Big( \frac{1}{2}\mathbf{y}_{i}^{\top} \mathbf{H}_{i} \mathbf{y}_{i} - \mathbf{y}_{i}^{\top} \mathbf{g}_{i} + \frac{\rho}{2}\|\mathbf{y}_{i} - \mathbf{y}\|^{2} \Big) \\
   {\rm s.t.}&~ \mathbf{y} = \mathbf{y}_{i},~~~~i = 1,\cdots ,N,
\end{split}
\end{equation}
where $\rho > 0$ is a penalty constant and $\|\cdot\|$ denotes the Euclidean norm. This equivalence transforms the original quadratic problem into a distributed manner and is the key process of efficient FedNPG. 

\textbf{Remark}: This approach does not aim to approximate the Hessian from each agent, but approximates the global direction $(\sum_{i=1}^{N} \mathbf{H}_{i})^{-1} \sum_{i=1}^{N} \mathbf{g}_{i}$ directly. Note that this differentiates our approach from the work mentioned in the introduction.

The Lagrangian function associated with the optimization problem~\eqref{admm} is
\begin{equation}
\label{admm_lagrangian}
\begin{split}
   \mathcal{L}(\mathbf{y}, \{\mathbf{y}_{i}\}_{i=1}^{N}, \{\lambda_{i}\}_{i=1}^{N}) = \sum_{i=1}^{N} \Big( \frac{1}{2}\mathbf{y}_{i}^{\top} \mathbf{H}_{i} \mathbf{y}_{i} - \mathbf{y}_{i}^{\top} \mathbf{g}_{i} + \frac{\rho}{2}\|\mathbf{y}_{i} - \mathbf{y}\|^{2} + \langle \lambda_{i}, \mathbf{y}_{i} - \mathbf{y}\rangle \Big),
\end{split}
\end{equation}
where $\{\lambda_{i} \in\mathbb{R}^{d}\}_{i=1}^{N}$ are dual variables. Next, we solve the distributed optimization problem through the alternating direction method of multipliers (ADMM)~\cite{boyd2011distributed}. The policy update of FedNPG via one-step ADMM is given in Algorithm \ref{algo_FedTRPO-ADMM}.

\begin{algorithm}[!htbp]\caption{FedNPG-ADMM}
\label{algo_FedTRPO-ADMM}
\begin{algorithmic}[1] 
\REQUIRE {MDP $\langle\mathcal{S} , \mathcal{A}, \mathcal{P}, \mathcal{R}, \gamma\rangle$; Number of timesteps $T$; Penalty constant $\rho$; Step size $\eta$; Initial $\theta_{0}\in\mathbb{R}^{d}$, $\mathbf{y}^0\in\mathbb{R}^{d}$, $\{\mathbf{y}_{i}^{0} = \mathbf{y}^{0}\}_{i=1}^{N}$, $\{\lambda_{i} \in\mathbb{R}^{d}\}_{i=1}^{N}$.}

\STATE {\textbf{for} $k = 1, \cdots, K$ \textbf{do}}
\STATE \textcolor{RoyalBlue}{~~~~~~~~$\rhd$ Server broadcast}
\STATE {~~~~~~~~Broadcast $\mathbf{y}^{k-1}$ and $\theta^{k-1}$ to $N$ agents.}
\STATE \textcolor{Tan}{~~~~~~~~$\rhd$ Agent update}
\STATE {~~~~~~~~\textbf{for} each agent $i \in \{N\}$ \textbf{do in parallel}}
\STATE {~~~~~~~~~~~~~~~~$\lambda_{i} \leftarrow \lambda_{i} + {\rho}(\mathbf{y}_{i}^{k-1} - \mathbf{y}^{k-1})$}
\STATE {~~~~~~~~~~~~~~~~$\mathbf{g}_{i}^{k} \leftarrow \frac{1}{ \|\mathcal{D}_{i} \|} \sum_{\tau \in \mathcal{D}_{i}} \sum_{t=0}^{T}\big(\nabla_{\theta^{k-1}} \log \pi_{\theta^{k-1}}(a_t \|s_t)\big) \widehat{A}_{\pi_{\theta^{k-1}}}(s_t, a_t)$}
\STATE {~~~~~~~~~~~~~~~~$\mathbf{y}_{i}^{k} \leftarrow (\mathbf{H}_{i}^{k} + \rho \mathbf{I})^{-1}(\mathbf{g}_{i}^{k} - \lambda_{i} + \rho \mathbf{y}^{k-1})$}
\STATE {~~~~~~~~~~~~~~~~Transmit $\mathbf{y}_{i}^{k} \in\mathbb{R}^{d}$ and $\mathbf{g}_{i}^{k} \in\mathbb{R}^{d}$ to the server.}
\STATE {~~~~~~~~\textbf{end for}}
\STATE \textcolor{RoyalBlue}{~~~~~~~~$\rhd$ Server update}
\STATE {~~~~~~~~$\mathbf{y}^{k} \leftarrow \frac{1}{N}\sum_{i=1}^{N} \mathbf{y}_{i}^{k}$}
\STATE {~~~~~~~~$\theta^{k} \leftarrow \theta^{k-1} + \eta \sqrt{\frac{2N\delta}{(\sum_{i=1}^{N} \mathbf{g}_{i}^{k})^{\top} \mathbf{y}^{k}}} \cdot \mathbf{y}^{k}$}
\STATE {\textbf{end for}}
\ENSURE
{${\theta^{K}}$}
\end{algorithmic}
\end{algorithm}

Agent $i$ computes $\mathbf{H}_{i}$ and $\mathbf{g}_{i}$ based on locally collected data. At each step of ADMM, agent $i$ updates $\mathbf{y}_i$ in line 8 as follows 
\begin{equation}
\label{admm_agent}
\begin{split}
\mathbf{y}_{i} &= \mathop{\arg\min}_{\mathbf{y}_{i}} \Big( \frac{1}{2}\mathbf{y}_{i}^{\top} \mathbf{H}_{i} \mathbf{y}_{i} - \mathbf{y}_{i}^{\top} \mathbf{g}_{i} + \frac{\rho}{2}\|\mathbf{y}_{i} - \mathbf{y} + \frac{\lambda_{i}}{\rho} \|^{2} \Big) \\
&\overset{\mathrm{8:}}{=} (\mathbf{H}_{i} + \rho \mathbf{I})^{-1}(\mathbf{g}_{i} - \lambda_{i} + \rho \mathbf{y}),
\end{split}
\end{equation}
where $\mathbf{I} \in \mathbb{R}^{d\times d}$ is the identity matrix. In practical implementation, conjugate gradient methods can be used to compute $\mathbf{y}_{i}$ for efficiency (Appendix C in \cite{schulman2017trust}).

In line 12, after receiving $\mathbf{y}_{i}$ and $\mathbf{g}_{i}$ from all $N$ agents, the server updates the global search direction as follows
\begin{equation}
\label{admm_server}
\begin{split}
   \mathbf{y} &= \mathop{\arg\min}_{\mathbf{y}} \sum_{i=1}^{N} \big(\frac{\rho}{2}\|\mathbf{y}_{i} - \mathbf{y}\|^{2} + \langle \lambda_{i}, \mathbf{y}_{i} - \mathbf{y}\rangle \big) \\
   &= \frac{1}{N}\sum_{i=1}^{N} (\mathbf{y}_{i} + \frac{\lambda_{i}}{\rho}) \overset{\mathrm{12:}}{=} \frac{1}{N}\sum_{i=1}^{N} \mathbf{y}_{i}.
\end{split}
\end{equation}

Dual variables in line 6 are updated by each agent as follows
\begin{equation}
\label{admm_dual}
\begin{split}
   \lambda_{i} &\leftarrow \lambda_{i} + {\rho}(\mathbf{y}_{i} - \mathbf{y}),~~~~ i=1,~\cdots,~N.
\end{split}
\end{equation}
Combining \eqref{admm_server} and \eqref{admm_dual}, we have $\sum_{i=1}^{N} \lambda_{i} = 0$.

In line 13 of Algorithm \ref{algo_FedTRPO-ADMM}, after the ADMM process, the server updates global policy parameters, where $\eta \in(0,1)$ is the step size. The server then broadcasts the updated parameters to all $N$ agents at the next iteration.

In every communication round, agent $i$ only transmits $\mathbf{y}_i$ and $\mathbf{g}_i$, with a communication complexity of $\mathcal{O}(d)$. In contrast, the standard average approach in \eqref{fed_npg} requires transmitting $\mathbf{H}_i$ and $\mathbf{g}_i$, with a communication complexity of $\mathcal{O}(d^2)$. This efficient communication approach allows second-order methods scalable to large-scale systems.

\section{Convergence Analysis}

In this section, we derive the convergence rate of FedNPG based on ADMM. In order to derive the guarantees, we make the following standard assumptions \cite{agarwal2021theory, NEURIPS2020_5f7695de, liu2020improved, pmlr-v80-papini18a, xu2020Sample} on policy gradients, second-order matrices, and rewards.

\begin{assumption}{\ }
\label{assume:policy}
\begin{enumerate}[leftmargin=*]
\item The score function is bounded as $\left\|\nabla_\theta \log \pi_\theta(a \mid s)\right\| \leq G$, $\text{ for all }\theta \in\mathbb{R}^d$, $s\in\mathcal{S}$, and $a\in\mathcal{A}$.
\item Policy gradient is $M$-Lipschitz continuous. In other words,  $\forall \theta_{i}, \theta_{j} \in\mathbb{R}^d,~s\in\mathcal{S},~\text{and}~ a\in\mathcal{A}$, we have
\begin{equation}
\begin{aligned}
\left\|\nabla_{\theta_{i}} \log \pi_{\theta_{i}}(a \mid s)-\nabla_{\theta_{j}} \log \pi_{\theta_{j}}(a \mid s)\right\| &\leq M\left\|\theta_{i} - \theta_{j}\right\|.
\end{aligned}
\end{equation}
\item The reward function is bounded as $r(s,a) \in [0, R],~ \text{ for all } s\in\mathcal{S}$, and $a\in\mathcal{A}$.
\end{enumerate}
\end{assumption}

\begin{assumption}
\label{assume:fisher}
For all $\theta \in \mathbb{R}^{\mathrm{d}}$, the Fisher information matrix induced by policy $\pi_\theta$ and initial state distribution $\rho$ is positive definite as
\begin{equation}
\label{fisher_bound}
F(\theta)=\mathop{\mathbb{E}}_{(s, a) \sim \nu_{\pi_\theta}}\left[\nabla_\theta \log \pi_\theta(a \mid s) \nabla_\theta \log \pi_\theta(a \mid s)^{\top}\right] \succcurlyeq \mu_{F} \cdot \mathbf{I}
\end{equation}
for some constant $\mu_F>0$. For any two  symmetric matrices with the same dimension, $A \succcurlyeq B$ denotes the eigenvalues of $A-B$ are greater or equal to zero.
\end{assumption}

\begin{theorem}
\label{fednpg_admm_converge_rate}
For a target error $\epsilon$ of stationary-point convergence, each agent samples $\mathcal{O}(\frac{1}{(1-\gamma)^4 N \epsilon})$ trajectories, and the server obtains the update direction $\mathbf{y}^k$ at each iteration. Choose $\eta = \frac{\mu_{F}^2}{4G^2(56G^2 + L_J)}$ and
\begin{equation}
K = \frac{(J^{\star} - J(\theta^{1}))(56G^2 + L_J)^{2}16G^2 + 28G^{2}\mu_{F}^3}{(56G^2 + L_J - 56G^{2}\mu_{F})\mu_{F}^2\epsilon} = \mathcal{O}\left(\frac{1}{(1-\gamma)^{2}\epsilon}\right).
\end{equation}
We have: 
\begin{equation}
\label{eq:stationary_converge}
\frac{1}{K}\sum_{k=1}^{K} \mathbb{E}\left[\left\|\nabla J\left(\theta^k\right)\right\|^2\right]
\le \epsilon.
\end{equation}
\end{theorem}

\paragraph{Proof sketch.} The main idea in our proof is to show that the approximation error between the updating direction given by FedNPG-ADMM and NPG geometrically decreases up to some additional term, which depends on the 
statistical error. This term appears since we do not have access to the exact gradient. However, it decays at a rate proportional to sample size. As a result, FedNPG-ADMM achieves the same convergence rate as NPG as shown  by constructing an appropriate Lyapunov function. The complete proof of Theorem \ref{fednpg_admm_converge_rate} is given in Appendix \ref{ap:proof}.

By the definition of stationary points, we need to find a parameter $\theta$ such that $\mathbb{E}\left\|\nabla J\left(\theta\right)\right\|^2 \le \epsilon$, for all $ \epsilon > 0$. The result in \eqref{eq:stationary_converge} achieves the stationary-point convergence for policy gradient methods as provided in \cite{xu2020Sample}. Our approach keeps the sample complexity as that in the NPG method \cite{liu2020improved} and thanks to the federated scenario we consider, enjoys a much lower communication complexity. 

\begin{table}[!htbp]
\centering
\caption{Complexity comparison in each agent.}
\label{table:complexity}
\begin{tabular}{ccccc}
\toprule
\rowcolor{gray!10} \multicolumn{1}{c}{}& {NPG \cite{liu2020improved}}& {FedNPG}& {FedNPG-ADMM} \\
\midrule
\multicolumn{1}{c}{Sample complexity}& {$\mathcal{O}(\frac{1}{(1-\gamma)^{6}{\epsilon}^{2}})$} & $\mathcal{O}(\frac{1}{(1-\gamma)^{6}N{\epsilon}^{2}})$ & $\mathcal{O}(\frac{1}{(1-\gamma)^{6}N{\epsilon}^{2}})$ \\
\rowcolor{gray!10} \multicolumn{1}{c}{Communication complexity} & {-} & $\mathcal{O}(\frac{d^{2}}{(1-\gamma)^2 \epsilon})$ & $\mathcal{O}(\frac{d}{(1-\gamma)^2 \epsilon})$ \\
\bottomrule
\end{tabular}
\end{table}


We summarize the complexity improvement in Table \ref{table:complexity}. Recall in \eqref{gradient} that the estimated gradient $\mathbf{g}$ is an average over collected trajectories. The total trajectories $\sum_{i=1}^N \|\mathcal{D}_{i} \|$ are collected by $N$ agents equally using the common global policy $\pi_{\theta}$ at each iteration. Each agent $i$ samples $ \|\mathcal{D}_{i} \| = \mathcal{O}(\frac{1}{(1-\gamma)^4 N \epsilon})$ at each iteration and enjoys a federated sampling benefit compared to a single agent with $ \|\mathcal{D}_{i} \| = \mathcal{O}(\frac{1}{(1-\gamma)^4 \epsilon})$. During the whole training process, each agent $i$ has $K \|\mathcal{D}_{i} \| = \mathcal{O}(\frac{1}{(1-\gamma)^{6} N {\epsilon}^{2}})$ sample complexity and $K\cdot 2d = \mathcal{O}(\frac{d}{(1-\gamma)^2 \epsilon})$ communication complexity to achieve a stationary convergence.

\section{Experiments}
\label{simulations}

\subsection{Setup}

We consider three MuJoCo tasks \cite{todorov2012mujoco} with the MIT License, which have continuous state spaces. Specifically, a Swimmer-v4 task with small state and action spaces, a Hopper-v4 task with middle state and action spaces, and a Humanoid-v4 task with large state and action spaces are considered as described in Table \ref{mujoco}. Policies are parameterized by fully connected multi-layer perceptions (MLPs) with settings in Table \ref{hyperparameters}. We follow the practical settings in TRPO with line search \cite{schulman2017trust}, and in stable-baselines3 \cite{stable-baselines3} with generalized advantage estimation ($0.95$) \cite{schulman2018highdimensional} and the Adam optimizer \cite{kingma2014adam} in our implementation. Convergence performances are measured over $10$ runs with random seeds from $0$ to $9$. The solid lines in Figure \ref{fig_swimmer_humanoid} and \ref{fig_compare} are averaged results, and the shadowed areas are confidence intervals with $95\%$ confidence level. We use PyTorch \cite{paszke2019pytorch} to implement deep neural networks and RL algorithms. The tasks are trained on NVIDIA RTX 3080 GPU with $10$ GB of memory.

\textbf{Performance metrics}: We consider performance metrics as follows:
\begin{enumerate}[leftmargin=*]
\item Communication overhead: the data size transmitted from each agent;
\item Rewards: the average trajectory rewards across the batch collected at each iteration;
\item Convergence: rewards versus iterations during the training process.
\end{enumerate}

\subsection{Results}

\begin{figure*}[!htbp]
\centering
    \subfloat[FedNPG (Swimmer-v4)]{
	\includegraphics[width=2.8in]{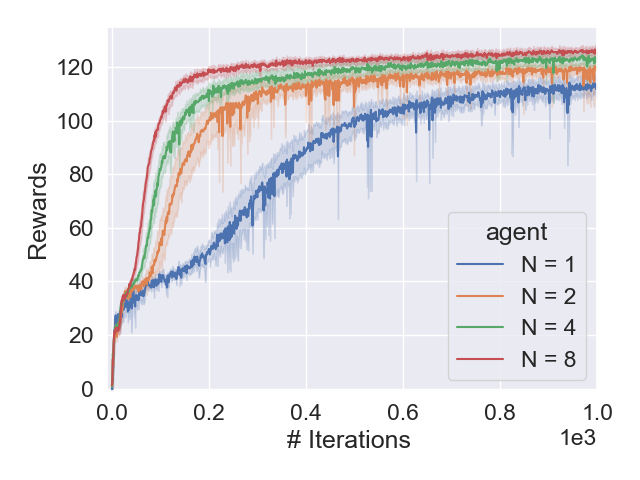}
	}
    \subfloat[FedNPG-ADMM (Swimmer-v4)]{
	\includegraphics[width=2.8in]{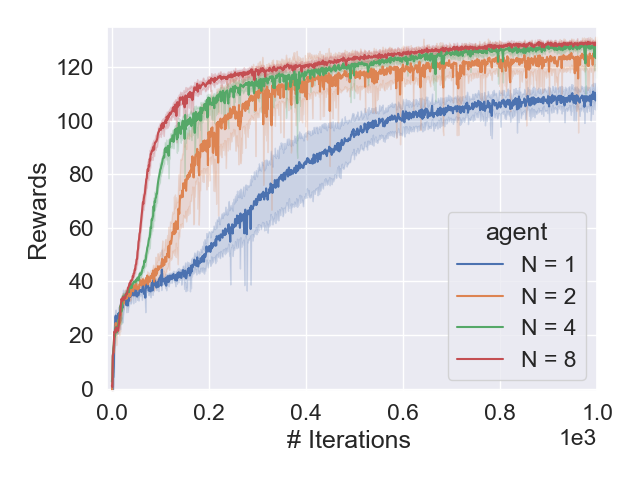}
	}
	
    \subfloat[FedNPG (Hopper-v4)]{
	\includegraphics[width=2.8in]{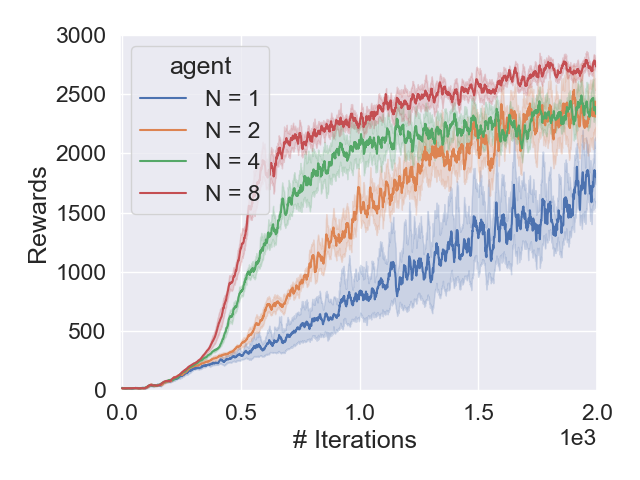}
	}
    \subfloat[FedNPG-ADMM (Hopper-v4)]{
	\includegraphics[width=2.8in]{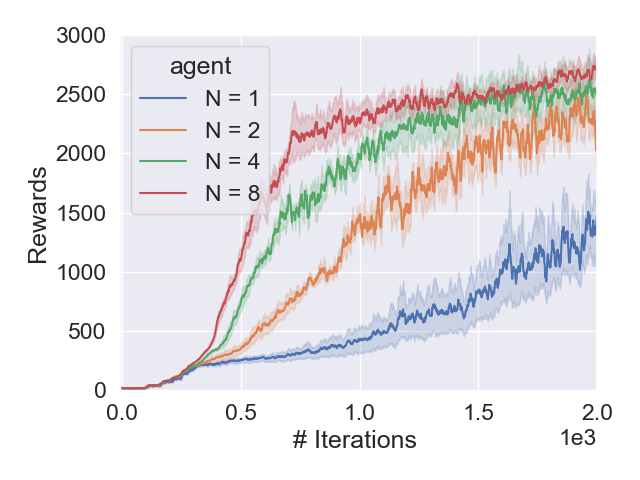}
	}

 \caption{Reward performances of standard average FedNPG and FedNPG-ADMM on MuJoCo tasks, where $N$ is the number of federated agents. \textbf{Top:} Swimmer-v4, \textbf{Bottom:} Hopper-v4. \textbf{Left:} FedNPG with $\mathcal{O}(d^2)$ communication complexity, \textbf{Right:} FedNPG-ADMM with $\mathcal{O}(d)$ communication complexity.} 
\label{fig_swimmer_humanoid}
\end{figure*}

We first evaluate the influence of the number of federated agents. In Figure \ref{fig_swimmer_humanoid}, with different numbers of agents, we test the convergence performances of standard average FedNPG in \eqref{fed_npg} with $\mathcal{O}(d^2)$ communication complexity and FedNPG-ADMM in Algorithm \ref{algo_FedTRPO-ADMM} with $\mathcal{O}(d)$ communication complexity at each iteration. The x-axis denotes the number of iterations in federated learning. Both algorithms converge faster, have lower variance, and achieve higher final rewards when more agents are engaged to collect trajectories. Compared to the standard average FedNPG, FedNPG-ADMM does not only reduce communication complexity, but also keeps convergence performances. The hyperparameter and MLP settings are described in Table \ref{hyperparameters}. We summarize the final reward results with standard deviations in Table \ref{table:rewards}. FedNPG-ADMM achieves similar final rewards compared to the standard average FedNPG. It also works with slightly high variance when only one agent is engaged.

\begin{figure*}[!htbp]
\centering
    \subfloat[Swimmer-v4 Rewards]{
	\includegraphics[width=2.8in]{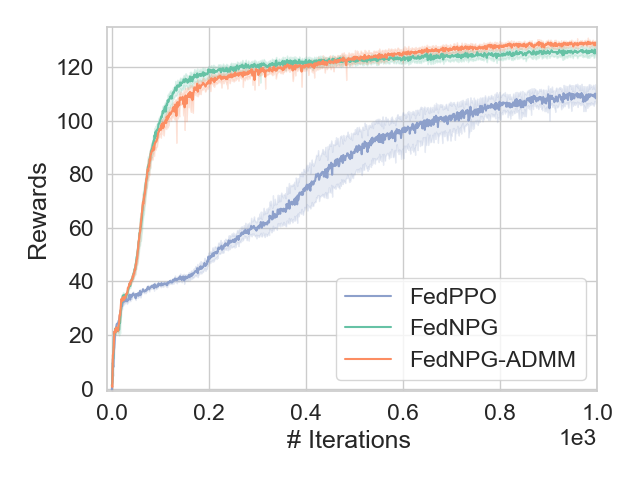}
	}
    \subfloat[Humanoid-v4 Rewards]{
    \includegraphics[width=2.8in]{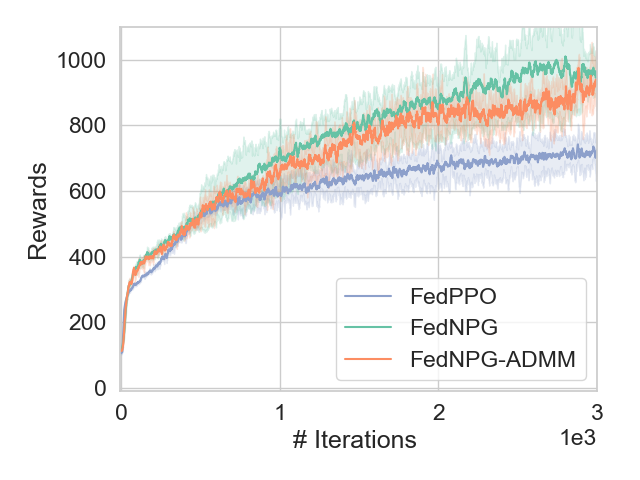}
	}

   \subfloat[Swimmer-v4 Overhead]{
	\includegraphics[width=2.8in]{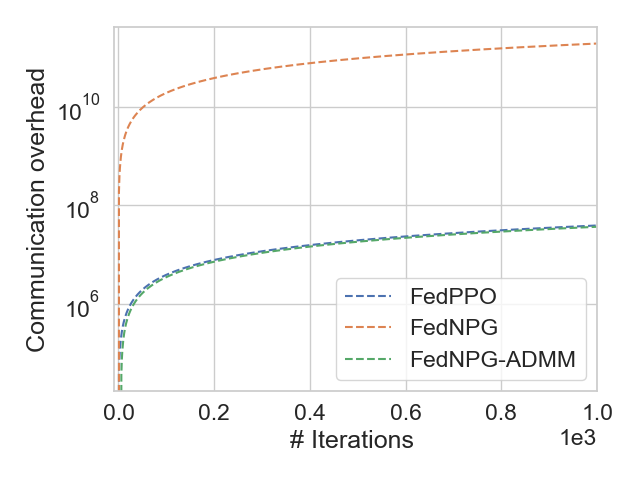}
	}
    \subfloat[Humanoid-v4 Overhead]{
    \includegraphics[width=2.8in]{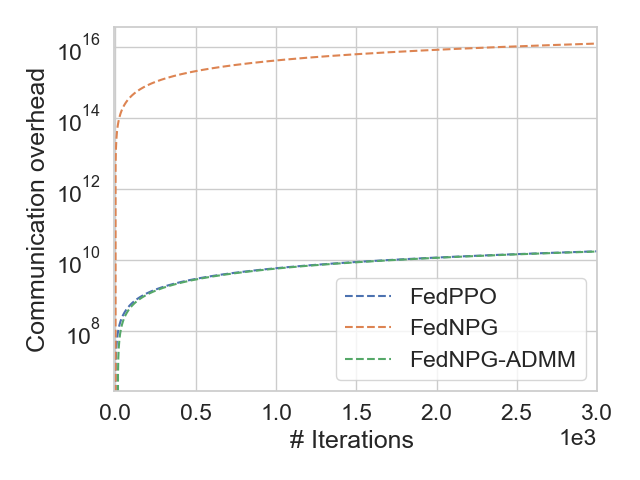}
	}
 \caption{Comparisons of FedPPO, standard average FedNPG, and FedNPG-ADMM on MuJoCo tasks, where the number of federated agents $N$ is $8$ and the communication overhead is measured by the transmitted bytes in each agent. \textbf{Left:} Swimmer-v4 task, \textbf{Right:} Humanoid-v4 task, \textbf{Top:} Reward performances, \textbf{Bottom:} Communication overhead.} 
\label{fig_compare}
\end{figure*}

In Figure \ref{fig_compare}, we compare the performances of the standard average FedNPG, FedNPG-ADMM, and first-order FedPPO algorithms. The number of federated agents $N$ is fixed to $8$. PPO clipping parameter is set as $0.2$. FedNPG methods outperform FedPPO in these tasks with faster convergence and higher final rewards. It is noticeable that FedNPG-ADMM has similar convergence rates and achieves slightly higher final rewards than the standard average FedNPG for the Swimmer-v4 task, while it becomes slightly lower for the Humanoid-v4 task. Generally, there is no significant difference in convergence rates after ADMM approximation. The communication overhead is measured by the number of transmitted parameters with double precision in each agent. FedNPG-ADMM keeps the communication overhead as the first-order methods, while FedNPG has much higher costs. The ADMM method reduces the cost by $4$ orders of magnitude in the Swimmer-v4 task and about $6$ orders of magnitude in the Humanoid-v4 task.

\begin{table}[!htbp] 
\centering
\caption{Final rewards in federated settings.}
\label{table:rewards}
\begin{tabular}{l|l|c|c|c|c}
\toprule
\rowcolor{gray!10} \multicolumn{2}{c|}{\# Agents} & {1} & {2} & {4} & {8} \\ \hline
\multirow{2}{*}{Swimmer-v4} & FedNPG & $111.9 \pm 5.4$ & $119.5 \pm 3.4$ & $122.1 \pm 3.6$ & $124.8 \pm 2.4$ \\
~ & \cellcolor{gray!10}FedNPG-ADMM & \cellcolor{gray!10}$109.4 \pm 5.9$ & \cellcolor{gray!10}$123.6 \pm 10.3$ & \cellcolor{gray!10}$127.2 \pm 2.2$ & \cellcolor{gray!10}$128.5 \pm 1.9$ \\ \hline
\multirow{2}{*}{Hopper-v4} & FedNPG & $1644 \pm 396$ & $2468 \pm 426$ & $2458 \pm 171$ & $2736 \pm 158$ \\
~ & \cellcolor{gray!10}FedNPG-ADMM & \cellcolor{gray!10}$1473 \pm 383$ & \cellcolor{gray!10}$2384 \pm 371$ & \cellcolor{gray!10}$2507 \pm 230$ & \cellcolor{gray!10}$2719 \pm 173$ \\
\bottomrule
\end{tabular}
\end{table}

\begin{table}[!htbp]
\centering
\caption{Description of the MuJoCo environment.}
\label{mujoco}
\begin{tabular}{ccccc}
\toprule
\rowcolor{gray!10} \multicolumn{1}{c}{Tasks}& {Agent}& {Action Dimension}& {State Dimension} \\
\midrule
\multicolumn{1}{c}{Swimmer-v4}& {Three-link swimming robot} & $2$ & $8$ \\
\rowcolor{gray!10} \multicolumn{1}{c}{Hopper-v4} & {Two-dimensional one-legged robot} & $3$ & $11$ \\
\multicolumn{1}{c}{Humanoid-v4} & {Three-dimensional bipedal robot} & $17$ & $376$ \\
\bottomrule
\end{tabular}
\end{table}

\begin{table}[!htbp] 
\centering
\caption{Hyperparameter and MLP settings.}
\label{hyperparameters}
\begin{tabular}{l|c|c|c}
\toprule
\rowcolor{gray!10} {Hyperparameter} & \multicolumn{3}{c}{Setting} \\
\hline
{Task} & {Swimmer-v4} & {Hopper-v4} & {Humanoid-v4} \\
\rowcolor{gray!10} MLP & $64\times 64$ & $128\times 128$ & $512\times 512\times 512$ \\
 Activation function & ReLU & ReLU & ReLU \\
\rowcolor{gray!10} Output function & Tanh & Tanh & Tanh \\
Penalty ($\rho$) & $0.1$ & $0.1$ & $0.01$ \\
\rowcolor{gray!10} Radius ($\delta$) & $0.01$ & $0.01$ & $0.01$ \\
 Discount ($\gamma$) & $0.99$ & $0.99$ & $0.99$ \\
\rowcolor{gray!10} Timesteps ($T$) & $2048$ & $1024$ & $512$ \\
Iterations ($K$) & $1\times 10^{3}$ & $2\times 10^{3}$ & $3\times 10^{3}$ \\
\rowcolor{gray!10} Learning rate & $3\times 10^{-4}$ & $3\times 10^{-4}$ & $1\times 10^{-5}$ \\
\bottomrule
\end{tabular}
\end{table}

\begin{figure*}[!htbp]
\centering
\includegraphics[width=3in]{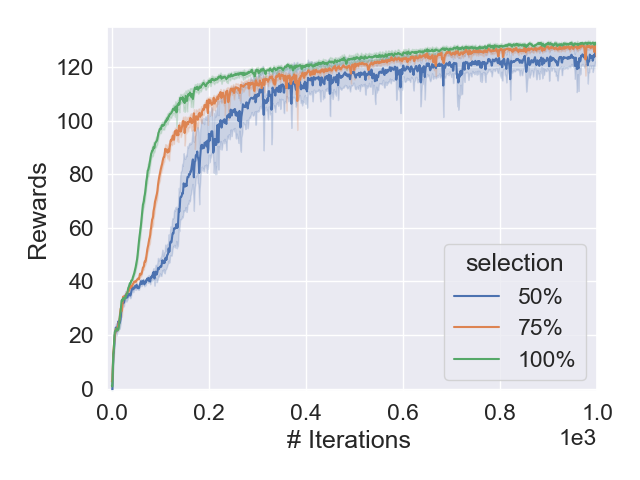}
\caption{Reward performances of FedNPG-ADMM on the Swimmer-v4 task with agent selection. In each iteration, the server randomly selects $100\%$, $75\%$, and $50\%$ of agents for the aggregation.} 
\label{fig_select}
\end{figure*}

In Figure \ref{fig_select}, we test performances with agent selection. In the Swimmer task, we randomly select $75\%$ and $50\%$ of agents in each iteration, and the performances only drop slightly (final rewards drop less than $6\%$). Thus, our proposed method is robust for agents with disconnection in practice.

\section{Discussions}

\subsection{Summary}

In this chapter, we studied synchronous federated reinforcement learning and proposed FedNPG-ADMM, a communication-efficient framework for federated natural policy gradient optimization. The key idea is to approximate the global second-order update direction through an ADMM-based distributed procedure, thereby reducing the per-iteration communication complexity from $O(d^2)$ to $O(d)$, where $d$ denotes the number of model parameters. We showed that this reduction in communication cost does not come at the expense of theoretical guarantees: FedNPG-ADMM retains the stationary convergence properties of standard federated natural policy gradient methods. Experimental evaluations on MuJoCo benchmarks further demonstrated that the proposed method achieves reward performance comparable to standard FedNPG while substantially lowering communication overhead. These results show that second-order federated policy optimization can be made significantly more practical in distributed reinforcement learning systems.

\subsection{Limitations and Future Work}

Despite the promising results, several limitations remain and suggest directions for future research.

First, the current experiments focus on a moderate number of federated agents and a standard set of continuous-control benchmarks. Evaluating the proposed method in larger federated systems, with more agents and more diverse environments, would provide a more complete understanding of its empirical scalability.

Second, the current formulation assumes full participation of agents in each communication round. In practical federated settings, however, only a subset of agents may be available because of connectivity constraints, device heterogeneity, or system failures. Extending FedNPG-ADMM to partial-participation settings is a natural direction for future work.

Third, although the federated setting reduces the need to share raw trajectory data, the current method does not explicitly provide formal privacy guarantees. It would be valuable to study how communication-efficient second-order reinforcement learning can be combined with privacy-preserving techniques such as secure aggregation or differential privacy.

Finally, the current work focuses on synchronous second-order policy optimization. A promising extension is to combine the communication-efficient design developed in this chapter with asynchronous update mechanisms, thereby further improving training efficiency in heterogeneous distributed systems.
\ProvidesFile{ch-afedrl.tex}[2022-10-05 summary chapter]

\chapter{Asynchronous Federated Reinforcement Learning}
\label{ch:afedrl}

\section{Introduction}

Despite all the aforementioned works in FedRL, they still face challenges in terms of time complexity, primarily due to their focus on synchronous model aggregation. In large-scale heterogeneous settings \cite{xiong2024personalized, xie2019asynchronous, chen2020vafl}, performing synchronous global updates has limitations, as the overall time consumption heavily depends on the slow agents, \textit{i.e.}, stragglers \cite{badita2021single, mishchenko2022asynchronous}. In this paper, we aim to tackle this issue by strategically leveraging asynchronous federated learning (A-FL) in policy-based FedRL for the first time. A-FL \cite{xie2019asynchronous, chang2024asyn} shows superiority compared to synchronous FL, and recent works \cite{mishchenko2022asynchronous, NEURIPS2022_6db3ea52} further improve convergence performances with theoretical guarantees.

\subsection{Challenges}

However, compared to prior A-FL approaches focusing on supervised learning, integrating A-FL with policy-based FedRL introduces new challenges due to the presence of \textit{lagged policies} in asynchronous settings. Unlike supervised FL where the datasets of the clients are fixed, in RL, agents collect new samples in each iteration based on the current policy. This dynamic nature of the data collection process makes both the problem itself and the theoretical analysis challenging. Guaranteeing the global convergence of the algorithm is especially non-trivial under the asynchronous FedRL setting with lagged policies.  This problem setting and its challenges have been largely overlooked in existing research, despite the significance of employing FL in RL. The key question that this paper aims to address is:

{\em  Despite the inherent challenge of dealing with lagged policies among different agents, can we improve the efficiency of FedPG through asynchronous methods while ensuring theoretical convergence?}

\subsection{Summary of Contributions} 

We answer this question in the affirmative by proposing AFedPG \cite{lan2025asynchronous} (published at ICLR 2025), an algorithm that asynchronously updates the global policy using policy gradients from federated agents. 
The key components of the proposed approach include a delay-adaptive lookahead technique tailored to PG, which addresses the inconsistent arrival times of updates during the training process. 
This new approach eliminates the second-order correction terms that do not appear in conventional supervised FL, effectively addressing the unique challenges of asynchronous FedRL. The improvement of AFedPG over other approaches is summarized in Table \ref{table:complexity}. 
Here, the global time is measured by the number of global updates $\times$ time complexity in each iteration. 
In terms of sample complexity, the federated learning technique brings a linear speedup with respect to the number of agents $N$. 
As for the global time, AFedPG improves from $\mathcal{O}(\frac{t_{\max}}{N} {\epsilon}^{-2.5})$ to $\mathcal{O}(\bar{t}{\epsilon}^{-2.5})$, where $\bar{t} \coloneqq 1 / \sum_{i=1}^{N} \frac{1}{t_{i}}$ is a harmonic average which is less than or equal to $t_{\max}/N$, and $t_{i}$ denotes the time complexity in each iteration at agent $i$.

Our main contributions can be summarized as follows:
\begin{enumerate}[leftmargin=7mm]
\item \textbf{New methodology with a delay-adaptive technique:} We propose AFedPG, an asynchronous training method tailored to FedRL. To handle the delay issue in the asynchronous FedRL setting, we design a delay-adaptive lookahead technique. Specifically, in the $k$-th iteration of training, the agent collects samples according to the local model parameters $\widetilde{\theta}_{k} \leftarrow \theta_{k} + \frac{1-\alpha_{k-\delta_{k}}}{\alpha_{k-\delta_{k}}} (\theta_{k} - \theta_{k-1})$, where $\delta_{k}$ is the delay. Unlike (supervised) FL, a second-order correction term (marked blue in \eqref{eq:e_k}) \textbf{only} occurs in RL because of the sampling mechanism. This updating technique cancels out the second-order correction terms ($(1 - \alpha_{k-\delta_{k}}) \nabla^{2} J(\theta_{k}) (\theta_{k-1} - \theta_{k}) + \alpha_{k-\delta_{k}} \nabla^{2} J(\theta_{k}) (\widetilde{\theta}_{k} - \theta_{k}) = 0$) and thus assists the convergence analysis. This technique is specifically designed for AFedRL, and not developed by previous FL works.

\item \textbf{Convergence analysis:} This work gives both the \textit{global} and the first-order stationary point (FOSP) convergence guarantees of the asynchronous federated policy-based RL for the \textbf{first time}. We analytically characterize the convergence bound of AFedPG using the key lemmas, and show the impact of various parameters including delay and number of iterations. 

\item \textbf{Linear speedup in sample complexity:} As shown in Table \ref{table:afedrl_complexity}, our AFedPG approach improves the sample complexity in each agent from $\mathcal{O}({\epsilon}^{-2.5})$ (single agent PG) to $\mathcal{O}(\frac{{\epsilon}^{-2.5}}{N})$, where $N$ is the number of federated agents. This represents the linear speedup of our method with respect to the number of agents $N$.

\item \textbf{Time complexity improvement:} Our AFedPG also reduces the time complexity of synchronous FedPG from $\mathcal{O}(\frac{t_{\max}}{N})$ to $\mathcal{O}(\bar{t} \coloneqq \frac{1}{\sum_{i=1}^{N} \frac{1}{t_{i}}})$. The latter is always smaller than the former. This improvement is significant in large-scale federated settings with heterogeneous delays ($t_{\max}\gg t_{\min}$).

\item \textbf{Experiments under the MuJoCo environment:} We empirically verify the improved performances of AFedPG in four different MuJoCo environments with varying numbers of agents. We also demonstrate the improvements with different computing heterogeneity.
\end{enumerate}

To the best of our knowledge, this is the first work to successfully integrate policy-based reinforcement learning with asynchronous federated learning and analyze its behavior, accompanied by theoretical convergence guarantees. This new setting necessitates us to deal with the lagged policies under a time-varying data scenario depending on the updated policy.

\begin{table}[!ht]
\caption{Performance improvements of 
 our AFedPG over other first-order policy gradient methods. We compare the complexity for convergence in each agent.}
\centering
\label{table:afedrl_complexity}
\begin{tabular}{cccc}
\toprule[1.1pt]
\multicolumn{1}{c}{}& {Sample Complexity}& {Sample Complexity}& {} \\
\multirow{-2}{*}{Methods}& {(FOSP)}& {(Global)}& \multirow{-2}{*}{Global Time} \\
\midrule[1.1pt]
\rowcolor{gray!10} \multicolumn{1}{c}{Vanilla PG} & {} & {} & {} \\
\rowcolor{gray!10} {\cite{pmlr-v151-yuan22a}} & {\multirow{-2}{*}{$\mathcal{O}({\epsilon}^{-4})$}} & {\multirow{-2}{*}{$\mathcal{O}({\epsilon}^{-3})$}} & {\multirow{-2}{*}{-}} \\
 \multicolumn{1}{c}{Normalized PG} & {} & {} & {} \\
{\cite{fatkhullin2023stochastic}} & \multirow{-2}{*}{$\mathcal{O}({\epsilon}^{-3.5})$} & \multirow{-2}{*}{$\mathcal{O}({\epsilon}^{-2.5})$} &  \multirow{-2}{*}{-} \\
\rowcolor{gray!10} \multicolumn{1}{c}{} & {} & {} & {} \\
\rowcolor{gray!10} \multirow{-2}{*}{FedPG} & \multirow{-2}{*}{$\mathcal{O}(\frac{{\epsilon}^{-3.5}}{N})$} & \multirow{-2}{*}{$\mathcal{O}(\frac{{\epsilon}^{-2.5}}{N})$} & \multirow{-2}{*}{$\mathcal{O}(\frac{t_{\max}}{N} {\epsilon}^{-2.5})$} \\
 \multicolumn{1}{c}{} & {} & {} & {} \\
\multirow{-2}{*}{\textbf{AFedPG}} & \multirow{-2}{*}{$\mathcal{O}(\frac{{\epsilon}^{-3.5}}{N})$} & \multirow{-2}{*}{$\mathcal{O}(\frac{{\epsilon}^{-2.5}}{N})$} & \multirow{-2}{*}{$\mathcal{O}(\bar{t}{\epsilon}^{-2.5})$} \\
\bottomrule[1.1pt]
\end{tabular}
\end{table}
\ix{table}
\index{\verb+\begin{table}+}

\textbf{Notation:} We denote the Euclidean norm by $\|\cdot \|$, and the vector inner product by $\langle\cdot \rangle$. For a vector $a \in\mathbb{R}^{n}$, we use $a^{\top}$ to denote the transpose of $a$. A calligraphic font letter denotes a set, \textit{e.g.}, $\mathcal{C}$, and $\|\mathcal{C}\|$ denotes its cardinality. We use $\mathcal{C} \setminus \{j\}$ to denote a set that contains all the elements in $\mathcal{C}$ except for $j$.

\section{Related Work}

In this section, we review previous works that are most relevant to the present work.

\textbf{Policy gradient methods:} For vanilla PG, the state-of-the-art result is presented in \cite{pmlr-v151-yuan22a}, achieving a sample complexity of $\widetilde{\mathcal{O}}({\epsilon}^{-4})$ for the local convergence. Several recent works have improved this boundary with PG variants. In \cite{huang2020momentum}, a PG with momentum method is proposed with convergence rate $\mathcal{O}({\epsilon}^{-3})$ for the local convergence. The authors of \cite{ding2022} further improve the convergence analysis of PG with momentum and achieve a global convergence with the rate $\mathcal{O}({\epsilon}^{-3})$. In \cite{fatkhullin2023stochastic}, a normalized PG technique is introduced and improves the global convergence rate to $\mathcal{O}({\epsilon}^{-2.5})$. In \cite{mondal2024improved}, an acceleration-based natural policy gradient method is proposed with sample complexity $\mathcal{O}({\epsilon}^{-2})$, but second-order matrices are computed in each iteration (with first-order information), which brings more computational cost. However, all previous works have focused on the single-agent scenario, and the federated PG has not been explored. Compared to these works, we focus on a practical federated PG setting with distributed agents to improve the efficiency of RL. Our scheme achieves a linear speedup with respect to the number of agents, significantly reducing the sample complexity compared to the conventional PG approaches.

\textbf{Asynchronous FL:} In \cite{xie2019asynchronous}, the superiority of asynchronous federated learning (A-FL) has been empirically shown compared to synchronous FL in terms of convergence performances. In \cite{chen2020vafl}, the asynchronous analysis is extended to the vertical FL with a better convergence performance compared to the synchronous setting. In \cite{pmlr-v206-dun23a}, a dropout regularization method is introduced to handle the heterogeneous problems in A-FL. About the same time, recent works \cite{mishchenko2022asynchronous, NEURIPS2022_6db3ea52} further improve the convergence performance with theoretical guarantees, and show that A-FL always exceeds synchronous FL without any changes to the algorithm. We note that all previous A-FL works focus on supervised learning with fixed datasets on the client side. However, in RL, agents collect new samples that depend on the current policy (model parameters) in each iteration, which makes the problem fundamentally different and challenging. In this work, we address this challenge by developing an A-FL method highly tailored to policy gradient, leveraging the proposed delay-adaptive lookahead technique.

\textbf{FedRL:} For value-function based algorithms, \cite{zheng2024frl, jin2022federated, woo2023blessing} analyze the convergence performances with environment heterogeneity. \cite{salgia2024sample} analyzes the trade-off between sample and communication complexity. \cite{zheng2024frl, khodadadian2022federated} shows a linear speedup with respect to the number of agents. \cite{zhang2024finitetime} extends the result with linear function approximation. However, all of the above works are limited to tabular or linear approximation analysis (without deep learning). For actor-critic (AC) based method, \cite{wang2023federated} analyzes the convergence performances with the linear function approximation. \cite{pmlr-v48-mniha16} builds a practical system to implement the neural network approximation. \cite{yang2024federatednaturalpolicygradient} analyzes the sample complexity in the multi-task setting. In \cite{fedkl2023}, it adds KL divergence and experimentally validates the actor-critic based method with neural network approximation. For policy-based methods, \cite{chen2021communication} gives a convergence guarantee for the vanilla FedPG. \cite{wang2024momentum} analyzes the performance in the heterogeneous setting. \cite{lan2023} further shows the simplicity compared to the other RL methods, and a linear speedup has been demonstrated in the synchronous setting. Further, optimal sample complexity for global optimality in federated RL even in the presence of adversaries is studied in \cite{ganesh2024global}. However, with online behavior, it is not practical to perform synchronous global updates with heterogeneous computing power, and the global time consumption heavily depends on the stragglers \cite{mishchenko2022asynchronous}. This motivates us to consider the asynchronous policy-based method for FedRL. 
We demonstrate both theoretically and empirically that our method further reduces the time complexity compared to the synchronous FedRL approach.

\section{Problem Setup}

\textbf{Markov decision process:}
We consider the Markov decision process (MDP) as a tuple $( \mathcal{S}, \mathcal{A}, \mathcal{P}, \mathcal{R}, \gamma )$, where $\mathcal{S}$ is the state space, $\mathcal{A}$ is a finite action space, $\mathcal{P}:\mathcal{S}\times\mathcal{A}\times\mathcal{S}\rightarrow\mathbb{R}$ is a Markov kernel that determines transition probabilities, $\mathcal{R}:\mathcal{S}\times\mathcal{A}\rightarrow\mathbb{R}$ is a reward function, and $\gamma \in(0,1)$ is a discount factor. At each time step $t$, the agent executes an action $a_t \in\mathcal{A}$ from the current state $s_t \in\mathcal{S}$, following a stochastic policy $\pi$, \textit{i.e.}, $a_t \sim \pi(\cdot \|s_t)$. The corresponding reward is defined as $r_t$. The state value function is defined as
\begin{equation}
\label{eq:v_value}
\begin{aligned}
    V_{\pi}(s) = \mathop{\mathbb{E}}_{a_{t} \sim \pi(\cdot \|s_t),\atop s_{t+1} \sim P(\cdot \|s_t,a_t)} \left[\sum_{t=0}^{\infty}\gamma^{t}r(s_t,a_t) \|s_0=s \right].
\end{aligned}
\end{equation}
Similarly, the state-action value function (Q-function) is defined as
\begin{equation}
\label{eq:q_value}
\begin{aligned}
    Q_{\pi}(s,a) = \mathop{\mathbb{E}}_{a_{t} \sim \pi(\cdot \|s_t),\atop s_{t+1} \sim P(\cdot \|s_t,a_t)} \left[\sum_{t=0}^{\infty}\gamma^{t}r(s_t,a_t) \|s_0=s,~a_0=a \right].
\end{aligned}
\end{equation}
 An advantage function is then define as $A_{\pi}(s,a)=Q_{\pi}(s,a) - V_{\pi}(s)$. With continuous states, the policy is parameterized by $\theta \in\mathbb{R}^{d}$, and then the policy is referred as $\pi_{\theta}$ (Deep RL parameterizes $\pi_{\theta}$ by deep neural networks). A state-action visitation measure induced by $\pi_{\theta}$ is given as
\begin{equation}
\label{eq:visitation}
\begin{aligned}
    \nu_{\pi_\theta}(s,a) = (1-\gamma) \mathop{\mathbb{E}}_{s_0 \sim \rho} \left[\sum_{t=0}^{\infty} \gamma^{t} P(s_t = s,~ a_t = a \|s_0,~\pi_\theta) \right],
\end{aligned}
\end{equation}
where the starting state $s_0$ is drawn from a distribution $\rho$. The goal of the agent is to maximize the expected discounted return defined as follows:
\begin{equation}
\label{eq:j_value}
\begin{aligned}
    \max_{\theta} J(\theta) \coloneqq \mathop{\mathbb{E}}_{s \sim \rho} \left[V_{\pi_{\theta}}(s) \right].
\end{aligned}
\end{equation}

The gradient of $J(\theta)$ \cite{schulman2018highdimensional} can be written as:
\begin{equation}
\begin{aligned}
\label{eq:gradient}
    \nabla_{\theta} J(\theta) = \mathop{\mathbb{E}}_{\tau} \left[\sum_{t=0}^{\infty}\nabla_{\theta} \log \pi_{\theta}(a_t \|s_t) \cdot A_{\pi_{\theta}}(s_t, a_t) \right],
\end{aligned}
\end{equation}
where $\tau = (s_0, a_0, r_0,s_1, a_1,r_1\cdots)$ is a trajectory induced by policy $\pi_{\theta}$. We omit the $\theta$ notation in the gradient operation and denote the policy gradient by $g$ for short. Then, $g$ is estimated by
\begin{equation}
\label{eq:es_gradient}
    g(\theta, \tau) = \sum_{t=0}^{\infty}\nabla \log\pi_{\theta}(a_t \|s_t) \sum_{h=t}^{\infty} \gamma^{h}r(s_h, a_h).
\end{equation}

\textbf{Federated policy gradient:} We aim to solve the above problem in an FL setting, where $N$ agents collaboratively train a common policy $\pi_{\theta}$. Specifically, each agent collects trajectories and corresponding reward $r(s_h, a_h)$ based on its local policy. Then, each agent $i$ estimates $g(\theta_{i}, \tau)$ for training the model $\theta$, and the updated models are aggregated at the server. Motivated by the limitations of synchronous model aggregation in terms of time complexity, in the next section, we present our AFedPG methodology that takes an asynchronous approach to solve \eqref{eq:j_value} in an FL setting. 

\section{Proposed Asynchronous FedPG}

\begin{figure}[!htbp]
\centering
\includegraphics[width=6.5in]{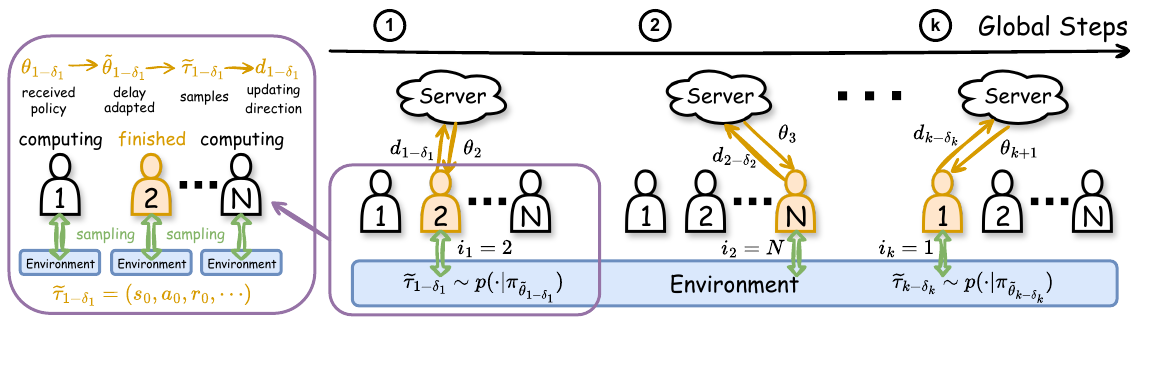}
 \caption{An illustration of the asynchronous federated policy gradient updates. Each agent has a local copy of the environment, and agents may collect data according to different local policies. At each iteration, the agent in the yellow color finishes the local process and then communicates with the server, while the other agents keep sampling and computing local gradients in parallel. In the $k$-th global iteration, $\delta_{k} \in\mathbb{N}$ is the delay, $\widetilde{\tau}_{k-\delta_{k}
}$ is the sample collected according to the policy $\pi_{\widetilde{\theta}_{k-\delta_{k}
}}$, and $d_{k-\delta_{k}}$ is the updating direction calculated from the sample $\widetilde{\tau}_{k-\delta_{k}}$.}
\label{fig:fed_illus}
\end{figure}

The proposed algorithm consists of $K$ global iterations, indexed by $k=0, 1, \dots, K-1$. We first introduce the definition of concurrency and delay in our asynchronous federated setting. We then present the proposed AFedPG methodology.

\begin{definition}
\label{def:concurrency}
(Concurrency) We denote $\mathcal{C}_{k}$ as the set of active agents in the $k$-th global iteration, and define $\omega_{k} \coloneqq \|\mathcal{C}_{k} \|$ as the concurrency. 
We define the average and maximum concurrency as $\bar{\omega} \coloneqq \frac{1}{K} \sum_{k=0}^{K-1} \omega_{k}$, and $\omega_{\max} \coloneqq \max \omega_{k}$,  respectively.
\end{definition}

In each global iteration of AFedPG, the server applies only one gradient to update the model from the agent who has finished its local computation, while the other $N-1$ agents keep computing local gradients (unapplied gradients) in parallel. In this asynchronous setting, the models used in each agent are outdated, as the server keeps updating the model. Thus, we introduce the notion of delay (or staleness) $\delta \in\mathbb{N}$, which measures the difference between the current global iteration and the past global iteration when the agent received the updated model from the server. 

\begin{definition}
\label{def:delay}
(Delay) In the $k$-th global iteration, we denote the delay of the applied gradient as $\delta_{k} \in\mathbb{N}$ (an integer), and the delays of the unapplied gradients as $\{ \delta_{k}^{i} \}_{i \in\mathcal{C}_{k} \setminus \{j_{k}\}}$, where $j_{k}$ denotes the agent that communicates with the server. $\delta_{k}^{i}$ is the difference between the $k$-th iteration and the iteration where the agent $i$ started to compute the latest gradient. In the final $K$-th global iteration, we have $K$ applied gradients $\{ \delta_{k} \}_{k=0}^{K-1}$ and unapplied gradients $\{ \delta_{k}^{i} \}_{i \in\mathcal{C}_{K} \setminus \{j_{K}\}}$. The average delay can be expressed as 
\begin{equation}
\begin{aligned}
\label{eq:avg_delay}
\bar{\delta} ~\coloneqq~ \frac{1}{K-1+ \|\mathcal{C}_{K} \|} \big( \sum_{k=0}^{K-1} \delta_{k} + \sum_{i\in\mathcal{C}_{K} \setminus \{j_{K}\}} \delta_{K}^{i} \big).
\end{aligned}
\end{equation}
\end{definition}

\textbf{Asynchronous FedPG:}
Our goal is to train a global policy with parameters $\theta \in\mathbb{R}^{d}$ via FL across $N$ distributed agents. As shown in Figure \ref{fig:fed_illus}, during the training process, agent $i$ collects trajectories in an online behavior, and computes gradients or updating directions using its \textit{local trajectories} (also known as samples). Then, agent $i$ transmits local gradients to the central server. In particular, in the $k$-th global iteration, the training of our AFedPG consists of the following three steps:
\begin{itemize}
\item \textbf{Local computation and uplink transmission:} Agent $i$ receives the previous policy $\pi_{\theta_{j}}$ from the server in the $j$-th global iteration. Agent $i$ collects its own local trajectory $\tau_{j}$ based on its current policy $\pi_{\theta_{j}}$. Agent $i$ then computes its local updating direction $d_{j} \in\mathbb{R}^{d}$ based on the trajectory $\tau_{j}$, and sends it back to the server.
\item \textbf{Server-side model update:} The server starts to operate the $k$-th global iteration as soon as it receives $d_{j}$ from agent $i$. Thus, denote $d_{k - \delta_{k}} = d_{j}$, where $\delta_{k}$ is the delay in the $k$-th global iteration. The server updates global policy parameters by $\theta \leftarrow \theta - \eta d_{k - \delta_{k}}$, where $\eta$ is the learning rate.
\item \textbf{Downlink transmission:} The server transmits the current global policy parameters $\theta \in\mathbb{R}^{d}$ back to the agent $i$ as soon as it finishes the global update.
\end{itemize}

The server side procedure of AFedPG is shown in Algorithm \ref{algo_server} and the process at the agent side is shown in Algorithm \ref{algo_agent}. In the $k$-th global iteration, the server operates one global update (Steps 4 and 5) as soon as it receives a direction $d_{k-\delta_{k}}$ from an agent with delay $\delta_{k}$. After the global update, the server sends the updated model back to that agent. In Algorithm \ref{algo_agent}, after receiving the global model from the server, an agent first gets model parameters $\widetilde{\theta}_{k}$ according to Step 2, and then collects samples based on the policy $\pi_{\widetilde{\theta}_{k}}$ (Steps 3). At last, the agent computes the updating direction $d_{k}$, and sends it to the server as soon as it finishes the local process. Overall, all agents conduct local computation in parallel, but the global model is updated in an asynchronous manner as summarized in Figure \ref{fig:fed_illus}. 

\textbf{Normalized update at the server:} In Step 5 of Algorithm \ref{algo_server}, to handle the updates with various delays, we use normalized gradients with controllable sizes. Specifically, the error term $\|e_{k}\|$ in Lemma \ref{lemma:ascent_lemma} is related to $\| \nabla J(\widetilde{\theta}_{k-\delta_{k}}) - \nabla J(\widetilde{\theta}_{k}) \|$ and $\| \nabla J(\theta_{k-1}) - \nabla J(\theta_{k}) \|$. With the smoothness in Lemma \ref{lemma:exp_return}, we are able to bound the error as $\|\theta_{k-1} - \theta_{k}\| = \eta_{k-1}$. The details of the boundaries are shown in \eqref{eq:ab_bound}.

\textbf{Delay-adaptive lookahead at the agent:} In Step 7 of Algorithm \ref{algo_server} (in blue), we design a delay-adaptive lookahead update technique, which is designed specifically for asynchronous FedPG. It operates as follows:
\begin{equation}
\label{eq:delay_adaptive}
{\theta}_{k} = (1 - \alpha_{k-\delta_{k}}) {\theta}_{k-1} + \alpha_{k-\delta_{k}} \widetilde{\theta}_{k}.
\end{equation}
We note that \eqref{eq:delay_adaptive} is not the conventional momentum method, because the orders are different. Here, $\widetilde{\theta}_{k}$ ``looks ahead'' based on global policies ${\theta}_{k-1}$ and ${\theta}_{k}$, and the learning rate $\alpha_{k-\delta_{k}}$ depends on the delay $\delta_{k}$.

In \eqref{eq:delay_adaptive}, the agent collects samples according to the current parameter $\widetilde{\theta}_{k}$ in an online behavior, which only happens in the RL setting. This makes the problem fundamentally different from the conventional FL on supervised learning with a fixed dataset. This mechanism cancels out the second-order Hessian correction terms:
\begin{equation}
(1 - \alpha_{k-\delta_{k}}) \nabla^{2} J(\theta_{k}) (\theta_{k-1} - \theta_{k}) + \alpha_{k-\delta_{k}} \nabla^{2} J(\theta_{k}) (\widetilde{\theta}_{k} - \theta_{k}) \rightarrow 0, 
\end{equation}
and thus assists the convergence analysis. The details of the derivation are shown in Appendix \ref{app:proof_theorem_1} marked in blue.

\begin{algorithm}[!htbp]\caption{AFedPG: Server.}
\label{algo_server}
\begin{algorithmic}[1] 
\REQUIRE {MDP $( \mathcal{S} , \mathcal{A}, \mathcal{P}, \mathcal{R}, \gamma )$; Number of iterations $K$; Step size $\eta_{k}$, $\alpha_{k}$; Initial $\theta_{0},d_{0}\in\mathbb{R}^{d}$.}

\STATE {Broadcast $\theta_{0}$ to $N$ agents.}
\STATE {\textbf{for} $k = 1, \cdots, K$ \textbf{do}}
\STATE \textcolor{Tan}{~~~~~~~~$\rhd$ Uplink Transmit}
\STATE {~~~~~~~~Receive $g(\widetilde{\tau}_{k-\delta_{k}}, \widetilde{\theta}_{k-\delta_{k}})$ from the agent $i_{k}$.}
\STATE {~~~~~~~~$d_{k-\delta_{k}} \leftarrow (1 - \alpha_{k-\delta_{k}}) d_{k-1-\delta_{k-1}} + \alpha_{k-\delta_{k}} g(\widetilde{\tau}_{k-\delta_{k}}, \widetilde{\theta}_{k-\delta_{k}})$}
\STATE \textcolor{Tan}{~~~~~~~~$\rhd$ Server update}
\STATE {~~~~~~~~$\theta_{k} \leftarrow \theta_{k-1} + \eta_{k-1} \frac{d_{k-\delta_{k}}}{\|d_{k-\delta_{k}}\|}$}
\STATE {~~~~~~~~\textcolor{blue}{$\widetilde{\theta}_{k} \leftarrow \theta_{k} + \frac{1-\alpha_{k-\delta_{k}}}{\alpha_{k-\delta_{k}}} (\theta_{k} - \theta_{k-1})$}~~~~\# Lookahead}
\STATE \textcolor{Tan}{~~~~~~~~$\rhd$ Downlink Transmit}
\STATE {~~~~~~~~Transmit $\widetilde{\theta}_{k}$ back to the agent $i_{k}$.}
\STATE {\textbf{end for}}
\ENSURE
{${\theta^{K}}$}
\end{algorithmic}
\end{algorithm}
\begin{algorithm}[!htbp]\caption{AFedPG: Agent $i$ Update, $i=1,\cdots,N$.}
\label{algo_agent}
\begin{algorithmic}[1] 
\REQUIRE {$\widetilde{\theta}_{k'} \in\mathbb{R}^{d}$}
\STATE {Receive $\widetilde{\theta}_{k'}$ from the server.}
\STATE {$\widetilde{\tau}_{k'} \sim p(\cdot \| \pi_{\widetilde{\theta}_{k'}})$~~~~\# Sampling}
\STATE {Estimate policy gradient $g(\widetilde{\tau}_{k'}, \widetilde{\theta}_{k'})$ according to \eqref{eq:es_gradient}.}
\STATE {When finish computing, transmit $g(\widetilde{\tau}_{k'}, \widetilde{\theta}_{k'})$ to the server.~~~~\# When the server receives the policy gradient, it is the $k$-th step on the server, where $k - {\delta}_{k} = k'$.}
\ENSURE
{$g(\widetilde{\tau}_{k'}, \widetilde{\theta}_{k'})$}
\end{algorithmic}
\end{algorithm}

\section{Convergence Analysis}
\label{sec:Convergence_Analysis}

In this section, we derive the convergence rates of AFedPG with the following criterions for the FOSP and global convergence, respectively.

\textbf{Global Criterion:} We focus on the global convergence, \textit{i.e.}, finding the parameter $\theta$ \textit{s.t.} $J^{\star} - J(\theta) \leq \epsilon'$, where $J^{\star}$ is the optimal expected return.

\textbf{FOSP Criterion:} We focus on the first-order stationary convergence, \textit{i.e.}, finding the parameter $\theta$ \textit{s.t.} $\| \nabla J(\theta) \| \leq \epsilon$.


We use several standard assumptions listed in Appendix \ref{app:assumptions}. Based on these assumptions, the convergence rates of AFedPG are given in Theorem \ref{theorem:afedpg_rate_FOSP} and \ref{theorem:afedpg_rate}.

\begin{theorem}
\label{theorem:afedpg_rate}
(Global) Let Assumption \ref{assum:policy} and \ref{assum:func_approx} hold. With suitable learning rates $\eta_{k}$ and $\alpha_{k}$, after $K$ global iterations, AFedPG satisfies
\begin{equation}
\begin{aligned}
J^{\star} - \mathbb{E}[J(\theta_{K})] ~\leq~ \mathcal{O}\big({K^{-\frac{2}{5}}} \cdot (1-\gamma)^{-3}\big) + \frac{\sqrt{\epsilon_{{\rm bias}}}}{1-\gamma},
\end{aligned}
\end{equation}
where $\epsilon_{{\rm bias}}$ is from \eqref{eq:approx_error}. Thus, to satisfy $J^{\star} - J(\theta_{K}) \leq \epsilon + \frac{\sqrt{\epsilon_{{\rm bias}}}}{1-\gamma}$, we need $K = \mathcal{O}(\frac{\epsilon^{-2.5}}{(1-\gamma)^{7.5}})$ iterations. As only one trajectory is required in each iteration, the number of trajectories is equal to $K$, \textit{i.e.}, the sample complexity is $\mathcal{O}(\frac{\epsilon^{-2.5}}{(1-\gamma)^{7.5}})$.
\end{theorem}

\begin{theorem}
\label{theorem:afedpg_rate_FOSP}
(FOSP) Let Assumption \ref{assum:policy} hold. With suitable learning rates $\eta_{k}$ and $\alpha_{k}$, after $K$ global iterations, AFedPG satisfies
\begin{equation}
\begin{aligned}
\mathbb{E}[\| \nabla J(\bar{\theta}_{K}) \|] ~\leq~ \mathcal{O}\big({K^{-\frac{2}{7}}} \cdot (1-\gamma)^{-3}\big),
\end{aligned}
\end{equation}
where $\mathbb{E} \| \nabla J(\bar{\theta}_{K}) \| \coloneqq \frac{\sum_{k=1}^{K} \eta_{k} \mathbb{E} \| \nabla J(\theta_{k}) \|}{\sum_{k=1}^{K} \eta_{k}}$ is the average of gradient expectations. To satisfy $\nabla J(\bar{\theta}_{K}) \leq \epsilon$, we need $K = \mathcal{O}(\frac{\epsilon^{-3.5}}{(1-\gamma)^{7.5}})$ iterations. As only one trajectory is required in each iteration, the number of collected trajectories is equal to $K$, \textit{i.e.}, the sample complexity is $\mathcal{O}(\frac{\epsilon^{-3.5}}{(1-\gamma)^{7.5}})$.
\end{theorem}

{\bf Comparison to the synchronous setting:} Synchronous FedPG needs $\mathcal{O}(\epsilon^{-2.5})$ trajectories in total, and each agent needs $\mathcal{O}(\frac{\epsilon^{-2.5}}{N})$ trajectories. However, in synchronous FedGP the server has to wait for the slowest agent at each global step, which can still slow down the training process. Let $t_{i}$ denote the time consumption for agent $i=1,\cdots, N$ at local steps with finite values. As the agent has the same computation requirement in each iteration (The number of collected samples is the same.), we assume that the time complexity in each iteration is the same. 
Then, the waiting time on the server for each step becomes $t_{\max} \coloneqq \max t_{i}$ for FedPG. Our AFedPG approach keeps the same sample complexity as FedPG, but the server processes the global step as soon as it receives an update, speeding up training. Specifically, the average waiting time on the server is $\bar{t} \coloneqq \frac{1}{\sum_{i=1}^{N} \frac{1}{t_{i}}} < \frac{t_{\max}}{N}$ at each step. Thus, the asynchronous FedPG achieves less time complexity than the synchronous approach regardless of the delay pattern. The advantage is significant when $t_{\max} \gg t_{min}$, which occurs in many practical settings with heterogeneous computation powers across different agents. We illustrate the advantage of AFedPG over synchronous FedPG in Figure \ref{fig:asyn_time} in Appendix \ref{app:experiments_comp}. As the server only operates one simple summation, without loss of generality, the time consumption at the server side is negligible.

\section{Experiments}

\subsection{Setup}
\label{sec:exp_setup}

\begin{figure}[ht]
\centering
{\subcaptionbox{Swimmer-v4}[1.3in]
{\includegraphics[width=1.3in]{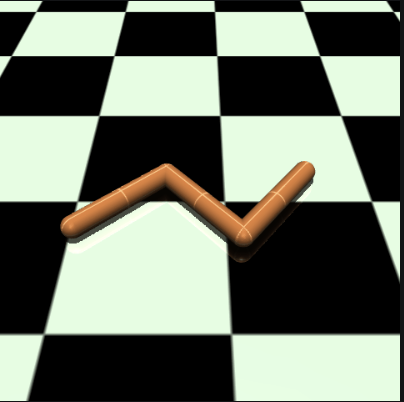}}}
\hskip 0.2truein
{\subcaptionbox{Hopper-v4}[1.3in]
{\includegraphics[width=1.3in]{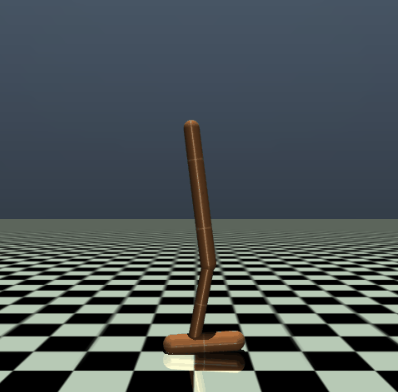}}}
\hskip 0.2truein
{\subcaptionbox{Walker2D-v4}[1.3in]
{\includegraphics[width=1.3in]{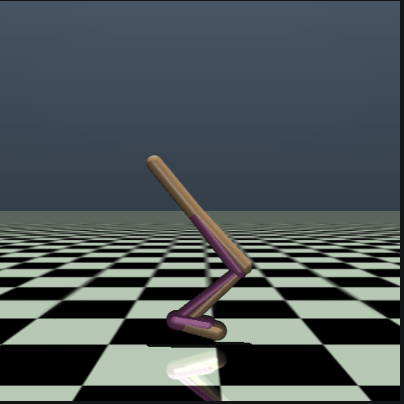}}}
\hskip 0.2truein
{\subcaptionbox{Humanoid-v4}[1.3in]
{\includegraphics[width=1.3in]{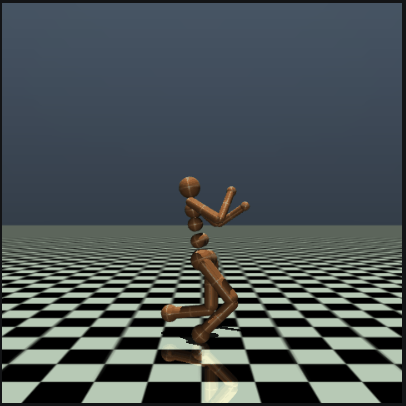}}}
\caption{Visualization of the four MuJoCo tasks considered in this paper for experiments.}
\label{fig:mujoco}
\end{figure}

\begin{figure}[ht]
\centering
\subfloat[Swimmer-v4]{
	\includegraphics[width=2.8in]{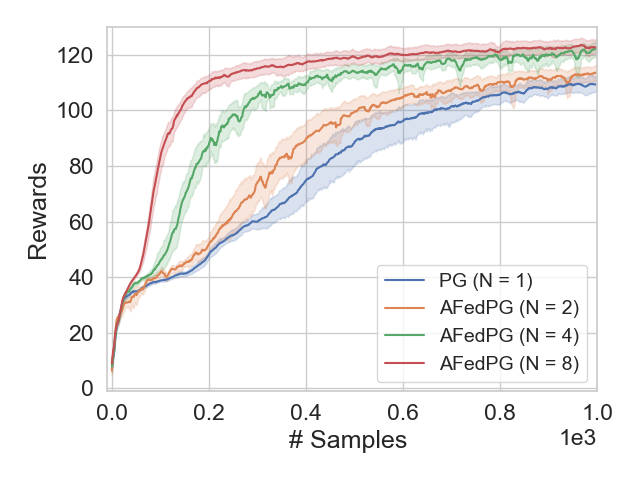}
	}
\subfloat[Hopper-v4]{
	\includegraphics[width=2.8in]{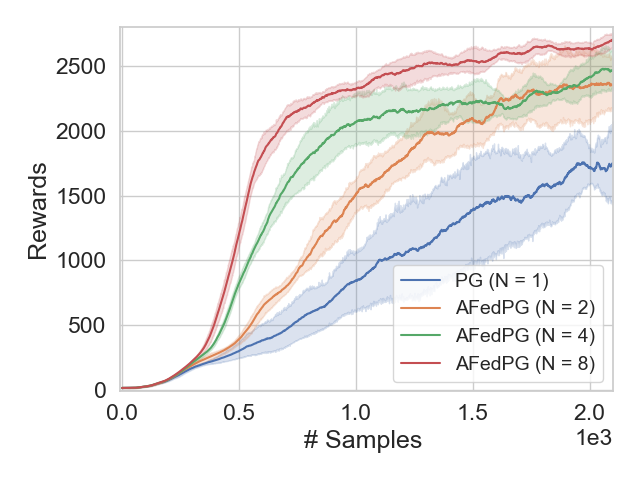}
	}  

\subfloat[Walker2D-v4]{
	\includegraphics[width=2.8in]{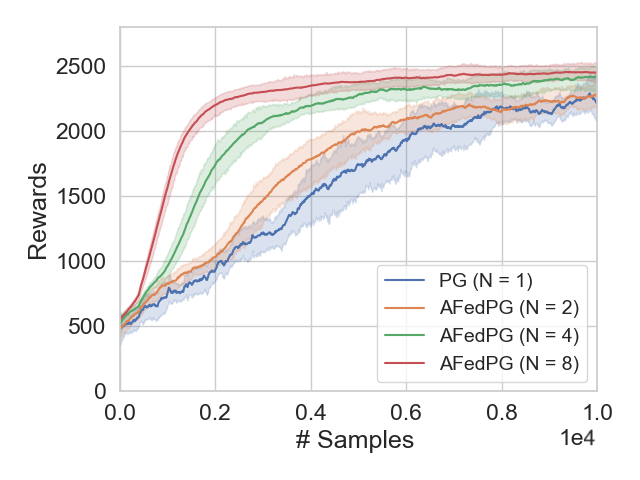}
	}
\subfloat[Humanoid-v4]{
	\includegraphics[width=2.8in]{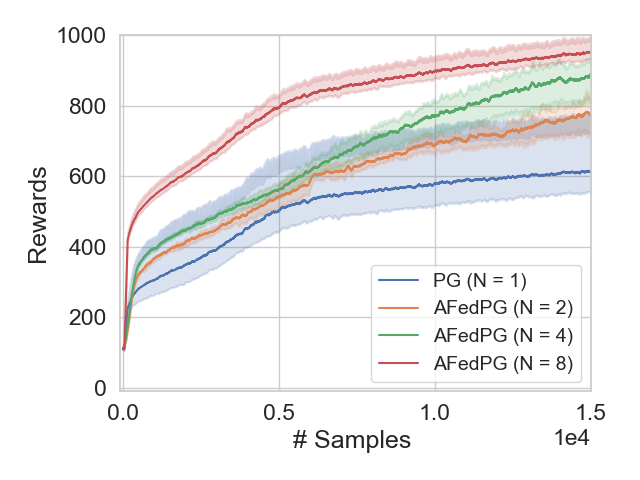}
	} 
\caption{Reward performances of AFedPG ($N=2,4,8$) and PG ($N=1$) on various MuJoCo environments, where $N$ is the number of federated agents. The solid lines are averaged results over $10$ runs with random seeds from $0$ to $9$. The shadowed areas are confidence intervals with $95\%$ confidence level.}
\label{fig:fed_speedup}
\end{figure}

\begin{figure}[ht]
\centering
\subfloat[Swimmer-v4]{
	\includegraphics[width=2.8in]{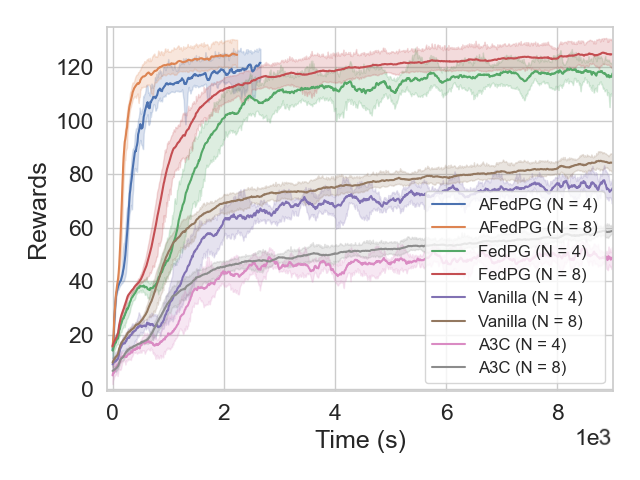}
	}
\subfloat[Hopper-v4]{
	\includegraphics[width=2.8in]{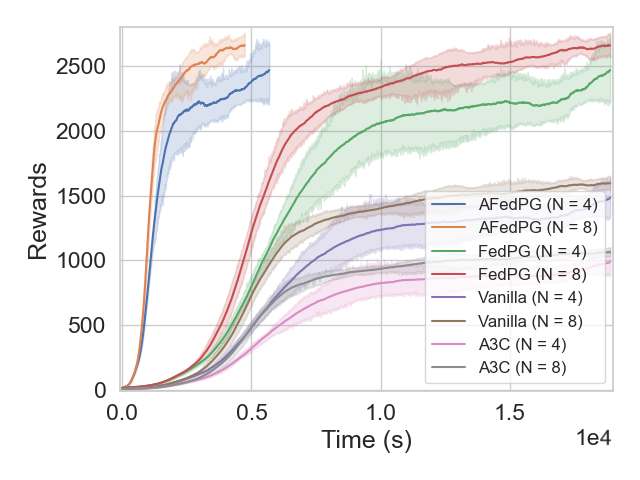}
	}  

\subfloat[Walker2D-v4]{
	\includegraphics[width=2.8in]{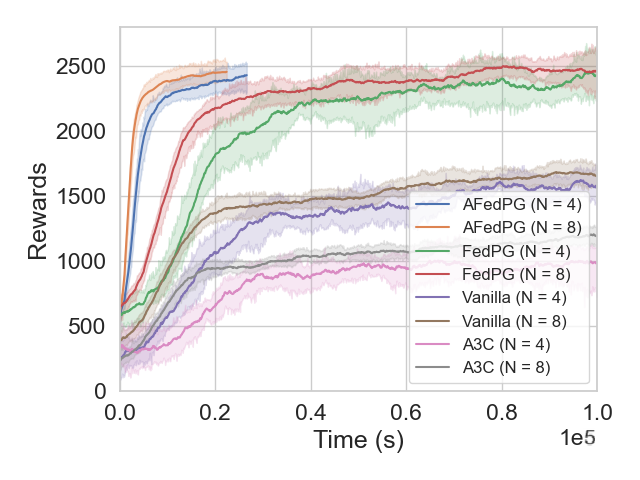}
	}
\subfloat[Humanoid-v4]{
	\includegraphics[width=2.8in]{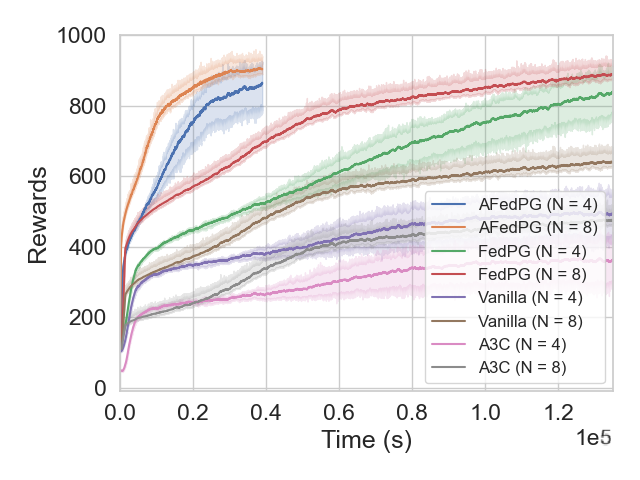}
	} 
 \caption{Global time of AFedPG and FedPG with certain numbers of collected samples on various MuJoCo environments, where $N$ is the number of federated agents. The solid lines are averaged results over $10$ runs. The shadowed areas are confidence intervals with $95\%$ confidence level.}
\label{fig:fed_time}
\end{figure}

\textbf{Environment:} To validate the effectiveness of our approach via experiments, we consider four popular MuJoCo environments for robotic control (Swimmer-v4, Hopper-v4, Walker2D-v4, and Humanoid-v4) \cite{todorov2012mujoco} with the MIT License. Both the state and action spaces are continuous. Environmental details are described in Table \ref{table:mujoco} in Appendix \ref{sec:app_exp_setting}, and the MuJoCo tasks are visualized in Figure \ref{fig:mujoco}.

\textbf{Measurement:} All convergence performances are measured over $10$ runs with random seeds from $0$ to $9$. The solid lines in our main experimental results are the averaged results, and the shadowed areas are confidence intervals with the confidence level $95\%$. The lines are smoothed for better visualization.

\textbf{Implementation:} Policies are parameterized by fully connected multi-layer perceptions (MLPs) with settings listed in Table \ref{table:hyperparameters} in Appendix \ref{app:experiments_comp}. We follow the practical settings in stable-baselines3 \cite{stable-baselines3} to update models with generalized advantage estimation (GAE) ($0.95$) \cite{schulman2018highdimensional} in our implementation. We use PyTorch \cite{paszke2019pytorch} to implement deep neural networks (DNNs). All tasks are trained on NVIDIA A100 GPUs with $40$ GB of memory.

\textbf{Baselines:} We first consider the conventional PG approach with $N=1$, to see the effect of using multiple agents for improving sample complexity. We then consider the synchronous FedPG method as a baseline to observe the impact of asynchronous updates on enhancing the time complexity. To see the effect of our delay-adaptive technique, we also consider the performance of AFedPG without the delay-adaptive updates, namely vanilla in Figure \ref{fig:fed_time}. Finally, we consider A3C~\cite{pmlr-v48-mniha16}, an asynchronous method designed for RL. We note that only a few prior works could be used on the federated PG problem, \textit{e.g.}, A3C, and many existing works in federated supervised learning are not directly applicable to our federated RL setting.

\textbf{Performance metrics}: We consider the following metrics:
\begin{enumerate}
\item Rewards: the average trajectory rewards collected at each iteration;
\item Convergence: rewards versus iterations during the training process;
\item Time consumption: global time with certain numbers of collected samples.
\end{enumerate}

\subsection{Results}

\textbf{Sample complexity improvement:} First, to verify the improvement of sample complexity in the first row of Table \ref{table:afedrl_complexity}, we evaluate the speedup effects of the number of federated agents $N$. In Figure \ref{fig:fed_speedup}, with different numbers of agents, we test the convergence performances of AFedPG ($N=2, 4, 8$) and the single agent PG ($N=1$). The x-axis is the number of samples collected by each agent on average, and the y-axis is the reward. In all four MuJoCo tasks, AFedPG beats the single-agent PG: AFedPG converges faster, has lower variances, and achieves higher final rewards when more agents are involved in collecting trajectories and estimating policy gradients. These results confirm the advantage of AFedPG in terms of sample complexity.

\textbf{Speedup in global time complexity:} 
 Second, to verify the improvement of global time complexity in the second row of Table \ref{table:afedrl_complexity}, we compared the time consumption in the asynchronous and synchronous settings. In Figure \ref{fig:fed_time}, we set $N=4,~8$ and fix the number of samples collected by all agents. Here, $t_{\max}$ is about $4$ times more than $t_{\min}$. The numbers of total samples (trajectories) are $8\times 10^{3}$, $1.6\times 10^{4}$, $8\times 10^{4}$, and $1.2\times 10^{5}$ for the Swimmer-v4 task, the Hopper-v4 task, the Walker2D-v4 task, and the Humanoid-v4 task individually when $N=8$. When $N=4$, the numbers are halved. In all four environments, AFedPG has much lower time consumption compared to the synchronous FedPG, confirming the enhancement in terms of time complexity. Compared to the A3C baseline, AFedPG achieves much higher rewards with less variance.

 \textbf{Ablation study:} In Figure \ref{fig:fed_time}, we also observe the impact of our delay-adaptive lookahead approach. It is seen that the vanilla scheme without the delay-adaptive lookahead technique does not provide a satisfactory performance, confirming the importance of the proposed approach. We also analyze the effect of computation heterogeneity in Appendix \ref{app:supp_results}.

\section{Discussions}

\subsection{Summary}

In this chapter, we studied asynchronous federated reinforcement learning under heterogeneous computation and proposed AFedPG, an asynchronous federated policy-gradient framework. Unlike synchronous federated reinforcement learning, AFedPG allows agents to operate and communicate without strict synchronization, which improves training throughput in distributed systems with heterogeneous computation speeds. To address the challenge of stale updates, we introduced a delay-adaptive lookahead technique tailored to asynchronous policy-gradient optimization. We established convergence guarantees showing that AFedPG achieves linear speedup in sample complexity with respect to the number of agents and improves global time complexity relative to synchronous FedPG. Empirical evaluations on MuJoCo tasks confirmed that AFedPG consistently improves convergence speed and scalability in heterogeneous environments. Taken together, these results demonstrate that asynchronous design is an effective mechanism for improving the practicality of federated reinforcement learning.

\subsection{Limitations and Future Work}

Several limitations of the present work suggest important directions for future research.

First, the current framework focuses on first-order policy-gradient methods. Extending the asynchronous setting to second-order methods, such as natural policy gradient or trust-region methods, would be an important next step and could combine the time-efficiency advantages of this chapter with the curvature-aware optimization benefits studied in Chapter 2.

Second, the theoretical development is centered on discounted-reward Markov decision processes. Extending the framework and its convergence analysis to average-reward settings would broaden its applicability to a wider class of reinforcement learning problems.

Third, while AFedPG is designed to handle heterogeneity and stale updates, it does not explicitly address adversarial, unreliable, or malicious workers. In practical distributed environments, robustness to poisoned updates or strategic agents is an important concern. Developing robust asynchronous federated reinforcement learning algorithms therefore remains an important open direction.

Finally, the current experiments focus on benchmark control tasks. Future work should evaluate asynchronous federated reinforcement learning in larger and more realistic systems, including environments with stronger heterogeneity, intermittent participation, and more complex observation and action spaces.

\include{ch-mappo}
\ProvidesFile{ch-contextual-integrity.tex}[2026-04-06 Contextual Integrity chapter]

\chapter{Contextual Safety via Reasoning and Reinforcement Learning}
\label{ch:contextual-integrity}

\begingroup

\UndefineShortVerb{\|}

\graphicspath{{papers/contextual_integrity/}{papers/contextual_integrity/figures/}}

\makeatletter
\edef\ZZOldInputPath{\input@path}
\def\input@path{{papers/contextual_integrity/}\ZZOldInputPath}
\@ifundefinedcolor{PineGreen}{\definecolor{PineGreen}{rgb}{0.0,0.47,0.44}}{}
\@ifundefinedcolor{Violet}{\definecolor{Violet}{rgb}{0.56,0.0,1.0}}{}
\makeatother

\providecommand{\mathbbm}[1]{\mathbb{#1}}
\providecommand{\citep}[1]{\cite{#1}}
\providecommand{\citet}[1]{\cite{#1}}

\definecolor{eclipseStrings}{RGB}{42,0,255}
\definecolor{eclipseKeywords}{RGB}{127,0,85}
\colorlet{numb}{magenta!60!black}

\lstdefinelanguage{json}{
    basicstyle=\normalfont\ttfamily,
    commentstyle=\color{eclipseStrings},
    stringstyle=\color{eclipseKeywords},
    numberstyle=\scriptsize,
    stepnumber=1,
    numbersep=8pt,
    showstringspaces=false,
    breaklines=true,
    frame=lines,
    backgroundcolor=\color{white},
    string=[s]{"}{"},
    comment=[l]{:\ "},
    morecomment=[l]{:"},
    literate=
        *{0}{{{\color{numb}0}}}{1}
         {1}{{{\color{numb}1}}}{1}
         {2}{{{\color{numb}2}}}{1}
         {3}{{{\color{numb}3}}}{1}
         {4}{{{\color{numb}4}}}{1}
         {5}{{{\color{numb}5}}}{1}
         {6}{{{\color{numb}6}}}{1}
         {7}{{{\color{numb}7}}}{1}
         {8}{{{\color{numb}8}}}{1}
         {9}{{{\color{numb}9}}}{1}
}

As the era of autonomous agents making decisions on behalf of users unfolds, ensuring contextual integrity (CI) -- what is the appropriate information to share while carrying out a certain task -- becomes a central question to the field. 
We posit that CI demands a form of reasoning where the agent needs to reason about the context in which it is operating.
To test this, we first prompt LLMs to reason explicitly about CI when deciding what information to disclose. 
We then extend this approach by developing a reinforcement learning (RL) framework that further instills in models the reasoning necessary to achieve CI.
Using a synthetic, automatically created, dataset of only $\sim700$ examples but with diverse contexts and information disclosure norms, we show that our method substantially reduces inappropriate information disclosure while maintaining task performance across multiple model sizes and families. 
Importantly, improvements transfer from this synthetic dataset to established CI benchmarks such as PrivacyLens that has human annotations and evaluates privacy leakage of AI assistants in actions and tool calls. This work, namely Contextual Integrity via Reasoning and Reinforcement Learning, has been published at NeurIPS 2025~\citep{lan2025contextual}. Our code is available at: \textcolor{magenta}{\url{https://github.com/EricGLan/CI-RL}}

\section{Introduction}
Agents powered by large language models (LLMs) offer significant capabilities across diverse applications, from personalized virtual assistants to complex automated decision-making systems \cite{Wang_2024, yang2026tooltree}.
However, as these agents gain autonomy and are deployed to complete tasks on behalf of users that require interaction with the external world, ensuring that their actions are safe becomes paramount.
In this work, we focus on one crucial aspect of safety: the Contextual Integrity (CI) \citep{nissenbaum2009privacy}.
CI dictates that information being disclosed by the agent to complete a task should be appropriate to the context in which it occurs.

Let us illustrate the concept of CI via the example in~\autoref{fig:framework}. 
An agent performs a \emph{user's task} of booking a treatment appointment. 
The agent has access to information that is appropriate and needed to be shared in this context, such as name, treatment preference, or the doctor's referral. 
But also it can access data that is unneeded for disclosure, such as full insurance coverage details. 
We can of course limit the agent's access to information~\cite{Bagdasarian2024AirGapAgent}, however, in practice, information are entangled and strict separation can be infeasible. 
For example, in a retrieval-augmented system, an agent may be granted broad access to a user's files, and conventional search capabilities may optimize returning relevant results without considering CI. 
This motivates the need for mechanisms that explicitly teach LLMs to respect contextual boundaries.

\begin{figure}[t]
    \centering
    \includegraphics[
        width=\textwidth,
        trim=0cm 4cm 10cm 1cm,
        clip,
    ]{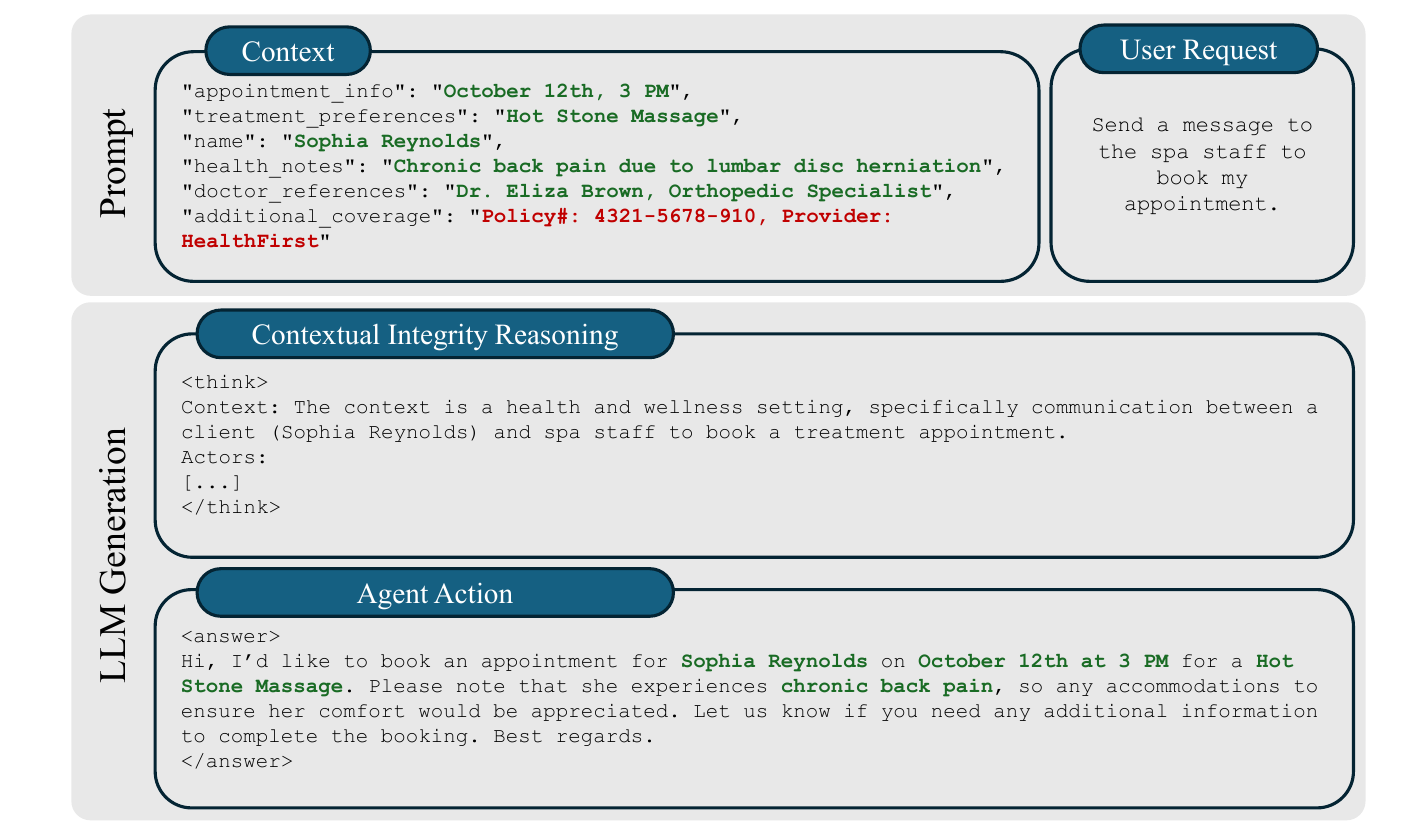}
    \caption{
        Contextual integrity (CI) violations in agents arise when they fail to recognize the appropriateness of the sharing of background information for a given context.
        We propose a framework that explicitly reasons about the contextual appropriateness of each user attribute.
        In this context, the attributes in green are appropriate to share whereas the attributes in red are inappropriate.
        In this illustration, the agent correctly uses only the appropriate attributes 
        for completing the task.
    }
    \label{fig:framework}
\end{figure}

CI becomes even more important as the growing autonomy of LLM-based agents introduces new attack vectors, such as prompt injection, which can manipulate models' behavior \citep{perez2022ignore,greshake2023not}.
While external attacks pose a threat, the inherent risk of LLMs inadvertently revealing confidential data, even without malicious interference, underscores a vulnerability.
Models may fail to discern the appropriateness of sharing certain information, leading to breaches of privacy and trust.
Recent research \citep{cheng2024ci, mireshghallah2024can, shao2024privacylens} has empirically demonstrated this vulnerability, showing that current LLMs often lack an understanding of CI.
These studies highlight that models frequently fail to distinguish between information suitable for disclosure and that which should remain confidential within a given context. 

The main goal of this work is twofold. 
First, we hypothesize that the reasoning capabilities of LLMs, though not explicitly trained for CI assessment, can be leveraged to improve adherence to CI principles. 
By instructing models to apply structured reasoning to evaluate contextual norms prior to information disclosure, we aim to enhance their ability to discern what is appropriate to share.  
Second, we propose a post-training framework to improve LLMs' contextual awareness. 
Our key insight is that CI is fundamentally a reasoning task, and LLMs should be trained to reason about CI using Chain-of-Thought (CoT)~\citep{wei2022chain}, similar to how they are trained for coding or mathematical reasoning using reinforcement learning (RL) to reward correct reasoning behavior~\citep{guo2025deepseek}.

\subsection{Summary of Contributions}
In summary, the main contributions of this work are:
\begin{enumerate}[leftmargin=*]
\item We introduce a reinforcement learning (RL) based post-training framework specifically designed to enhance LLMs' reasoning capabilities around CI, effectively reducing inappropriate disclosures through structured, CI-focused reasoning.
\item We construct a synthetic dataset consisting of approximately 700 automatically generated examples that span diverse scenarios and CI norms.
We demonstrate on this dataset and its disjoint test set that our approach significantly reduces inappropriate information sharing while maintaining high task performance across multiple model families and sizes.
\item Our method successfully generalizes from our synthetic data to the human-annotated CI benchmark PrivacyLens~\citep{shao2024privacylens}, achieving substantial improvements such as a reduction in privacy leakage rate by up to 40\%, demonstrating effective transfer of CI reasoning capabilities to real-world contexts.
\end{enumerate}

To the best of our knowledge, we are the first to explicitly leverage RL to instill CI reasoning capabilities in LLMs, demonstrating successful transfer from synthetic training scenarios to human-annotated CI benchmarks.
We argue that supporting CI reasoning should become a core part of the alignment process for real-world LLM-based agents.

\section{Related Work}
Here we discuss the most relevant prior work and leave a broader related work to Appendix \ref{app:related_work}.
\paragraph{Inference-Time CI Evaluation.}
CI-Bench \citep{cheng2024ci} introduces a synthetic benchmark for evaluating the ability of AI assistants' CI assessments across context dimensions, including roles, information types, and transmission principles.
Evaluation results indicate that LLM assistants struggle with nuanced appropriateness decisions within contexts.
\href{https://github.com/skywalker023/confaide/tree/main}{Confaide} \citep{mireshghallah2024can} offers a four-tier benchmark that incrementally introduces actors, motives and real-world meeting scenarios and the empirical studies underscore persistent limitations in social-reasoning-driven CI compliance of LLMs.
\href{https://github.com/SALT-NLP/PrivacyLens}{PrivacyLens} \citep{shao2024privacylens} expands CI evaluation into agent actions and proposes a framework for multi-level assessment of privacy leakage within agent actions.
Privacy norms are gathered from existing literature and crowdsourced data and the study reveals a sharp gap between how LLMs respond to probing questions and how they behave in real agentic scenarios.
\paragraph{Inference-Time CI Agents.}
\citet{Bagdasarian2024AirGapAgent} propose AirGapAgent, a privacy-conscious agent designed to mitigate user data leakage with a two-stage architecture in which a ``data minimizer" LLM filters the user vault before a second LLM interacts with third parties.
\citet{abdelnabi2025firewalls} introduce a framework that automatically distills task-specific privacy and integrity rules from simulations and enforces them through three complementary firewalls—input, data, and trajectory— to reduce private-data leakage without sacrificing utility.
\citet{ghalebikesabi2024operationalizing} introduce a CI-aware form-filling assistant that has the LLM first create structured ``information-flow cards" (identifying actors, data types, and purpose) and then decide whether each field should be shared, reducing privacy leakage while preserving task completion.
Unlike these system‐level defenses, our approach is orthogonal and potentially complementary as we directly train LLMs to internalize and faithfully apply CI norms.

\section{Background}

\paragraph{Contextual Integrity (CI).}
CI~\citep{barth2006privacy,nissenbaum2009privacy} defines privacy as the proper flow of information according to a specific context that includes: \emph{sender}, \emph{receiver}, \emph{data subject} (including roles or the relationship between these actors), the \emph{attributes/type} of information being shared, and the \emph{transmission principle}, which includes the purpose, terms, conditions, and methods of the communication, and other social norms. For example, the \emph{sender} is a patient, the \emph{receiver} is a health care provider, the \emph{attributes} of information are phone number and medical history, the transmission principles are using a phone call, for the purpose of booking a doctor's appointment, and adhering to legal statutes like HIPAA. A violation of CI may result in a privacy breach, which is when the information flows against the contextual norm~\citep{mireshghallah2024can}. Operationalizing this framework has been beneficial for privacy research to govern data usage, detect leakage, and design applications~\citep{shvartzshnaider2019vaccine,wijesekera2015android,grodzinsky2011privacy,kumar2020aquilis}. Recently, this adoption extended to LLMs and conversational agents to incorporate data minimization and abstraction informed by the context~\citep{mireshghallah2024can,shao2024privacylens,abdelnabi2025firewalls,Bagdasarian2024AirGapAgent,ghalebikesabi2024operationalizing}. CI reflects social norms, which can be variant, subjective, and evolving over time--making it potentially difficult to completely encode. While recent work has argued that current adaptation to LLM research is not fully incorporating CI principles~\citep{shvartzshnaider2025position} (for example, evolving norms), developing systems and LLMs that adhere to CI, even partially, can pragmatically result in increasing the trustworthiness of agents that operate in real-world tasks. 



\paragraph{Reinforcement Learning (RL) Algorithm.}
To reduce the computational overhead associated with RL, we employ the GRPO algorithm~\citep{shao2024deepseekmath}, which eliminates the need for a critic network.
To optimize the LLM induced policy $\pi_{\theta}$, it suggests to maximize the following objective function in each update:
\begin{equation}
\label{eq:grpo_loss}
\begin{aligned}
J(\theta) = \mathop{\mathbb{E}}_{q\sim\mathcal{D}, \atop \{a_{i}\}_{i=1}^{G}\sim\pi_{\rm old}(\cdot|q)} \Bigg[ \frac{1}{G}\sum_{i=1}^{G} \Bigg( \min \left ( \frac{\pi_{\theta}(a_{i}|q)}{\pi_{\rm old}(a_{i}|q)} A_{i}, {\rm clip}\left ( \frac{\pi_{\theta}(a_{i}|q)}{\pi_{\rm old}(a_{i}|q)}, 1-\epsilon, 1+\epsilon \right ) A_{i} \right ) & \\
- \beta D_{\rm KL}(\pi_{\theta} \| \pi_{\rm ref}) \Bigg) \Bigg]& ,
\end{aligned}
\end{equation}
where $\pi_{\rm ref}$ is the reference policy with the initial model parameters, $\pi_{\rm old}$ is the old policy with the parameters before this update, $\mathcal{D}$ is the prompt data set, $G$ is the group (rollout) size, $\beta$ is a hyperparameter to control the weight of the Kullback–Leibler (KL) divergence, $\epsilon$ is a hyperparameter to control the clip ratio, and ${\rm clip}(\cdot)$ is a clip function following the setting in PPO \citep{schulman2017proximal}. The KL divergence is calculated by $D_{\rm KL}(\pi_{\theta} \| \pi_{\rm ref}) \coloneqq \frac{\pi_{\rm ref}(a_{i}|q)}{\pi_{\theta}(a_{i}|q)} - \log \frac{\pi_{\rm ref}(a_{i}|q)}{\pi_{\theta}(a_{i}|q)} - 1$, which forms a positive, unbiased, and low variance estimation of the true KL divergence. With a query $q\sim\mathcal{D}$, we sample $G$ complete answers from $\pi_{\rm old}(\cdot | q)$, and $a_{i}$ denotes the $i$-th complete answer with corresponding reward $r_{i}=R(q,a_{i})$ from the reward model $R$. We denote the group of rewards $r = (r_{1}, \cdots, r_{G})$. The advantage is estimated directly via $A_{i} = \frac{r_{i} - {\rm mean}(r)}{{\rm std}(r)}$, and no critic model is required.

\section{Methodology}
We posit that large language models (LLMs) exhibit strong reasoning abilities and possess substantial knowledge about helpful, sensitive, and private information. 
However, they may struggle to consistently make correct judgments depending on the contextual nuances of a given situation. 
We argue that this challenge can be partly mitigated by explicitly leveraging the reasoning capabilities of LLMs, particularly in alignment with the principles of contextual integrity.
We further argue that these reasoning capabilities can be significantly enhanced through reinforcement learning (RL). 
To facilitate this, we curate a dataset specifically designed to improve LLM reasoning in the context of contextual integrity. 
This synthetic dataset enables us to efficiently explore and validate our RL approach.
We subsequently evaluate our model using existing contextual integrity benchmarks derived from human feedback, ensuring the validity and relevance of our framework.

\subsection{Chain of Thought Reasoning for Contextual Integrity}
\label{sec:cicot}

Prior research demonstrates that chain-of-thought (CoT) prompting significantly enhances the reasoning capabilities of autoregressive large language models \citep{wei2022chain, kojima2022large, yao2023tree}.
This technique requires models to generate intermediate reasoning steps prior to producing a final answer, which improves performance on complex reasoning tasks.
While initially applied to logical reasoning and computational problems, we extend this approach to contextual integrity reasoning.

Our method employs a structured prompt that directs the model to explicitly reason through contextual integrity considerations before addressing the user's request.
Specifically, we instruct the model to document its reasoning process within \texttt{<think>} and \texttt{</think>} tags, followed by the final response within \texttt{<answer>} and \texttt{</answer>} tags.
Within the reasoning phase, the model is guided to analyze the task and context, evaluate each personal attribute's relevance, and determine whether sharing it is necessary, helpful, optional, or inappropriate.
We term this approach \textbf{Contextual Integrity Chain of Thought (CI-CoT)} and present the prompt template in Figure \ref{fig:cicotprompt}.

\begin{figure}[ht]
\begin{tcolorbox}[colback=gray!10, colframe=gray!50, title=\textbf{Contextual Integrity Chain of Thought (CI-CoT)}, fonttitle=\bfseries, boxrule=0.5mm]
It is critical that you do not share inappropriate data.
Use the Contextual Integrity framework to guide your decisions.
Follow this structured process:
\begin{enumerate}
  \item \textbf{Reasoning Phase} (within \verb|<think>| and \verb|</think>| tags)
     \begin{itemize}
     \item Understand the task and context: Briefly describe what the user is trying to achieve and the broader context of the request.
     \item Evaluate each personal attribute:
       \begin{itemize}
       \item For each provided personal attribute, explain whether sharing it is necessary, helpful, optional, or inappropriate for fulfilling the user's request.
       \item Justify your decision based on the principle of Contextual Integrity: information should only be shared if it aligns with the norms of the context (including purpose, role, and transmission principles).
       \end{itemize}
     \end{itemize}
  \item \textbf{Response Phase} (within \verb|<answer>| and \verb|</answer>| tags)
     \begin{itemize}
     \item Output your final answer.
     \end{itemize}
\end{enumerate}

Always complete the Reasoning Phase carefully before generating the final response and use the appropriate tags.
\end{tcolorbox}
\caption{Prompt template for contextual integrity reasoning.}
\label{fig:cicotprompt}
\end{figure}
\subsection{Synthetic Dataset Curation}
\label{sec:dataset}
\begin{figure}[ht]
\centering
\begin{tikzpicture}[
    stage/.style   = {draw, rounded corners=2pt, minimum width=2.6cm,
                      minimum height=1.2cm, align=center, font=\small},
    arr/.style     = {thick, -{Triangle[length=3pt,width=5pt]}},
    node distance = 0.6cm
]

\node[stage, fill=gray!10] (seed)  {\textbf{Seed}\\\footnotesize scenario, domain,  transmission principle};
\node[stage, fill=gray!10, right=of seed] (vign) {\textbf{Vignette}\\\footnotesize actors + CI slots};
\node[stage, fill=gray!10, right=of vign] (json) {\textbf{Dataset item}\\\footnotesize \texttt{\{task, info, annotation\}}};

\draw[arr] (seed) -- (vign);
\draw[arr] (vign) -- (json);

\end{tikzpicture}
\vspace{-0.5em}
\caption{Three‑stage synthetic dataset curation pipeline used in Section~\ref{sec:dataset}.}
\label{fig:pipeline}
\end{figure}
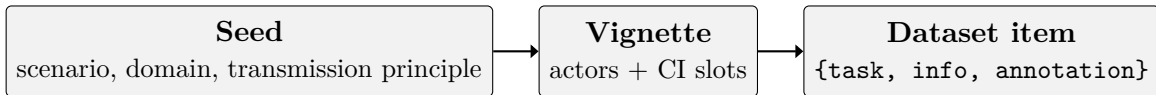
We build each dataset example in three stages (Figure\,\ref{fig:pipeline}).
In essence, each dataset example includes a clearly defined \textit{user task} $T$ that the AI assistant must complete, a set of \textit{required information} $A$ that is permissible for sharing to achieve this task, and a set of \textit{restricted information} $D$ that is inappropriate for disclosure within the given context.
Our goal is to train LLMs to distinguish between required and restricted information while completing the user task when these types of information are intermixed in the context without explicit labels.
We employ GPT-4 for the generation of our synthetic dataset.

\paragraph{Initial seeds.} The initial seeds vary the scenarios under which the AI assistant operates, such as sending emails or chat messages, to diversify the dataset contexts.
We also vary the task domains; following~\citep{cheng2024ci}, we include domains such as \emph{Hospitality}, \emph{Healthcare}, \emph{Entertainment}, \emph{Finance}, \emph{eCommerce}, \emph{Education}, \emph{Government}, \emph{Family}, \emph{Friends}, and \emph{Work}. 
Each seed includes a transmission principle that outlines the terms and conditions governing the interaction between the sender and recipient, aligning with relevant social norms and constructs~\citep{barth2006privacy}.
The transmission principle grounds the creation of the subsequent vignettes and final dataset examples.
In this work, we focus on three common transmission principles and examine their relevance to AI assistants: (1) \textit{Confidentiality}; information unrelated to the context (i.e., the task being performed, the sender, or the recipient contextual relationship) should not be shared; (2) \textit{Proportionality}; shared information should be proportionate to the task and not excessive; and (3) \textit{Consent}; information sharing depends on the awareness and consent of the data subject.
We then construct the final random seeds by sampling a scenario, a domain, and a transmission principle.

\paragraph{Vignettes.} The initial seeds are expanded by GPT-4 into vignettes that (1) state the user's task and (2) fill in the remaining contextual‑integrity (CI) fields (sender, recipient, subject). 
For each vignette two disjoint sets of information
categories are also generated by GPT-4 — those that the task \textit{requires} and those that the principle \textit{restricts}, for example:
\begin{lstlisting}[language=json, frame=single, basicstyle=\ttfamily\small]
"information_type": {
  "required": ["name", "event date", "number of guests"],
  "restricted": ["personal financial details", "medical history"]}
\end{lstlisting}
Each vignette is automatically generated using a prompt that includes concise explanations of the governing transmission principles to ground the context.
For each initial seed, we generate 3–5 vignettes, resulting in a total of 795 vignettes. 
Examples of the complete vignettes and the prompt used to generate them are provided in Appendix~\ref{app:dataset_example}.

\paragraph{Final dataset examples.} The final step is to transform the vignettes into examples presented in a more natural format.
We feed the vignettes into GPT-4 and prompt it to populate the specific values for the user’s query directed to the agent, the names and roles of senders and recipients, and each information type.
To induce diversity, we also prompt GPT-4 to generate natural conversations that incorporate the information specified in the vignettes.
The information is organized as key-value pairs.
Each key-value pair corresponds to an information item specified in the vignette.
The LLM is also prompted to generate neutral names for the keys (to avoid introducing cues about the flow annotation) and to produce an annotation indicating whether each key–value pair belongs to the \textit{required} or \textit{restricted} category. 
Within the annotation, specific \textbf{\texttt{keywords}} are extracted from the required and restricted values to facilitate a string-matching mechanism for a rule-based reward function that scores the presence of required and restricted values.
Examples of the final dataset items are provided in Appendix~\ref{app:dataset_example} along with the generation prompt.
\subsection{Contextual Integrity Optimized RL Training}

We further argue that the prompt-based reasoning ability can be enhanced through RL.

We employ GRPO with a rule-based reward function to enhance reasoning aligned with contextual integrity.
Our reward function comprises two parts, similar to \cite{guo2025deepseek}: a scoring mechanism for contextual integrity, and a formatting criterion that assesses whether responses adhere to a specified structured format. 
Each response must include an explicit reasoning component enclosed in \texttt{<think>} and \texttt{</think>} tags, and a task completion enclosed in \texttt{<answer>} and \texttt{</answer>} tags.

For the contextual integrity scoring mechanism, we extract the model’s attempt to complete the user task from within the \texttt{<answer>} and \texttt{</answer>} tags, assuming the required format is followed.
We define a reward function \( R \) as follows. Let \( A \) denote the set of all required keywords, and \( D \) denote the set of all restricted keywords. Let \( A_{\text{present}} \subseteq A \) denote the subset of required keywords that appear in the user task completion, and similarly \( D_{\text{present}} \subseteq D \) denote the restricted keywords that appear in the user task completion.

The reward function is defined as:

\begin{equation}
R =
\begin{cases}
    -1 & \text{if the response violates the required format} \\
    \frac{|A_{\text{present}}|}{|A|} - \frac{|D_{\text{present}}|}{|D|} & \text{otherwise}
\end{cases} \; ,
\label{eq:reward}
\end{equation}

where a format violation occurs if the response is missing valid \texttt{<think>} or \texttt{<answer>} tags.

We call this approach \textbf{Contextual Integrity Reinforcement Learning (CI-RL)}.

\section{Experiments}
\label{sec:experiments}
Having detailed our data-generation pipeline and training procedure, we now turn to an empirical evaluation. 
We next outline the experimental setup.
\paragraph{Dataset.} We separate the dataset\footnote{Synthetic dataset: https://huggingface.co/datasets/huseyinatahaninan/ContextualIntegritySyntheticDataset} generated in Section~\ref{sec:dataset} into disjoint training, evaluation, and test subsets, containing 590, 66, and 73 examples, respectively.
We provide a training sample in Appendix~\ref{app:example_prompt}.
During training, the models are periodically evaluated, and the best checkpoint is selected based on the highest validation score. 
This checkpoint is subsequently evaluated on the test set, and its performance is reported as the main result of this section.
\paragraph{Models.} We select a series of models along two dimensions: (1) \textit{Model size}; experimenting with Qwen2.5-1.5B-Instruct, Qwen2.5-3B-Instruct, Qwen2.5-7B-Instruct\footnote{Checkpoint trained from Qwen2.5-7B-Instruct: \url{https://huggingface.co/huseyinatahaninan/Qwen2.5-7B-Instruct-CI}}, and Qwen2.5-14B-Instruct \citep{yang2024qwen2} and (2) \textit{Model family}; experimenting with Llama-3.1-8B-Instruct \citep{llama3} and Mistral-7B-Instruct-v0.3 \citep{jiang2024mistral}.
\paragraph{Training details.}
We base our training method on the VERL framework \citep{sheng2024hybridflow}, adapting it to our tasks\footnote{Code at GitHub: \url{https://github.com/EricGLan/CI-RL}}.
The hyperparameters, dataset statistics, and computer resources are outlined in Appendix \ref{sec:append_Hyperparameters}.
\paragraph{Metrics.}
Let \( A_i \) denote the set of all required keywords, and \( D_i \) denote the set of all restricted keywords for a test example $s_i$ for $i \in \{1, 2, \ldots, N\}$. 
Let $g_i$ denote the corresponding model generation for the user task completion in $s_i$.
We write $\mathbbm{1}[\cdot]$ for the indicator function.
We consider the following metrics in our tasks:
\begin{enumerate}[leftmargin=*]
    \item Integrity ($\mathcal{I}$): Excludes all restricted information in the task, averaged over the test examples. Formally, $\mathcal{P} = \frac{1}{N} \sum_{i=1}^{N} \mathbbm{1}[\{d\in D_i| \ g_i \ \textnormal{does not contain} \ d \} = D_i]$.
    \item Utility ($\mathcal{U}$): Includes all required information to complete the task, averaged over the test examples. Formally, $\mathcal{U} = \frac{1}{N} \sum_{i=1}^{N} \mathbbm{1}[\{a\in A_i| \ g_i \ \textnormal{contains} \ a \} = A_i]$.
    \item Complete ($\mathcal{C}$): Includes all required information and excludes all restricted information to complete the task, averaged over the test examples. Formally,
    $\mathcal{C} = \frac{1}{N} \sum_{i=1}^{N} \mathbbm{1}[\{a\in A_i| \ g_i \ \textnormal{contains} \ a \} = A_i \ \& \ \{d\in D_i| \ g_i \ \textnormal{does not contain} \ d \} = D_i]$.
\end{enumerate}

\subsection{Results}
\begin{table}[ht]
    \centering
    \caption{Main results across models. 
    We observe that \textbf{CI-RL} consistently improves both {Integrity} ($\mathcal{I}$) and {Complete} ($\mathcal{C}$) metrics for all models while maintaining comparable or improved Utility ($\mathcal{U}$).}
    \begin{tabular}{lc|cc|cc|cc}
    \toprule
    \multicolumn{2}{l}{Model} & \multicolumn{2}{|c}{$\mathcal{I}$ (in \%) $\uparrow$} &  \multicolumn{2}{c}{$\mathcal{U}$ (in \%) $\uparrow$} & \multicolumn{2}{c}{$\mathcal{C}$ (in \%) $\uparrow$} \\
    \midrule
    Qwen2.5-1.5B-IT & \color{PineGreen}{+ \textbf{CI-RL}} & 37.5 & \color{PineGreen}{\textbf{59.4}} & 35.9 & \color{PineGreen}{\textbf{43.7}} & 4.7 & \color{PineGreen}{\textbf{26.6}} \\
    Qwen2.5-3B-IT & \color{PineGreen}{+ \textbf{CI-RL}} & 31.2 & \color{PineGreen}{\textbf{57.8}} & 53.1 & \color{PineGreen}{\textbf{51.6}}  & 12.5 & \color{PineGreen}{\textbf{28.1}} \\
    Qwen2.5-7B-IT & \color{PineGreen}{+ \textbf{CI-RL}} & 46.9 & \color{PineGreen}{\textbf{75.0}} & 62.5 & \color{PineGreen}{\textbf{67.2}}  & 29.7 & \color{PineGreen}{\textbf{48.4}} \\
    Mistral-7B-IT & \color{PineGreen}{+ \textbf{CI-RL}} & 38.8 & \color{PineGreen}{\textbf{89.1}}  & 67.3 & \color{PineGreen}{\textbf{82.8}} & 24.5 & \color{PineGreen}{\textbf{73.4}} \\
    Llama-3.1-8B-IT & \color{PineGreen}{+ \textbf{CI-RL}} & 61.9 & \color{PineGreen}{\textbf{79.7}} & 64.3 & \color{PineGreen}{\textbf{79.7}} & 38.1 & \color{PineGreen}{\textbf{62.5}} \\
    Qwen2.5-14B-IT & \color{PineGreen}{+ \textbf{CI-RL}} & 51.6 & \color{PineGreen}{\textbf{78.1}} & 67.2 & \color{PineGreen}{\textbf{64.1}} & 37.5 & \color{PineGreen}{\textbf{50.0}} \\
    \bottomrule
\end{tabular}

    \label{tab:main_results}
\end{table}

We present the results of our experiments in Table~\ref{tab:main_results}, which demonstrates the following key findings:
\begin{itemize}[leftmargin=*]
    \item \textbf{CI-RL consistently improves Integrity and Complete metrics across all models.} Across all model sizes and families, the application of CI-RL increases both Integrity and Complete metrics compared to their baseline models. Notably, these improvements are achieved while maintaining strong utility across models, demonstrating that our approach preserves required information sharing for the tasks. See Appendix~\ref{app:output_synthetic} for an illustrative generation trajectory during training.
    \item \textbf{Larger models achieve higher absolute scores.} The larger models achieve higher overall scores, suggesting that scaling up model size enhances reasoning capability, which in turn contributes to improved integrity adherence, as expected.
    \item \textbf{CI-RL enables smaller models to outperform larger baseline models.} For example, Qwen2.5-7B-Instruct after CI-RL achieves a Integrity score of 75.0\% and a Complete score of 48.4\%, both surpassing the baseline Qwen2.5-14B-Instruct (Integrity: 51.6\%, Complete: 37.5\%). This highlights the effectiveness of reinforcement learning in closing, and even reversing, the performance gap between smaller and larger models for contextual integrity.
\end{itemize}

\subsection{Ablation Studies}

\paragraph{LLMs vs LRMs.} Large reasoning models (LRMs) are language models that are explicitly encouraged or optimized to perform multi-step reasoning and structured problem solving.
Unlike standard large language models (LLMs), which may rely on surface-level statistical patterns, LRMs are designed to articulate intermediate reasoning steps, improving interpretability and control. 
For tasks requiring nuanced judgment, such as determining whether information flows align with contextual integrity principles, LRMs offer a promising approach by enabling the model to reason explicitly about contextual integrity within the context rather than depending solely on implicit knowledge.
Motivated by these insights, we compare Llama-3.1-8B-Instruct with DeepSeek-R1-Distill-Llama-8B \citep{guo2025deepseek}, which extends Llama-3.1-8B and Qwen2.5-14B-Instruct with DeepSeek-R1-Distill-Qwen-14B, which extends Qwen2.5-14B and report the results in Table~\ref{tab:llm_lrm}.

\begin{table}[ht]
    \centering
    \caption{Comparison of LLMs and LRMs. Our evaluation reveals that LRMs fall short of LLMs in overall performance in this task.}

\begin{tabular}{lc|cc|cc|cc}
    \toprule
    \multicolumn{2}{l}{Model} & \multicolumn{2}{|c}{$\mathcal{I}$ (in \%) $\uparrow$} & \multicolumn{2}{c}{$\mathcal{U}$ (in \%) $\uparrow$} & \multicolumn{2}{c}{$\mathcal{C}$ (in \%) $\uparrow$} \\
    \midrule
    Llama-3.1-8B-Instruct & \color{PineGreen}{+ \textbf{CI-RL}} & 61.9 & \color{PineGreen}{\textbf{79.7}} & 64.3 & \color{PineGreen}{\textbf{79.7}} & 38.1 & \color{PineGreen}{\textbf{62.5}} \\
    DeepSeek-R1-Distill-Llama-8B & \color{PineGreen}{+ \textbf{CI-RL}} & 35.9 & \color{PineGreen}{\textbf{68.7}} & 57.8 & \color{PineGreen}{\textbf{65.6}} & 20.3 & \color{PineGreen}{\textbf{45.3}} \\
    Qwen2.5-14B-Instruct & \color{PineGreen}{+ \textbf{CI-RL}} & 51.6 & \color{PineGreen}{\textbf{78.1}} & 67.2 & \color{PineGreen}{\textbf{64.1}} & 37.5 & \color{PineGreen}{\textbf{50.0}} \\
    DeepSeek-R1-Distill-Qwen-14B & \color{PineGreen}{+ \textbf{CI-RL}} & 29.7 & \color{PineGreen}{\textbf{75.0}} & 73.4 & \color{PineGreen}{\textbf{60.1}} & 18.7 & \color{PineGreen}{\textbf{46.9}} \\
    \bottomrule
\end{tabular}
    \label{tab:llm_lrm}
\end{table}

{The results demonstrate that instruction-tuned LLMs achieve substantially higher integrity, utility, and task completion scores compared to LRMs after CI-RL training. }
However, we do not believe this gap is inherent to the LRM paradigm. 
We hypothesize that the performance difference may stem from the fact that the distilled models have been primarily optimized for scientific and code-related domains, at the expense of broader domain coverage. 
As a result, their performance on CI tasks, which require diverse, real-world understanding, lags behind that of instruction-tuned LLMs.
\paragraph{Integrity-utility trade-off via reward function design.}
By adjusting the weighting of required and restricted keywords in the reward function defined in Equation~\eqref{eq:reward}, we can influence the model's prioritization of integrity versus utility. 
We present the results of this ablation in Appendix~\ref{sec:supplementary_results}.

\section{Evaluation -- PrivacyLens}
\label{sec:PrivacyLens}
Having introduced a promising training pipeline, we still need to verify that the results translate to a real world setting.
We employ PrivacyLens \citep{shao2024privacylens} as a comprehensive benchmark to evaluate our method's performance in improving contextual integrity.
The benchmark provides a standardized framework for assessing the contextual integrity in large language model outputs.
Below we summarize the primary evaluation metrics utilized in our analysis:
\paragraph{Helpfulness.}
To quantify model utility, \citet{shao2024privacylens} employ an LLM judge to evaluate the helpfulness of the final action on a scale from 0 (poor) to 3 (excellent).
This metric assesses whether the model's final action fulfills the user's intention. 
\paragraph{Leakage Rate.}
To measure privacy leakage, \citet{shao2024privacylens} implement a few-shot classifier to detect whether the final action contains any sensitive attributes.
The leakage rate (LR) is calculated as the percentage of responses containing disclosure of sensitive information.
\paragraph{Adjusted Leakage Rate.}
To compensate for the safety-helpfulness trade-off (as models that refuse to respond are technically safe but not helpful), \citet{shao2024privacylens} propose the adjusted leakage rate (ALR).
This metric is defined as the leakage rate calculated exclusively for helpful responses (those receiving a helpfulness score of 2 or 3).
ALR provides an assessment of how models balance privacy protection with information provision in scenarios where responses are actually useful to users.

\subsection{Results}

In all our experiments, we employ GPT-4o (version 2024-11-20) with a temperature of 0 as the judge model for PrivacyLens evaluations.
To avoid the bias of judges towards their own generations \cite{panickssery2024llm}, we do not present any results for OpenAI models.
All quantitative results are summarized in Table \ref{tab:privacylens}.
\paragraph{Chain of Thought Reasoning for Contextual Integrity.}
Our investigation begins with examining whether explicit reasoning about contextual integrity enhances LLM performance across both safety and helpfulness dimensions.
To address this question, we evaluate several model categories, including frontier models such as Claude 3.7 Sonnet (S) \citep{claude2025} and Gemini 2.5 Flash \citep{geminiflash2025}.
Additionally, we assess large reasoning models (LRMs), specifically Claude 3.7 Sonnet Thinking (S-T) and Gemini 2.5 Pro \citep{geminipro2025}.
Table \ref{tab:privacylens} presents a comparison of these models alongside the open-weight models described in Section \ref{sec:experiments}.
The results consistently demonstrate that the CI-CoT prompt yields improvements in both the leakage rate (LR) and the adjusted leakage rate (ALR), with the latter metric accounting for the helpfulness-safety trade-off.
The CI-CoT approach makes models more conservative regarding sensitive information disclosure, resulting in reduced helpfulness scores.
Notably, even when accounting for this trade-off through the adjusted leakage rate, our results still demonstrate a positive reduction in information leakage.

\begin{table}[ht]
    \centering
    \small
    \caption{
        PrivacyLens Results.
        We compare the performance of different models on the PrivacyLens benchmark.
        The leakage rate (LR) and adjusted leakage rate (ALR) are both lower when reasoning about CI using our CI-CoT prompt.
    }
    \begin{tabular}{ll|lll}
    \toprule
    \multicolumn{2}{l|}{Model} & \multicolumn{1}{c}{LR (in \%) $\downarrow$} & \multicolumn{1}{c}{ALR (in \%) $\downarrow$} & \multicolumn{1}{c}{Helpful [0-3] $\uparrow$} \\
    \midrule
    \rowcolor{gray!20} \multicolumn{5}{c}{\textit{Baseline LLMs}} \\
    \midrule
    Claude 3.7 S & \color{Violet}{+ \textbf{CI-CoT}} & 30.4 \ \color{Violet}{\textbf{23.1}} & 35.9 \ \color{Violet}{\textbf{25.4}} & 2.49 \ \color{Violet}{\textbf{2.69}} \\
    Gemini 2.5 F & \color{Violet}{+ \textbf{CI-CoT}} & 29.0 \ \color{Violet}{\textbf{19.7}} & 30.8 \ \color{Violet}{\textbf{24.0}} & 2.75 \ \color{Violet}{\textbf{2.31}} \\
    \midrule
    \rowcolor{gray!20} \multicolumn{5}{c}{\textit{Baseline LRMs}} \\
    \midrule
    Claude 3.7 S-T & \color{Violet}{+ \textbf{CI-CoT}} & 32.0 \ \color{Violet}{\textbf{20.1}} & 34.6 \ \color{Violet}{\textbf{22.6}} & 2.75 \ \color{Violet}{\textbf{2.63}} \\
    Gemini 2.5 Pro & \color{Violet}{+ \textbf{CI-CoT}} & 37.3 \ \color{Violet}{\textbf{25.3}} & 38.2 \ \color{Violet}{\textbf{26.9}} & 2.84 \ \color{Violet}{\textbf{2.72}} \\
    \midrule
    \rowcolor{gray!20} \multicolumn{5}{c}{\textit{Open Weights}} \\
    \midrule
    Mistral-7B-IT & \color{Violet}{+ \textbf{CI-CoT}} \ \color{PineGreen}{+ \textbf{CI-RL}} & 47.9 \ \color{Violet}{\textbf{28.8}} \ \color{PineGreen}{\textbf{31.2}} & 52.1 \ \color{Violet}{\textbf{46.6}} \ \color{PineGreen}{\textbf{29.6}} & 1.78 \ \color{Violet}{\textbf{1.17}} \ \color{PineGreen}{\textbf{1.84}} \\
    Qwen-7B-IT & \color{Violet}{+ \textbf{CI-CoT}} \ \color{PineGreen}{+ \textbf{CI-RL}} & 50.3 \ \color{Violet}{\textbf{44.8}} \ \color{PineGreen}{\textbf{33.7}} & 52.4 \ \color{Violet}{\textbf{45.7}} \ \color{PineGreen}{\textbf{33.9}} & 1.99 \ \color{Violet}{\textbf{2.13}} \ \color{PineGreen}{\textbf{2.08}} \\
    Llama-8B-IT & \color{Violet}{+ \textbf{CI-CoT}} \ \color{PineGreen}{+ \textbf{CI-RL}} & 18.2 \ \color{Violet}{\textbf{21.3}} \ \color{PineGreen}{\textbf{18.5}} & 38.9 \ \color{Violet}{\textbf{31.5}} \ \color{PineGreen}{\textbf{29.4}} & 1.05 \ \color{Violet}{\textbf{1.29}} \ \color{PineGreen}{\textbf{1.18}} \\
    Qwen-14B-IT & \color{Violet}{+ \textbf{CI-CoT}} \ \color{PineGreen}{+ \textbf{CI-RL}} & 52.9 \ \color{Violet}{\textbf{42.8}} \ \color{PineGreen}{\textbf{33.9}} & 51.2 \ \color{Violet}{\textbf{44.4}} \ \color{PineGreen}{\textbf{34.4}} & 2.37 \ \color{Violet}{\textbf{2.27}} \ \color{PineGreen}{\textbf{2.30}} \\
    \bottomrule
\end{tabular}

    \label{tab:privacylens}
\end{table}

\paragraph{Reinforcement Learning for Contextual Integrity.}
Building on our findings, we next explore whether reinforcement learning can further enhance LLM performance regarding both safety and helpfulness metrics.
As demonstrated in the previous section, while explicit reasoning about contextual integrity improves safety metrics, it often comes at the expense of reduced helpfulness.
To address this trade-off, we implement the reward function defined in Equation \eqref{eq:reward}, which balances penalties for inappropriate information disclosure with rewards for appropriate disclosure.
This approach enables models to learn more nuanced information-sharing policies aligned with contextual integrity principles. We show qualitative examples in Appendix~\ref{app:privacylens_examples} for Qwen2.5-7B-IT CI-RL on PrivacyLens. As can be observed, this benchmark has significant differences than our synthetic dataset. The context provided to the LLM is considerably longer and the benchmark is centered around detailed tool use. Nevertheless, CI-RL shows improvement compared to just using CI-CoT. 

In summary, our evaluation shows:
\begin{itemize}[leftmargin=*]
    \item CI-CoT serves as an effective mechanism for reducing leakage rate (LR) and adjusted leakage rate (ALR), though with a modest decrease in overall helpfulness.
    \item CI-RL further optimizes this balance by achieving even lower leakage rates and adjusted leakage rates while preserving or enhancing helpfulness metrics.
    \item Across our experiments, frontier models consistently demonstrate lower leakage rates and higher helpfulness compared to their significantly smaller open-weight counterparts.
\end{itemize}

\section{Discussions}
\label{sec:discussions}

\subsection{Summary}

In this chapter, we improve the ability of LLMs to reason about contextual integrity (CI), a framework that governs the appropriateness of information flow. 
We first demonstrate that prompting LLMs to engage in explicit CI reasoning can reduce inappropriate information leakage. 
Building on this, we introduce an RL framework that further enhances CI-aligned reasoning by optimizing models on a synthetic dataset spanning diverse scenarios. 
The experiments show that our approach significantly reduces inappropriate information leakage while preserving task utility, with consistent improvements observed across various model families. 
We further demonstrate remarkable improvements on CI benchmarks such as PrivacyLens. 
These findings highlight the promise of combining structured reasoning and RL to develop safer and more context-aware AI agent systems.

\subsection{Limitations and Future Work}

Despite our promising results, several limitations remain and suggest future directions:
\paragraph{Human-Annotated CI Data.} 
While high-quality contextual integrity (CI) data ideally relies on nuanced human annotation, such data remains scarce and expensive to collect at scale. As a result, we used a synthetic dataset to demonstrate the feasibility of our RL approach. Future work should incorporate human-annotated CI datasets to further validate and refine our findings.
\paragraph{Scaling and Model Generalization.}
Our results show that larger models consistently outperform smaller ones on CI tasks and generalize well to external benchmarks like PrivacyLens. This suggests that scaling plays an important role in enabling nuanced CI reasoning. Future work should explore applying our method to models larger than 14B. Furthermore, more empirical studies on large reasoning models (LRMs) that are explicitly trained for multi-domain reasoning, including visual reasoning~\citep{hou2025tdbench, shi2025towards, zeng2025futuresightdrive}, would provide a better understanding of the relative strengths and limitations of LLMs versus LRMs in CI tasks.
\paragraph{Learning and Prompting-Based Reward Supervision.}
Our keyword-based reward is simple and auditable but coarse. With abundant, high-quality data, training a CI-specific reward model or using rubric-guided LLM judges could better capture context-sensitive norms and raise recall.
\paragraph{Comparison with previous methods.}
Our approach is largely orthogonal to and potentially complementary with prior privacy agents and guardrails. For instance, \emph{AirGapAgent}~\citep{Bagdasarian2024AirGapAgent} enforces privacy via inference-time gating: a judge LLM decides which inputs are strictly necessary for the task. By contrast, \emph{CI-CoT} prompts the model to reason about contextual integrity, and \emph{CI-RL} trains the policy itself to internalize these norms during task completion. System defenses (firewalls, minimizers, trajectory filters) act \emph{outside} the model, whereas CI-RL shifts behavior \emph{inside} the model. As future work, we will run head-to-head and hybrid evaluations, pairing CI-RL with AirGapAgent-style gatekeepers, reporting integrity/utility trade-offs, false positive/negative rates, and cost/latency on shared benchmarks (PrivacyLens, ConfAIde).
\paragraph{Evolving norms and user customization.}
Privacy norms are context-dependent and drift over time; users and organizations also need custom policies. Our CI-CoT component is naturally adaptable at inference (via prompt/policy conditioning), while CI-RL’s rule-based reward is modular and can be updated by swapping restricted/required lists or adding user/tenant-specific constraints. Future work can support policy injection and versioning, periodic rule refresh (e.g., active learning from new decisions), and lightweight policy extractors that map governance documents into CI rules to enable per-user/per-organization customization with minimal retraining.
\paragraph{RL vs. SFT for CI.}
In agentic scenarios where the user task is open-ended and information flows are annotated, RL offers a natural fit. 
It allows the model to generate full task completions and be rewarded directly based on the presence or absence of specific information types in its output.
Moreover, RL is often more data-efficient than supervised fine-tuning (SFT); recent work has shown that RL can yield improvements with as little as a single training example~\citep{wang2025reinforcementlearningreasoninglarge}. 
Nevertheless, comparing SFT and RL-based approaches on CI frameworks remains an important direction.
\paragraph{Unstructured and Retrieval-Augmented Contexts.}
We constructed a relatively simple training dataset with semi-structured input. 
Yet our method yields consistent gains on more natural, free-form chats with conversation history (PrivacyLens) and shows the same trend on an external replication with ConfAIde (single-model slice; Appendix~\ref{sec:supplementary_results}).
Extending the training and CI reasoning to more complex settings would further validate the robustness of our approach.

\DefineShortVerb{\|}

\endgroup
\ProvidesFile{ch-conclusion.tex}[2026-04-07 conclusion chapter]

\chapter{Conclusion and Future Directions}
\label{ch:conclusion}

\section{Conclusion}

This dissertation studied reinforcement learning as a unifying framework for advancing intelligent systems along two complementary dimensions: scalability and trustworthiness. The central premise has been that reinforcement learning is not only a tool for maximizing task reward, but also a general methodology for optimizing policies under real-world constraints. In distributed environments, these constraints arise from limited communication, heterogeneous computation, and the need to learn from decentralized data. In language-based intelligent systems, they arise from the need to align behavior with human preferences and to respect contextual norms governing appropriate information disclosure.

The first part of this dissertation addressed the scalability of reinforcement learning in federated settings. In Chapter~\ref{ch:sfedrl}, we proposed \emph{FedNPG-ADMM}, a communication-efficient framework for synchronous federated reinforcement learning. By integrating the alternating direction method of multipliers with natural policy gradient optimization, FedNPG-ADMM reduces the per-iteration communication complexity from $O(d^2)$ to $O(d)$ while preserving the stationary convergence guarantees of standard federated natural policy gradient methods. The results show that second-order policy optimization can be made significantly more practical in large-scale federated systems without sacrificing theoretical guarantees or empirical reward performance.

In Chapter~\ref{ch:afedrl}, we turned from communication efficiency to time efficiency and studied \emph{asynchronous federated reinforcement learning}. We introduced \emph{AFedPG}, a policy-gradient framework tailored to heterogeneous federated systems in which agents operate at different speeds and local updates may be delayed. To address the challenge of stale policies in the asynchronous setting, AFedPG employs a delay-adaptive lookahead technique that enables stable updates and yields both theoretical and practical benefits. In particular, the method achieves linear speedup in sample complexity with respect to the number of agents and improves global time complexity relative to synchronous federated policy-gradient methods. Together, Chapters~\ref{ch:sfedrl} and~\ref{ch:afedrl} demonstrate that reinforcement learning can be adapted to the systems constraints of modern distributed environments through principled optimization design.

The second part of this dissertation addressed the trustworthiness of reinforcement learning for language-based intelligent systems. In Chapter~\ref{ch:mappo}, we introduced \emph{Maximum a Posteriori Preference Optimization (MaPPO)}, a principled preference-optimization framework that incorporates prior reward knowledge into the training objective. By extending the conventional maximum-likelihood view of preference optimization to a Maximum a Posteriori formulation, MaPPO mitigates confidence degeneration, yields a more calibrated training signal, and remains compatible with both offline and online preference-optimization settings. Empirically, MaPPO consistently improves alignment performance across multiple model families, model sizes, and evaluation benchmarks, while preserving computational efficiency and maintaining strong performance on academic benchmarks beyond alignment-specific evaluations.

In Chapter~\ref{ch:contextual-integrity}, we studied \emph{contextual integrity} as a concrete safety property for large language models and autonomous agents. We showed that explicit reasoning about contextual integrity can improve a model's ability to determine what information is appropriate to disclose in a given context. Building on this insight, we proposed a reinforcement-learning-based post-training framework that combines explicit reasoning with contextual-integrity-aware reward design. Using a synthetic dataset spanning diverse domains and disclosure norms, and evaluating on the human-annotated PrivacyLens benchmark, we demonstrated that reinforcement learning can substantially reduce inappropriate information disclosure while preserving task utility. This result highlights the broader role of reinforcement learning in shaping not only the capability of language models, but also their safety-sensitive behavior.

Taken together, the contributions of this dissertation support a common conclusion: reinforcement learning provides a unified perspective for optimizing policies in intelligent systems under both systems-level and behavior-level constraints. In federated reinforcement learning, the central challenge is to scale policy optimization across distributed agents while respecting communication and computation limits. In language-model post-training, the central challenge is to shape policy behavior so that it is aligned, privacy-aware, and contextually appropriate. While these domains differ in application and methodology, they are linked by the same core idea: a policy must be optimized not only for effectiveness, but also for deployability in the real world.

\section{Future Directions}

The results of this dissertation also suggest several promising directions for future research. These directions arise both from the limitations of the individual chapters and from broader questions at the intersection of scalable and trustworthy reinforcement learning.

\subsection{Future Directions for Scalable Reinforcement Learning}

A first important direction is to further strengthen the practicality of federated reinforcement learning in more realistic distributed settings. Although Chapter~\ref{ch:sfedrl} substantially reduces the communication cost of second-order federated policy optimization, several challenges remain open. One such challenge is \emph{partial participation}, where only a subset of agents is available or selected in each communication round. Extending the current framework to explicitly model partial participation could further reduce communication cost and improve robustness in real-world deployments with intermittent connectivity. A related direction is to scale the evaluation to substantially larger numbers of federated agents and more realistic environments, thereby better characterizing the empirical behavior of communication-efficient second-order methods at scale.

A second direction concerns the relationship between communication efficiency and privacy. The federated setting already limits raw data sharing, but more explicit privacy guarantees remain largely unexplored in the present work. In particular, connecting communication-efficient policy optimization with differential privacy, secure aggregation, or other privacy-preserving mechanisms would be valuable for applications where both efficient learning and formal privacy guarantees are required.

A third direction is to broaden the asynchronous federated reinforcement learning framework. While Chapter~\ref{ch:afedrl} shows that AFedPG improves both sample efficiency and global time complexity, the analysis and experiments focus on first-order policy-gradient methods in discounted-reward settings. Extending these results to \emph{asynchronous second-order policy optimization}, such as natural policy gradient or trust-region methods, would be a natural next step. Such an extension would combine the communication and curvature advantages explored in Chapter~\ref{ch:sfedrl} with the heterogeneity-aware asynchronous design of Chapter~\ref{ch:afedrl}. In addition, extending the framework from discounted-reward Markov decision processes to \emph{average-reward settings} would broaden its theoretical and practical applicability.

A fourth direction is to address \emph{robustness in asynchronous federated environments}. Although AFedPG is designed to handle stragglers and stale updates, the presence of adversarial or malicious workers remains an open problem. Developing asynchronous federated reinforcement learning algorithms that remain robust to poisoned updates, unreliable participants, or strategic agents would be particularly important for security-sensitive distributed learning systems, including edge device deployment~\cite{lan2023fl} and digital twin~\cite{ren2025digital}.

\subsection{Future Directions for Trustworthy Reinforcement Learning}

On the trustworthiness side, an important direction is to further improve preference optimization for large language models. Chapter~\ref{ch:mappo} showed that incorporating prior reward knowledge into preference optimization leads to more calibrated and effective training. However, several open questions remain. First, the current evaluation suggests that larger models consistently benefit from the MaPPO framework, motivating future work on scaling the method to models larger than those studied in this dissertation. Second, the design of the prior knowledge function still relies on domain-informed choices. A promising direction is to study more adaptive, data-driven, or theoretically grounded prior-function designs that can better reflect specific application domains or more nuanced human preferences. It is also worth to explore the design pattern for diffusion-based models~\cite{zhang2024bridging,zhouneural,li2026comprehensive}, and multimodals~\cite{xu2024fakeshield,yang2025magic,zhou2025opening,chen2025think,chen2025visrl,li2025information}.

Another natural direction is to improve the evaluation of aligned behavior, e.g., with rubric rewards~\cite{lan2026alternating}. While the current results demonstrate gains on standard automatic benchmarks, future work should incorporate more extensive \emph{human evaluation} to assess whether the improved calibration and benchmark performance translate into better real-world alignment. This is particularly important because alignment is inherently tied to human judgment, and automatic metrics may fail to capture subtle failures or trade-offs.

A broader direction is to study the interaction between \emph{preference optimization and RL-based post-training methods}. The current dissertation treats preference optimization and contextual safety as two related but distinct threads. Future work could explore tighter integration between prior-informed preference optimization and online reinforcement learning methods, including settings where reward models, preference data, and policy rollouts are jointly refined.

\subsection{Future Directions for Contextual Safety and Privacy}

Chapter~\ref{ch:contextual-integrity} suggests several concrete directions for improving contextual safety in language models. The first is the need for richer and more diverse \emph{human-annotated contextual integrity data}. In this dissertation, synthetic data provided a practical and scalable way to train and validate the method, and the transfer to PrivacyLens demonstrates that the learned behavior is meaningful. Nevertheless, contextual integrity is inherently nuanced and socially situated. Future work would benefit from larger human-annotated datasets or sythetic data using distillation \cite{zhang2025find} that reflect a broader range of real-world information-sharing norms and tool-use ability \cite{jiang2026scribe, xu2025learning}.

A second direction is \emph{scaling and model generalization}. The experiments indicate that larger models generally perform better on contextual integrity tasks and transfer more effectively to external benchmarks. Future studies should examine how contextual-integrity reasoning scales to larger instruction-following models and large reasoning models, and whether models specifically optimized for multi-step reasoning exhibit qualitatively different strengths and failure modes in privacy-sensitive tasks. It is also worthy to explore the efficient reasoning, e.g., distillation \cite{jiang2025drp, li2025frequency}, pruning \cite{li2026sepprune} and memorization-constrained reasoning \cite{jiang-ferraro-2026-beyond} under the concept of contextual integrity.

A third direction is to improve the reward supervision used for contextual safety. The keyword-based reward employed in this dissertation is simple, interpretable, and auditable, which makes it attractive for an initial study. However, it is also coarse. Future work could develop \emph{contextual-integrity-specific reward models}, rubric-guided LLM judges, incorrect information detection~\cite{zhang2026bimind}, or hybrid supervision methods that better capture subtle context-dependent disclosure norms. Such approaches may provide higher recall and more nuanced training signals than keyword-based matching alone.

A fourth direction is to integrate policy-level learning with \emph{system-level privacy defenses}. The contextual-integrity framework developed in this dissertation is largely orthogonal to inference-time privacy agents, data minimizers, firewalls, and trajectory filters. A promising research avenue is to combine these approaches: system-level defenses can constrain the information available to the model, while reinforcement learning can shape how the model internally reasons about context and disclosure, or different domain applications, e.g., UAV~\cite{zeng2026priordrive,zeng2025futuresightdrive,yang2025joint,huang2025erasure}, 5G networks~\cite{kouchaki2024enhanced}, decentralized web~\cite{zhang2023kadabra, zhang2025honeybee}, smart contracts~\cite{hu2026zero}, and malware detection~\cite{hu2025flowmaltrans}. Their combination may yield more robust privacy-preserving assistants than either approach alone.

\subsection{Toward a Unified Research Agenda}

Beyond the chapter-specific future directions, the broadest opportunity is to more tightly connect the two main themes of this dissertation. Scalable reinforcement learning and trustworthy reinforcement learning are often treated as separate research programs, but future intelligent systems will increasingly require both simultaneously. Distributed and federated systems are likely to support language-based agents that must be aligned with human values, while trustworthy assistants may themselves operate across decentralized devices, tools, and data sources.

This suggests a long-term agenda centered on \emph{scalable and trustworthy policy optimization}. Examples include federated post-training of language models, privacy-aware distributed preference optimization, asynchronous reinforcement learning for safety-critical autonomous agents, and reinforcement learning frameworks that jointly optimize system efficiency and behavioral guarantees. The central message of this dissertation is that reinforcement learning provides a common conceptual and mathematical language for all of these problems. Advancing this agenda will require bringing together insights from optimization, distributed systems, human feedback, privacy, and AI safety.

\section{Closing Remarks}

As intelligent systems become increasingly capable, interactive, and autonomous, the demands placed on their learning algorithms will continue to grow. They must be trainable across distributed and resource-constrained infrastructures, yet also behave in ways that are aligned with human preferences and societal norms. This dissertation has argued that reinforcement learning is uniquely well positioned to address both requirements.

The contributions presented here represent one step toward that goal. They show that reinforcement learning can be made more scalable through communication-efficient and asynchronous federated optimization, and more trustworthy through preference-aware and context-aware post-training of language models. More broadly, they suggest that progress in intelligent systems should not be measured only by raw capability, but also by whether those capabilities can be deployed efficiently, safely, and responsibly. In that broader sense, the future of reinforcement learning lies not only in solving harder problems, but also in solving them in ways that are scalable, aligned, and worthy of trust.



\makeatletter  
  \defbibenvironment{bibliography}
    {%
      \list
        {%
          \printtext[labelnumberwidth]%
          {%
            \printfield{prefixnumber}%
            \printfield{labelnumber}%
          }%
        }%
        {%
          \setlength{\bibhang}{1in} 
          \setlength{\itemindent}{1in}
          \setlength{\itemsep}{\bibitemsep}%
          \setlength{\leftmargin}{0pt}
          \setlength{\parsep}{\bibparsep}%
          \setlength{\rightmargin}{0.33in}%
        }%
    }
    {\endlist}
    {\item}
\makeatother  

\setlength{\labelwidth}{1.5in}

{%
  \catcode`*=\active
  \def*{\char'137}
  \PrintBibliography
}

%
\appendices


  \ProvidesFile{ap-appendix_sfedrl.tex}[2023-09-01 mathematics appendix]

\chapter{SFedRL Supplementary Results}
\label{app:sfedrl_experiments}

\section{Proof of Theorem \ref{fednpg_admm_converge_rate}}
\label{ap:proof}

With Assumption \ref{assume:policy} and \ref{assume:fisher}, we restate Lemma B.1. and Proposition G.1. in \cite{liu2020improved} as follows:
\begin{lemma}
\label{lemma:j_value}
$J(\cdot)$ is $L_J$-smooth
\begin{equation}
\left\|\nabla J(\theta_{i}) - \nabla J(\theta_{j})\right\| \leq L_J \left\| \theta_{i} - \theta_{j} \right\|,~\forall \theta_{i}, \theta_{j} \in\mathbb{R}^d,
\end{equation}
where $L_J=\frac{M R}{(1-\gamma)^2}+\frac{2 G^2 R}{(1-\gamma)^3}$. Furthermore, gradients are bounded as $\|\nabla J(\theta)\| \leq \frac{G R}{(1-\gamma)^2}$, $\forall \theta \in\mathbb{R}^d$.
\end{lemma}

\begin{lemma}
\label{lemma:sample_complexity}
At the $k$-th iteration, to achieve $\mathbb{E}\left[ \| w_\star^k - \hat{w}_\star^k \|^2 \right] \le \epsilon'$, it requires $\mathcal{O}(\frac{1}{(1-\gamma)^4 \epsilon'})$ sampling, where $w_\star^k$ is the exact NPG direction and $\hat{w}_\star^k$ is the estimated one.
\end{lemma}

Based on these lemmas, we then give the proof of Theorem \ref{fednpg_admm_converge_rate}.
\begin{proof}
At the $k$-th iteration, let the objective be 
\begin{equation}
\begin{aligned}
l(w)=\mathbb{E}_{(s, a) \sim \nu^{\pi_ {\theta^k}}} \Big[&w^{\top}A_{\pi_{\theta^k}}(s, a)\nabla_{\theta^{k}} \log \pi_{\theta^k}(a \mid s) \\
& -w^{\top} \nabla_{\theta^{k}} \log \pi_{\theta^k}(a \mid s)\nabla_{\theta^{k}} \log \pi_{\theta^k}(a \mid s)^{\top} w\Big],
\end{aligned}
\end{equation}
and the minimizer of $l(w)$ is $w_{\star}^k$. Thus, the exact NPG update is $\theta_{\star}^{k+1}=\theta^{k} + \eta w_{\star}^k$. From Assumption \ref{assume:policy} and \ref{assume:fisher}, we have 
\begin{equation}
\label{npg_bound}
\left\|w_{\star}^{k}\right\|=\left\|F^{-1}(\theta^{k}) \nabla J(\theta^{k})\right\| \leq \frac{G R}{\mu_F(1-\gamma)^2}.
\end{equation}
The difference between the ADMM update and the exact NPG update is 
\begin{equation}
\label{admm_optim_dif}
\theta^{k+1} -\theta^{k+1}_{\star} = \eta (\mathbf{y}^k -w_\star^k).
\end{equation}

However, in practice, we have no access to $\nu^{\pi_\theta}$, and can only compute $w^k_{\star}$ through the empirical loss as follows 
\begin{equation}
\begin{aligned}
\label{eq:empirical_object}
\hat{l}(w)=\frac{1}{\sum_{i=1}^N \|\mathcal{D}_{i} \|} \sum_{\tau \in \cup_{i=1}^N\mathcal{D}_{i}} \Big[& w^{\top}\widehat{A}_{\pi_{\theta^k}}(s, a)\nabla_{\theta^{k}} \log \pi_{\theta^k}(a \mid s) \\
& -w^{\top} \nabla_{\theta^{k}} \log \pi_{\theta^k}(a \mid s)\nabla_{\theta^{k}} \log \pi_{\theta^k}(a \mid s)^{\top} w \Big],
\end{aligned}
\end{equation}
where $\widehat{A}^{\pi_{\theta^k}}(s, a)$ is an unbiased estimate of $A^{\pi_{\theta^k}}(s, a)$. Sample $(s, a) \sim \nu^{\pi_{\theta^{k}}}$ and obtain $\widehat{A}^{\pi_{\theta^{k}}}(s, a)$. By sampling according to the method described in~\cite{agarwal2021theory}, $\widehat{A}^{\pi_{\theta^{k}}}(s, a)$ is guaranteed to be unbiased. We then define its minimizer $\hat{w}^k_\star \coloneqq \mathop{\arg\min}_{w} \hat{l}(w)$.

From the linear convergence of quadratic ADMM, we have 
\begin{equation}
\| \mathbf{y}^k - \hat{w}_\star^k \|^2 \le (1-\frac{\mu_F}{G^2}) \| \mathbf{y}^{k-1} - \hat{w}_\star^k \|^2,
\end{equation}
where the linear convergence coefficient equals the conditional number of $F(\theta)$ (in Assumption~\ref{assume:fisher}) from~\cite{makhdoumi2017convergence}. Let $\zeta \coloneqq (1-\frac{\mu_F}{G^2})$. The variance between the ADMM updating term and the empirical updating term is
\begin{equation}
\label{y_w_variance}
\begin{aligned}
\| \mathbf{y}^k - \hat{w}_\star^k \|^2 \le& (1+c_1)\zeta \| \mathbf{y}^{k-1} - w_\star^k \|^2 + (1+ \frac{1}{c_1})\zeta \| w_\star^k - \hat{w}_\star^k \|^2 \\
\le& (1+c_2) (1+c_1)\zeta \| \mathbf{y}^{k-1} - w_\star^{k-1} \|^2 +(1+\frac{1}{c_2}) (1+c_1)\zeta  \| w_\star^k - w_\star^{k-1} \|^2 \\
&+ (1+ \frac{1}{c_1})\zeta \epsilon_{\text{stats}} \\
\overset{\eqref{npg_bound}}{=}& (1+c_2) (1+c_1)\zeta \| \mathbf{y}^{k-1} - w_\star^{k-1} \|^2 + (1+ \frac{1}{c_1})\zeta \epsilon_{\text{stats}} \\
&+(1+\frac{1}{c_2}) (1+c_1)\zeta \| F^{-1}(\theta^k) \nabla J(\theta^k) - F^{-1}(\theta^{k-1}) \nabla J(\theta^{k-1}) \|^2\\
\le& (1+c_2) (1+c_1)\zeta \| \mathbf{y}^{k-1} - w_\star^{k-1} \|^2 +\frac{2}{\mu_F^2}(1+\frac{1}{c_2}) (1+c_1)\zeta \| \nabla J(\theta^k)\|^2  \\
&+\frac{2}{\mu_F^2}(1+\frac{1}{c_2}) (1+c_1)\zeta \| \nabla J(\theta^{k-1})\|^2+ (1+ \frac{1}{c_1})\zeta \epsilon_{\text{stats}}, \\
\end{aligned}
\end{equation}
where we denote $\epsilon_{\text{stats}} \coloneqq \| w_\star^k - \hat{w}_\star^k \|^2$ for short, and $c_1, c_2 > 0$ are constants. Therefore, from \eqref{admm_optim_dif} with a constant $c_3 > 0$, the variance between the ADMM updating result and the optimal updating result is
\begin{equation}
\label{eq: recursive_eq}
\begin{aligned}
\|\theta^{k+1} -\theta^{k+1}_{\star}\|^2 \le& \eta^2(1+c_3)\| \mathbf{y}^k -\hat{w}_\star^k\|^2 +  \eta^2(1+\frac{1}{c_3})\|\hat{w}_\star^k -w_\star^k\|^2 \\
\overset{\eqref{y_w_variance}}{\le}& \eta^2(1+c_3) (1+c_2) (1+c_1)\zeta \| \mathbf{y}^{k-1} - w_\star^{k-1} \|^2 \\
&+\frac{2\eta^2}{\mu_F^2}(1+{c_3})(1+\frac{1}{c_2}) (1+c_1)\zeta \big(\| \nabla J(\theta^k)\|^2 + \| \nabla J(\theta^{k-1})\|^2 \big) \\
&+ \eta^2 \left[1+ \frac{1}{c_3}+(1+{c_3})(1+\frac{1}{c_1})\zeta\right] 
\epsilon_{\text{stats}} \\
\overset{\eqref{admm_optim_dif}}{=}& (1+c_3)(1+c_2) (1+c_1)\zeta \| \theta^{k} -\theta^{k}_{\star} \|^2\\
&+\frac{2\eta^2}{\mu_F^2}(1+{c_3})(1+\frac{1}{c_2}) (1+c_1)\zeta \big(\| \nabla J(\theta^k)\|^2 + \| \nabla J(\theta^{k-1})\|^2 \big) \\
&+ \eta^2 \left[1+ \frac{1}{c_3}+(1+{c_3})(1+\frac{1}{c_1})\zeta\right] 
\epsilon_{\text{stats}}.
\end{aligned}
\end{equation}
Choose $c_1 = c_2 =c_3 =\frac{\mu_F}{14G^2} \le \frac{1}{14}$. Then we have 
\begin{equation}
\label{eq:constants_c}
(1+c_1)(1+c_2)(1+c_3)\zeta \le (1+7c_1)\zeta \le 1 -\frac{\mu_F}{2G^2}.
\end{equation}
The above inequality can be viewed as 
\begin{equation}
\|\theta^{k+1} -\theta^{k+1}_{\star}\|^2 \le \alpha \|\theta^{k} -\theta^{k}_{\star}\|^2 +\gamma_{k},
\end{equation}
where $ \alpha \coloneqq 1 -\frac{\mu_F}{2G^2}$ and 
\begin{equation}
\label{eq:gamma_expression}
\begin{aligned}
\gamma_{k} \coloneqq& \frac{2\eta^2}{\mu_F^2}(1+\frac{1}{c_3})(1+\frac{1}{c_2}) (1+c_1)\zeta\| \nabla J(\theta^k)\|^2 \\
&+ \frac{2\eta^2}{\mu_F^2}(1+{c_3})(1+\frac{1}{c_2}) (1+c_1)\zeta(\| \nabla J(\theta^k)\|^2 + \| \nabla J(\theta^{k-1})\|^2) \\
&+ \eta^2 \left[1+ \frac{1}{c_3} + (1+{c_3})(1+ \frac{1}{c_1})\zeta\right] 
\epsilon_{\text{stats}}.
\end{aligned}
\end{equation}

From Lemma \ref{lemma:j_value} and the Descent Lemma, we have 
\begin{equation}
\label{J_iteration}
\begin{aligned}
J\left(\theta^{k+1}\right) \geq&  J\left(\theta^k\right) +\left\langle\nabla J\left(\theta^k\right), \theta_{\star}^{k+1}-\theta^k\right\rangle+\left\langle\nabla J\left(\theta^k\right), \theta^{k+1}-\theta_{\star}^{k+1}\right\rangle-\frac{L_J}{2}\left\|\theta^{k+1}-\theta^k\right\|^2 \\
=& J\left(\theta^k\right) + \eta\left\langle\nabla J\left(\theta^k\right), F^{-1}\left(\theta^k\right) \nabla J\left(\theta^k\right)\right\rangle \\
& +\left\langle\nabla J\left(\theta^k\right), \theta^{k+1}-\theta_{\star}^{k+1}\right\rangle-\frac{L_J}{2}\left\|\theta^{k+1}-\theta^k\right\|^2 \\
\geq& J\left(\theta^k\right)+\frac{\eta}{G^2}\left\|\nabla J\left(\theta^k\right)\right\|^2 +\left\langle\nabla J\left(\theta^k\right), \theta^{k+1}-\theta_{\star}^{k+1}\right\rangle -\frac{L_J}{2}\left\|\theta^{k+1}-\theta^k\right\|^2 \\
\geq& J\left(\theta^k\right)+\frac{\eta}{2 G^2}\left\|\nabla J\left(\theta^k\right)\right\|^2-\frac{G^2}{2 \eta}\left\|\theta^{k+1}-\theta_{\star}^{k+1}\right\|^2-\frac{L_J}{2}\left\|\theta^{k+1}-\theta^k\right\|^2 \\ 
\geq&  J\left(\theta^k\right)+\frac{\eta}{2 G^2}\left\|\nabla J\left(\theta^k\right)\right\|^2-\left(\frac{G^2}{2 \eta}+L_J\right)\left\|\theta^{k+1}-\theta_{\star}^{k+1}\right\|^2-L_J\left\|\theta_{\star}^{k+1}-\theta^k\right\|^2 \\ 
\overset{\eqref{npg_bound}}{\geq}&  J\left(\theta^k\right)+\left(\frac{\eta}{2 G^2}-\frac{L_J \eta^2}{\mu_F^2}\right)\left\|\nabla J\left(\theta^k\right)\right\|^2-\left(\frac{G^2}{2 \eta}+L_J\right)\left\|\theta^{k+1}-\theta_{\star}^{k+1}\right\|^2.
\end{aligned}\end{equation}

Construct a potential value $\Psi_{k} {\coloneqq} J(\theta^{k}) - \Phi_k\left\|\theta^{k}-\theta_{\star}^{k}\right\|^2$, where $\Phi_k$ is a scaling scalar. We have
\begin{equation}\label{eq:error_decrease}
\begin{aligned}
\Psi_{k+1} \overset{\eqref{J_iteration}}{\ge}& \Psi_{k} +\left(\frac{\eta}{2 G^2}-\frac{L_J \eta^2}{\mu_F^2}\right)\left\|\nabla J\left(\theta^k\right)\right\|^2\\
&-\left( \Phi_{k+1}+\frac{G^2}{2 \eta}+L_J\right)\left\|\theta^{k+1}-\theta_{\star}^{k+1}\right\|^2 + \Phi_k\left\|\theta^{k}-\theta_{\star}^{k}\right\|^2 \\
\stackrel{\eqref{eq: recursive_eq}}{\ge}& \Psi_{k} +\left(\frac{\eta}{2 G^2}-\frac{L_J \eta^2}{\mu_F^2}\right)\left\|\nabla J\left(\theta^k\right)\right\|^2 \\
&+ \left[\Phi_{k}-\alpha\left( \Phi_{k+1}+\frac{G^2}{2 \eta}+L_J\right)\right]\left\|\theta^{k}-\theta_{\star}^{k}\right\|^2 -\gamma_k.
\end{aligned}
\end{equation}
Here we require $\Phi_{k}-\alpha\left( \Phi_{k+1}+\frac{G^2}{2 \eta}+L_J\right) = 0$, i.e., $\Phi_k = \frac{\alpha(\frac{G^2}{2 \eta}+L_J)}{(1-\alpha)}.$ Sum Eq~\eqref{eq:error_decrease} from $k=1$ to $k=K$, and we then have
\begin{equation}
\begin{aligned}
\label{eq:sum_potential}
&\left(\frac{\eta}{2 G^2}-\frac{L_J \eta^2}{\mu_F^2}\right)\sum_{k=1}^{K} \mathbb{E}\left[\left\|\nabla J\left(\theta^k\right)\right\|^2\right] \\
\le& \Psi_{K+1}-\Psi_1 + \sum_{k=1}^K\gamma_k \\
\stackrel{\eqref{eq:gamma_expression}}{\le}& \Psi_{K+1}-\Psi_1 +\eta^2 \sum_{k=1}^K\left[1+ \frac{1}{c_3} + (1+{c_3})(1+ \frac{1}{c_1})\zeta
\epsilon_{\text{stats}}\right] \\
&+ \sum_{k=1}^K\left[\frac{2\eta^2}{\mu_F^2}(1+{c_3})(1+\frac{1}{c_2}) (1+c_1)\zeta\big(\| \nabla J(\theta^k)\|^2 + \| \nabla J(\theta^{k-1})\|^2\big)\right]\\
\stackrel{\eqref{eq:constants_c}}{\le}& J^{\star} - J(\theta^{1}) + \Phi_1 \lVert \theta^{1}-\theta_{\star}^{1}\rVert^2 +\eta^2 \frac{28G^2}{\mu_F} \sum_{k=1}^K\left[1 + 2\zeta
\epsilon_{\text{stats}}\right] \\
&+ \eta^2 \frac{2}{\mu_F^2} \frac{14G^2}{\mu_F} (1 -\frac{\mu_F}{2G^2}) \sum_{k=1}^K\left[\big(\| \nabla J(\theta^k)\|^2 + \| \nabla J(\theta^{k-1})\|^2\big)\right],
\end{aligned}
\end{equation}
where $J^{\star}$ is the maximized value by the optimal policy and $\theta_{\star}^{1}=\theta^{1}$ is the initial setting. Sample $\mathcal{O}(\frac{1}{(1-\gamma)^4 \epsilon})$ trajectories for $\hat{w}_\star^k$ in Lemma \ref{lemma:sample_complexity} and then achieve $\sum_{k=1}^{K} \epsilon_{\text{stats}} \le K\epsilon$. Choose $\eta = \frac{\mu_{F}^2}{4G^2(56G^2 + L_J)}$ and
\begin{equation}
K = \frac{(J^{\star} - J(\theta^{1}))(56G^2 + L_J)^{2}16G^2 + 28G^{2}\mu_{F}^3}{(56G^2 + L_J - 56G^{2}\mu_{F})\mu_{F}^2\epsilon} = \mathcal{O}\left(\frac{1}{(1-\gamma)^{2}\epsilon}\right).
\end{equation}
Then, rewrite \eqref{eq:sum_potential} and we arrive at
\begin{equation}
\begin{aligned}
\frac{1}{K}\sum_{k=1}^{K} \mathbb{E}\left[\left\|\nabla J\left(\theta^k\right)\right\|^2\right]
\le \frac{J^{\star} - J(\theta^{1}) + \eta^2\frac{28G^2}{\mu_F} + \eta^2 \frac{56G^2}{\mu_F} K\epsilon}{K\eta \left[ 
\frac{1}{2G^2} - \frac{\eta}{\mu_{F}^2}(56G^2 + L_J) \right] } = \epsilon.
\end{aligned}
\end{equation}
\end{proof}
Recall that the total trajectories $\sum_{i=1}^N \|\mathcal{D}_{i} \|$ in \eqref{eq:empirical_object} are collected by $N$ agents equally using the common global policy $\pi_{\theta}$ at each iteration. Each agent $i$ samples $ \|\mathcal{D}_{i} \| = \mathcal{O}(\frac{1}{(1-\gamma)^4 N \epsilon})$ at each iteration and enjoys a federated sampling benefit compared to a single agent with $ \|\mathcal{D}_{i} \| = \mathcal{O}(\frac{1}{(1-\gamma)^4 \epsilon})$. During the whole training process, each agent $i$ has $K \|\mathcal{D}_{i} \| = \mathcal{O}(\frac{1}{(1-\gamma)^{6} N {\epsilon}^{2}})$ sample complexity and $K\cdot 2d = \mathcal{O}(\frac{d}{(1-\gamma)^2 \epsilon})$ communication complexity to achieve a stationary point.

  \ProvidesFile{ap-appendix_afedrl.tex}[2023-09-01 mathematics appendix]

  

\chapter{AFedRL Supplementary Results}
\label{app:experiments}

In this section, we first compare the time complexity between the synchronous and the asynchronous settings. We then list the experimental settings in Appendix \ref{sec:app_exp_setting}. At last, we have supplementary experiments in Appendix \ref{app:supp_results}.

\section{Comparison to the Synchronous Setting}
\label{app:experiments_comp}

Let $T$ be the computation time during the entire training process to achieve a given number $K$ of cumulative communication rounds of all agents, where $K = \mathcal{O}({\epsilon}^{-2.5})$ for a global convergence according to Theorem \ref{theorem:afedpg_rate}.

In AFedPG, for agent $i$, the number of communication rounds is $\frac{T}{t_{i}}$. For all agents, the total number is $\sum_{i=1}^{N} \frac{T}{t_{i}}$. As $\sum_{i=1}^{N} \frac{T}{t_{i}} = K$, we have $T = \frac{K}{\sum_{i=1}^{N} \frac{1}{t_{i}}} = \mathcal{O}(\frac{1}{\sum_{i=1}^{N} \frac{1}{t_{i}}} {\epsilon}^{-2.5})$ as a harmonic average of all agents.

In FedPG, the number of global communication rounds on the server is $\frac{T}{t_{\max}}$. As $\frac{T}{t_{\max}} = \frac{K}{N}$, we have $T = \frac{K t_{\max}}{N} = \mathcal{O}(\frac{t_{\max}}{N} {\epsilon}^{-2.5}) \ge \mathcal{O}(\frac{1}{\sum_{i=1}^{N} \frac{1}{t_{i}}} {\epsilon}^{-2.5})$ as the harmonic mean is always smaller or equal to the maximum one.

\begin{figure}[!ht]
\centering
\includegraphics[width=0.8\linewidth]{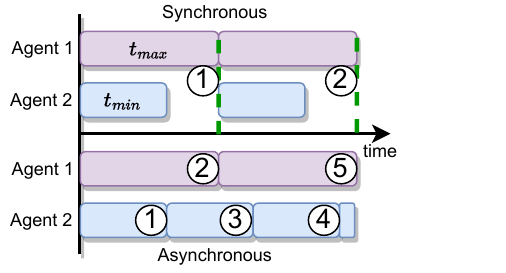}
\caption{Comparison of time consumptions between synchronous and asynchronous approaches. The circled numbers denote the indices of global steps.}
\label{fig:asyn_time}
\end{figure}

The asynchronous FedPG achieves better time complexity than the synchronous approach regardless of the delay pattern. The advantage is significant when $t_{\max} \gg t_{min}$, which occurs in many practical settings with heterogeneous computation powers across different agents. We illustrate the advantage of AFedPG over synchronous FedPG in Figure \ref{fig:asyn_time}. As the server only operates one simple summation, without loss of generality, the time consumption at the server-side is negligible.

It is noticeable that we do not make any assumptions or requirements on $t_{\max}$ in the analysis of AFedPG. In the extreme case, the slowest agent does not communicate with the server, and thus, $t_{\max}$ is infinite. In this scenario, the time consumption of AFedPG does not hurt a lot, while the time consumption of FedPG becomes infinite.

\section{Supplementary Experiments}

\subsection{Supplementary Experimental Settings}
\label{sec:app_exp_setting}

In Section \ref{sec:exp_setup}, we list the key experimental settings. We list the rest with details in this subsection. The environmental details (the four MuJoCo tasks) are described in Table \ref{table:mujoco}. The policies $\pi_{\theta}$ are parameterized by fully connected multi-layer perceptions (MLPs) with settings listed in Table \ref{table:hyperparameters}.

\begin{table}[t]
\centering
\renewcommand\arraystretch{1.1} 
\caption{Detailed descriptions of the four tasks in the MuJoCo environment.}
\label{table:mujoco}
\begin{tabular}{ccccc}
\toprule[1.1pt]
\rowcolor{gray!10} {}& {}& {Action Space}& {State Space} \\
\rowcolor{gray!10} \multirow{-2}{*}{MuJoCo Tasks} & \multirow{-2}{*}{Agent Types} & {Dimension} & {Dimension} \\
\midrule[1.1pt]
 {} & {Three-link} & {} & {} \\
\multirow{-2}{*}{Swimmer-v4} & {swimming robot} & \multirow{-2}{*}{$2$} & \multirow{-2}{*}{$8$} \\
\hline
\rowcolor{gray!10} {} & {Two-dimensional} & {} & {} \\
\rowcolor{gray!10} \multirow{-2}{*}{Hopper-v4} & {one-legged robot} & \multirow{-2}{*}{$3$} & \multirow{-2}{*}{$11$} \\
\hline
{} & {Two-dimensional} & {} & {} \\
\multirow{-2}{*}{Walker2D-v4} & {bipedal robot} & \multirow{-2}{*}{$6$} & \multirow{-2}{*}{$17$} \\
\hline
\rowcolor{gray!10} {} & {Three-dimensional} & {} & {} \\
\rowcolor{gray!10} \multirow{-2}{*}{Humanoid-v4} & {bipedal robot} & \multirow{-2}{*}{$17$} & \multirow{-2}{*}{$376$}\\
\bottomrule[1.1pt]
\end{tabular}
\end{table}

\begin{table}[t]
\centering
\caption{Hyperparameters of AFedPG and the MLP policy parameterization settings.}
\label{table:hyperparameters}
\begin{tabular}{l|c|c|c|c}
\toprule[1.1pt]
\rowcolor{gray!10} {Hyperparameter} & \multicolumn{4}{c}{Setting} \\
\midrule[1.1pt]
{Task} & {Swimmer-v4} & {Hopper-v4} & {Walker2D-v4} & {Humanoid-v4} \\
\rowcolor{gray!10} MLP & $64\times 64$ & $256\times 256$ & $512\times 512$ & $512\times 512\times 512$ \\
 Activation function & ReLU & ReLU & ReLU & ReLU \\
\rowcolor{gray!10} Output function & Tanh & Tanh & Tanh & Tanh \\
 Learning rate ($\alpha$) & $1\times 10^{-3}$ & $1\times 10^{-3}$ & $1\times 10^{-3}$ & $1\times 10^{-3}$ \\
\rowcolor{gray!10} Discount ($\gamma$) & $0.99$ & $0.99$ & $0.99$ & $0.99$ \\
Timesteps ($T$) & $2048$ & $1024$ & $1024$ & $512$ \\
\rowcolor{gray!10} Iterations ($K$) & $1\times 10^{3}$ & $2\times 10^{3}$ & $5\times 10^{3}$  & $1.5\times 10^{4}$ \\
Learning rate ($\eta$) & $3\times 10^{-4}$ & $1\times 10^{-4}$ & $5\times 10^{-5}$  & $1\times 10^{-5}$ \\
\bottomrule[1.1pt]
\end{tabular}
\end{table}

\subsection{Supplementary Experimental Results}
\label{app:supp_results}

In this subsection, we show three experimental results to study the effect of computation heterogeneity, communication overhead, and reward performances with long runs.

\textbf{Effect of computation heterogeneity.} 
We study the effect of the computation heterogeneity among federated agents. The heterogeneity of computing powers is measured by the ratio $\frac{t_{\max}}{t_{\min}}$, and the effect is measured by the speedup, which is the global time of FedPG divided by that of AFedPG. Without loss of generality, we test the performances with (1) One straggler, which is one agent has $t_{\min}$ time consumption, and all the other agents have $t_{\max}$ time consumption. This is the scenario with the largest speedup. (2) One leader, which is one agent has $t_{\max}$ and the others have $t_{\min}$. This is the scenario with the smallest speedup. 

\begin{figure}[!ht]
\centering
{\subcaptionbox{With one straggler}[2.8in]{\includegraphics[width=2.8in]{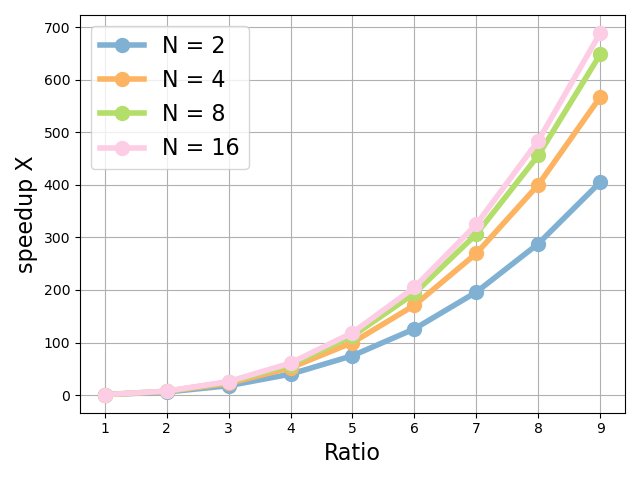}}}
\hskip 0.4truein
{\subcaptionbox{With one leader}[2.8in]{\includegraphics[width=2.8in]{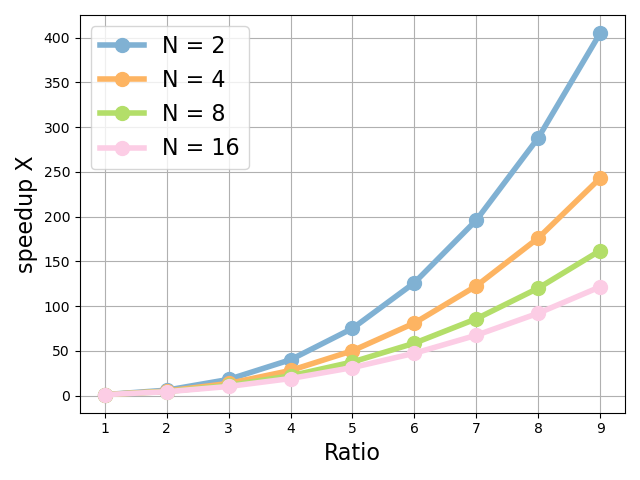}}}
\caption{The time complexity speedup of AFedPG compared to FedPG. $N$ is the number of agents. The x-axis is the heterogeneity ratio of computing power measured by $\frac{t_{\max}}{t_{\min}}$. The y-axis is the global time of FedPG divided by that of AFedPG.}
\label{fig:speedup}
\end{figure}

In Figure \ref{fig:speedup}, the results show that the speedup increases as the heterogeneity ratio $\frac{t_{\max}}{t_{\min}}$ increases. With one leader, as more agents participate, the speedup approximately decreases to a quadratic function w.r.t. the time heterogeneity ratio. With one straggler, the more agents that participate in the training process, the higher speedup they achieve. The overall results indicate that AFedPG has the potential to scale up to a very large RL system, particularly in scenarios with extreme stragglers (The ratio is large: $t_{\max}\gg t_{\min}$).

\begin{figure}[!ht]
\centering
{\subcaptionbox{On the agent side}[2.8in]{\includegraphics[width=2.8in]{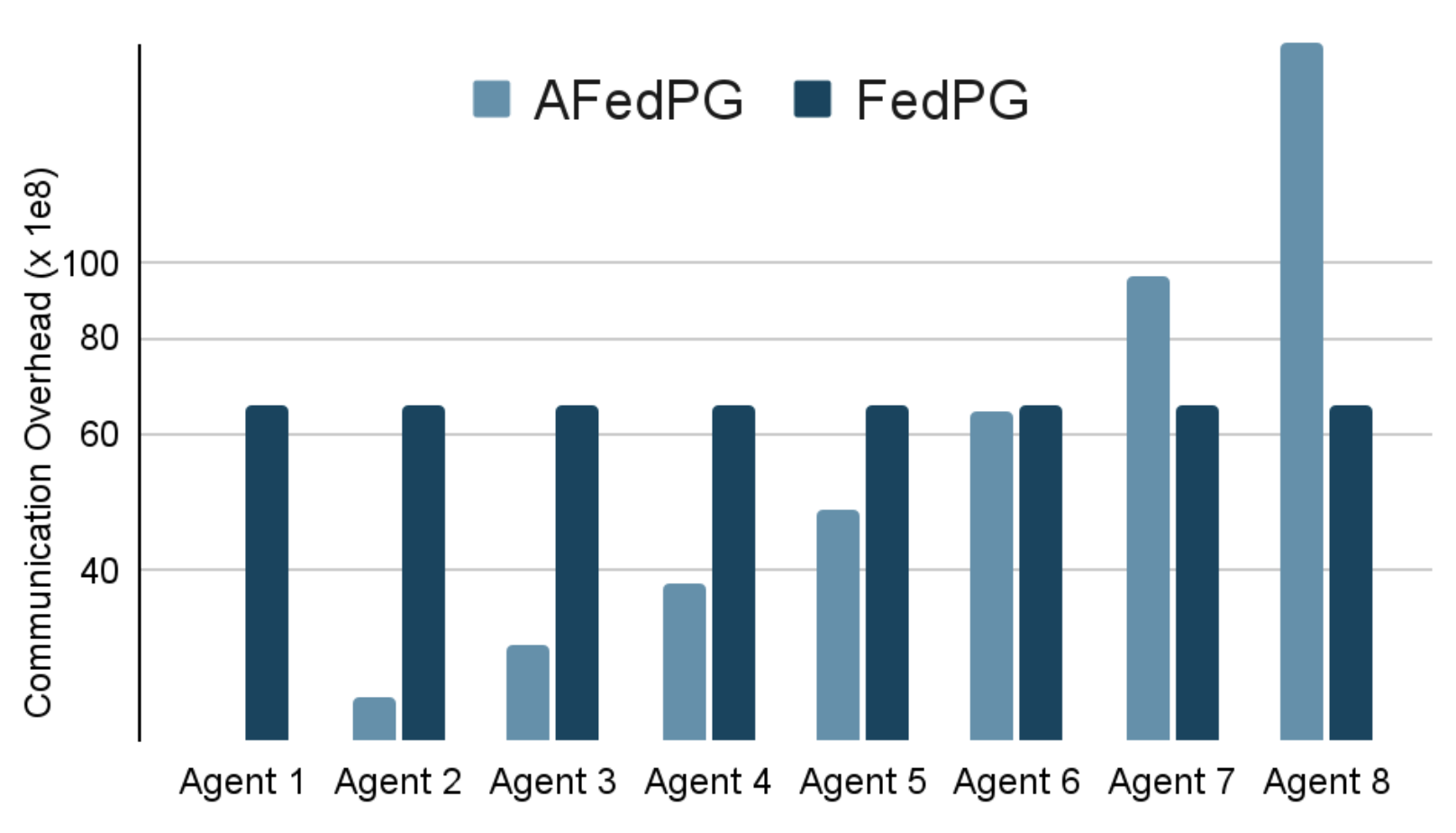}}}
\hskip 0.4truein
{\subcaptionbox{Downlink on the server side}[2.8in]{\includegraphics[width=2.8in]{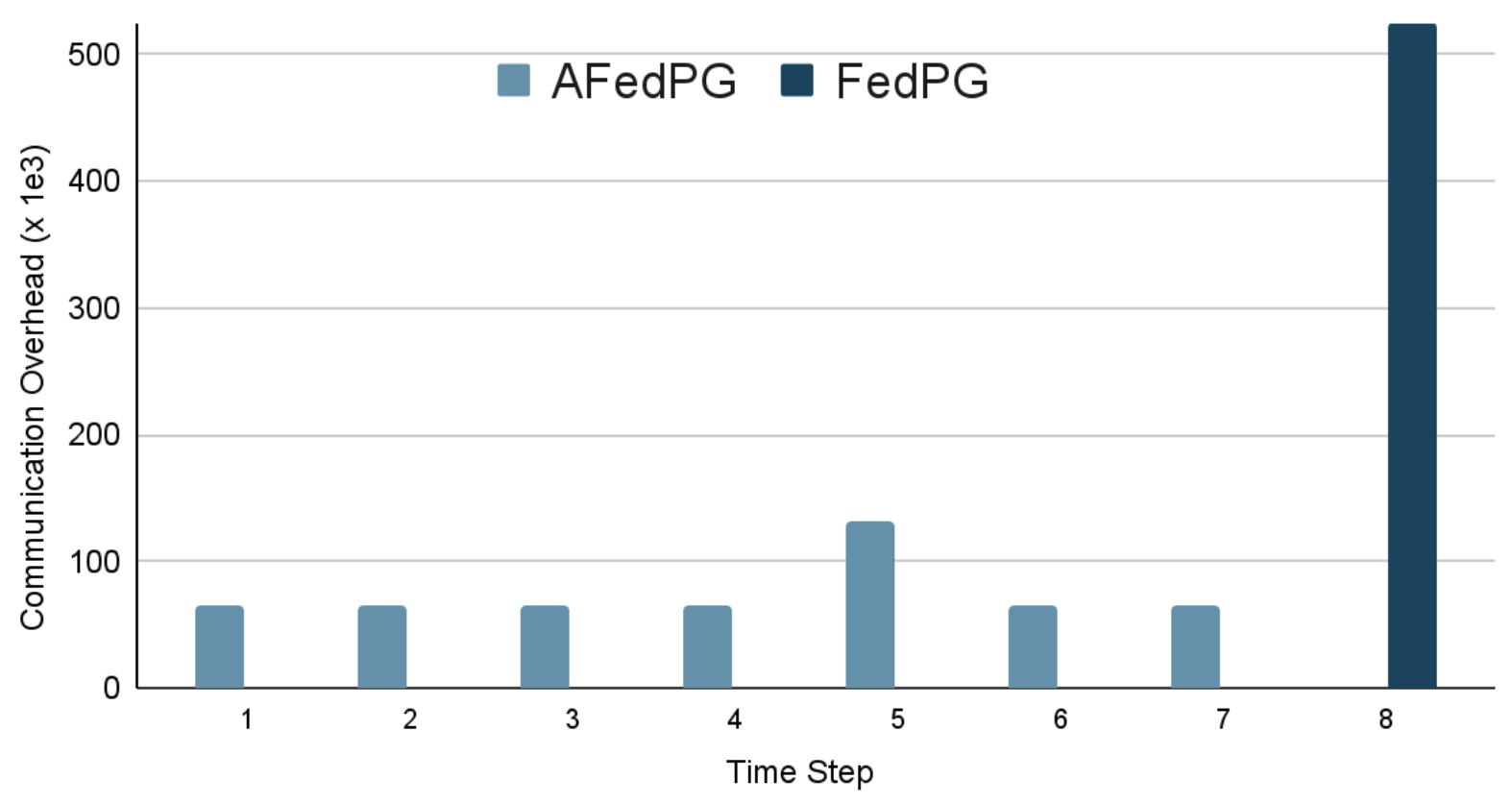}}}
\caption{The communication overhead of AFedPG compared to FedPG. The number of agents is $N=8$. (a) The cumulative communication overhead on the agent side. (b) The downlink communication overhead in a time window on the server side.}
\label{fig:commun_a}
\end{figure}

\textbf{Communication overhead analysis.}
We compare the communication overhead of FedPG and AFedPG in Figure \ref{fig:commun_a}. For neural network parameters, we use the standard format float32. The communication overhead is measured by the number of transmitted bytes. The number of federated agents $N$ is set to $8$. The MuJoCo task is Swimmer-v4. Overall, the commutative communication overhead is similar. However, when we dive deep into the agent side and the server side separately in fine-grained time, AFedPG shows advantages on both sides.

On the agent side, Figure \ref{fig:commun_a}~(a) shows the cumulative communication bytes during the training process. The cumulative communication overhead is similar in FedPG and AFedPG. In AFedPG, the faster ones, \textit{e.g.}, Agent 8, communicate more, and the slower ones, \textit{e.g.}, Agent 1, communicate less, which is more reasonable than the equal allocation in the synchronous setting. The Agent 1, in fact, commutes with the server during the training process, but it is insignificant in the plot with a log scale.

The agents have heterogeneous resources, but equal allocation could bring too much burden for the slower ones. In AFedPG, it naturally shifts these burdens to the faster ones considering the local resources.

On the server side, we show the downlink communication overhead in a time window in Figure \ref{fig:commun_a}~(b). The time window is set as ``one global round in the synchronous setting''. At time step 5, it is higher because two agents communicate with the server in a short time period. The total amounts of AFedPG and FedPG are similar. In AFedPG, it is almost evenly distributed during the time span, while in FedPG, the server has a huge burden with a peak. This makes the server in FedPG require huge resources.

\begin{figure}[!ht]
\centering
\subfloat[Swimmer-v4]{
	\includegraphics[width=2.8in]{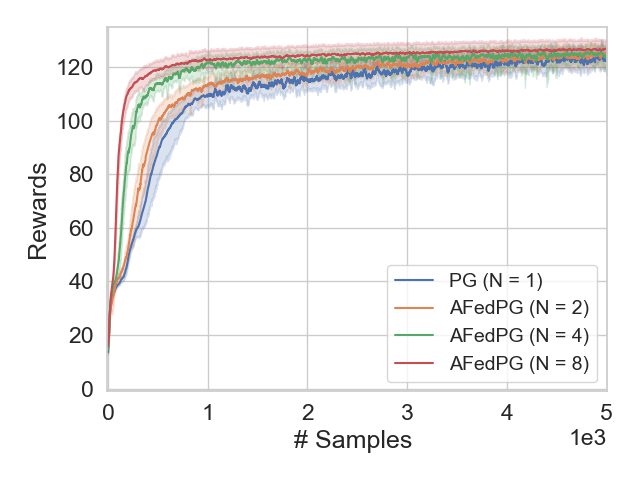}
	}
\subfloat[Hopper-v4]{
	\includegraphics[width=2.8in]{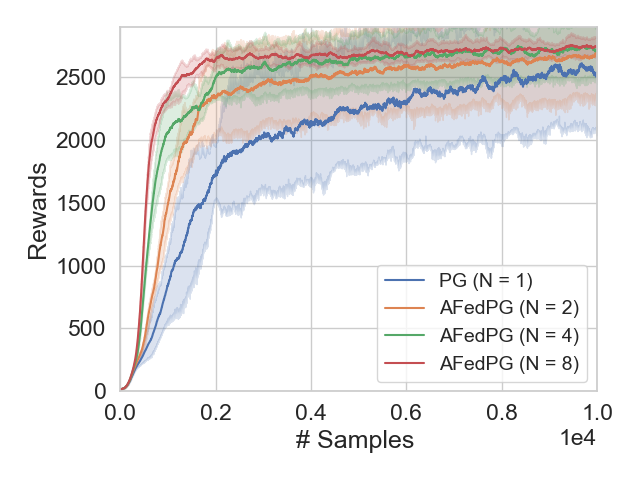}
	}  

\subfloat[Walker2D-v4]{
	\includegraphics[width=2.8in]{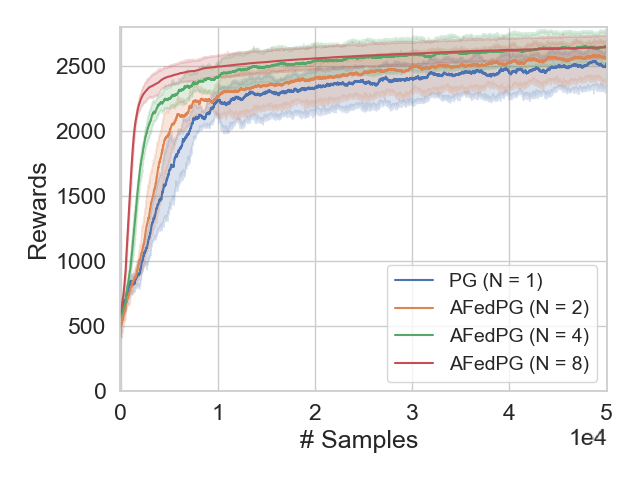}
	}
\subfloat[Humanoid-v4]{
	\includegraphics[width=2.8in]{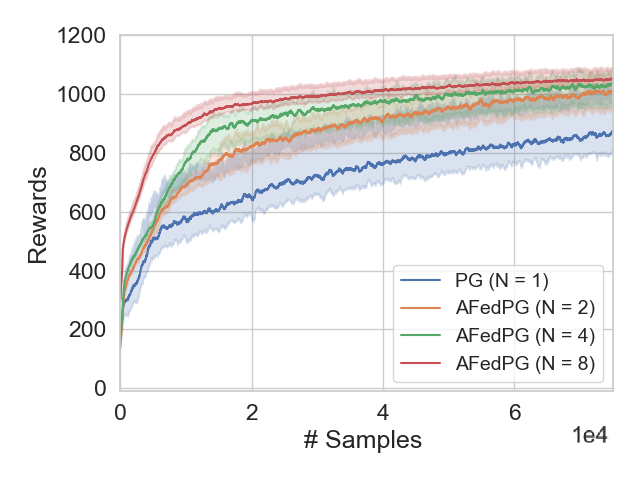}
	} 
\caption{Reward performances of AFedPG ($N=2,4,8$) and PG ($N=1$) on various MuJoCo environments, where $N$ is the number of federated agents. The solid lines are averaged results over $5$ runs with random seeds from $0$ to $4$. The shadowed areas are confidence intervals with $95\%$ confidence level.}
\label{fig:fed_speedup_long}
\end{figure}

\textbf{Reward performances with long runs.}
To further enhance the results in Figure \ref{fig:fed_speedup}, we extend the running time with more samples. Compared to the results in Figure \ref{fig:fed_speedup}, we increase the number of samples by $5$ times in each Mujoco task in Figure \ref{fig:fed_speedup_long}. The solid lines are averaged results over $5$ runs with random seeds from $0$ to $4$. The shadowed areas are confidence intervals with $95\%$ confidence level. 

With enough samples, it basically achieves a similar reward performance for different numbers of agents $N$ in AFedPG. However, the more agents engage, the faster it achieves. Notably, the solid line is the average result with $5$ independent runs. With different numbers of agents, the shadowed area has a large overlap. The more overlaps they have, the more runs that have similar reward performances because of the inherent randomness (with different random seeds) in the deep reinforcement learning tasks.

For the Humanoid-v4 task, though in some runs (shadowed area), PG achieves the optimal reward performance, the average (solid line) performance of PG is relatively lower than the others in AFedPG. The reason is that the Humanoid-v4 task has the largest state and action space, which makes the hyperparameter tuning difficult, \textit{e.g.}, learning rates, and brings huge GPU hours. The hyperparameter setting of PG is suboptimal here, as it has no contribution to our main claim. Recall that we aim to use these experiments to verify the speedup effect in AFedPG in Table \ref{table:afedrl_complexity}. The suboptimal hyperparameters of PG in the Humanoid-v4 task do not influence the conclusion: The more agents in AFedPG, the faster the optimal reward will be achieved.

\section{Theoretical Proofs}
\label{app:proof}

In this section, we give the assumptions in Appendix \ref{app:assumptions}, the technical lemmas in Appendix \ref{app:Technical_Lemmas}, our key lemmas in Appendix \ref{app:Key_Lemmas}, the proof of Theorem \ref{theorem:afedpg_rate} (the global convergence) in Appendix \ref{app:proof_theorem_1}, and the proof of Theorem \ref{theorem:afedpg_rate_FOSP} (the FOSP convergence) in Appendix \ref{app:proof_theorem_2}.

\subsection{Assumptions}
\label{app:assumptions}

In order to derive the global convergence rates, we make the following standard assumptions \cite{agarwal2021theory, NEURIPS2020_5f7695de, liu2020improved, pmlr-v80-papini18a, xu2020Sample} on policy gradients, and rewards.

\begin{assumption}{\ }
\label{assum:policy}
\begin{enumerate}
\item The score function is bounded as $\left\|\nabla \log \pi_\theta(a \mid s)\right\| \leq M_{g}$, $\text{ for all }\theta \in\mathbb{R}^d$, $s\in\mathcal{S}$, and $a\in\mathcal{A}$.
\item The score function is $M_{h}$-Lipschitz continuous. In other words, for all $\theta_{i},~\theta_{j} \in\mathbb{R}^d,~s\in\mathcal{S},~\text{and}~ a\in\mathcal{A}$, we have
\begin{equation}
\begin{aligned}
\left\|\nabla \log \pi_{\theta_{i}}(a \mid s) - \nabla \log \pi_{\theta_{j}}(a \mid s)\right\| &\leq M_{h} \left\|\theta_{i} - \theta_{j}\right\|.
\end{aligned}
\end{equation}
\item The reward function is bounded as $r(s,a) \in [0, R], \text{ for all } s\in\mathcal{S}$, and $a\in\mathcal{A}$.
\end{enumerate}
\end{assumption}

These standard assumptions naturally state that the reward, the first-order, and the second-order derivatives of the score function are not infinite, and they hold with common practical parametrization methods, \textit{e.g.}, softmax policies. With function approximation $\pi_{\theta}$, the approximation error may not be $0$ in practice. 

Thus, we follow previous works \cite{liu2020improved, agarwal2021theory, ding2022, huang2020momentum} with an assumption on the expressivity of the policy parameterization class.

\begin{assumption}
\label{assum:func_approx}
(Function approximation)
$\exists~\epsilon_{{\rm bias}}\geq 0$ \textit{s.t.} for all $\theta \in\mathbb{R}^{d}$, the transfer error satisfies
\begin{equation}
\label{eq:approx_error}
\mathbb{E}[\big( A_{\pi_{\theta}}(s,a) - (1-\gamma)u^{\star}(\theta)^{\top} \nabla \log \pi_{\theta}(a \mid s) \big)^{2}] \leq \epsilon_{{\rm bias}},
\end{equation}
where $u^{\star}(\theta) \coloneqq F_{\rho}(\theta)^{\dagger} \nabla J(\theta)$, and $F_{\rho}(\theta)^{\dagger}$ is the Moore-Penrose pseudo-inverse of the Fisher matrix $F_{\rho}$.
\end{assumption}
It means that the parameterized policy $\pi_{\theta}$ makes the advantage function $A_{\pi_{\theta}}(s, a)$ approximated by the score function $\nabla \log \pi_{\theta}(a \mid s)$ as the features. This assumption is widely used with Fisher-non-degenerate parameterization. $\epsilon_{{\rm bias}}$ can be very small with rich neural network parameterization, and $0$ with a soft-max parameterization \cite{Wang2020Neural}.

To achieve the global convergence, we make a standard assumption on the Fisher information matrix \cite{liu2020improved, ding2022, lan2023}.

\begin{assumption}
\label{assum:semi_pos}
(Positive definite) For all $\theta \in\mathbb{R}^d$, there exists a constant $\mu_{F} > 0$ \textit{s.t.} the Fisher information matrix $F_{\rho}(\theta)$ induced by the policy $\pi_{\theta}$ and the initial state distribution $\rho$ satisfies
\begin{equation}
F_{\rho}(\theta) \succcurlyeq \mu_{F} \cdot I,
\end{equation}
where $I \in\mathbb{R}^{d \times d}$ is an identity matrix.
\end{assumption}

For any two symmetric matrices $A$ and $B$ with the same dimension, $A \succcurlyeq B$ denotes that the eigenvalues of $A - B$ are greater or equal to zero.

\subsection{Technical Lemmas}
\label{app:Technical_Lemmas}

\begin{lemma}
\label{lemma:triangle}
For arbitrary $n$ vectors $\{\mathbf{a}_{i} \in\mathbb{R}^{d}\}_{i=1}^{n}$, we have
\begin{equation}
\label{eq:triangle}
\|\sum_{i=1}^{n} \mathbf{a}_{i} \|^{2} ~\leq~ n\sum_{i=1}^{n}\| \mathbf{a}_{i} \|^{2}.
\end{equation}
\end{lemma}

\begin{lemma}
\label{lemma:lr_bound}
Let $\alpha_{k} = (\frac{c}{t+c})^{p}$. $\forall p \in [0, 1]$ and $c\ge 1$, we have
\begin{equation}
1 - \alpha_{k+1} ~\leq~ \frac{\alpha_{k+1}}{\alpha_{k}}.
\end{equation}
\begin{proof}
\begin{equation}
\begin{aligned}
1 - \alpha_{k+1} ~=&~ 1 - (\frac{c}{t+1 + c})^{p} \\
~\leq&~ 1 - \frac{1}{t+1 + c} \\
~\leq&~ \frac{\alpha_{k+1}}{\alpha_{k}}.
\end{aligned}
\end{equation}
Straightforwardly, we have
\begin{equation}
\begin{aligned}
\label{eq:lr_bound}
\prod_{i=k}^{K-1} (1 - \alpha_{i+1}) ~\leq&~ \frac{\alpha_{K}}{\alpha_{k}}.
\end{aligned}
\end{equation}
\end{proof}
\end{lemma}

\begin{lemma}
\label{lemma:lr_seq_bound}
Let $\alpha_{k} = (\frac{1}{k+1})^{p}$ and $\eta_{k} = \eta_{0}(\frac{1}{k+1})^{q}$. $\forall p \in [0, 1)$, $q \ge 0$ and $\eta_{0} \ge 0$, we have
\begin{equation}
\label{eq:lr_seq_bound}
\sum_{k=0}^{K-1}\eta_{k} \prod_{i=k+1}^{K-1}(1 - \alpha_{i}) ~\leq~ c(p,q) \frac{\eta_{K}}{\alpha_{K}},
\end{equation}
where $c(p,q) \coloneqq \frac{2^{q-p}}{1-p} a \exp{\big((1-p)2^{p}a^{1-p}\big)}$ is a constant with specified $p$ and $q$, and $a = \max \big( (\frac{q}{(1-p)2^{p}})^{\frac{1}{1-p}}, (\frac{2(q-p)}{(1-p)^2})^{\frac{1}{1-p}} \big)$.
\end{lemma}

\subsection{Key Lemmas}
\label{app:Key_Lemmas}

In this subsection, we list four useful (key) lemmas to construct the proofs of Theorem \ref{theorem:afedpg_rate} and Theorem \ref{theorem:afedpg_rate_FOSP}. 

Under Assumption \ref{assum:policy} on score functions, the following lemma holds based on the results (Lemma 5.4) in \cite{zhang2020global}.
\begin{lemma}
\label{lemma:exp_return}
The gradient of the expected return is $L_{g}$-continuous and $L_{h}$-smooth as follows
\begin{equation}
\begin{aligned}
\|\nabla J(\theta) - \nabla J(\theta')\| ~\leq&~ L_{g}\|\theta - \theta'\|, \\
\|\nabla^{2} J(\theta) - \nabla^{2} J(\theta')\| ~\leq&~ L_{h}\|\theta - \theta'\|,
\end{aligned}
\end{equation}
where $L_{g} \coloneqq \frac{R(M_{g}^{2} + M_{h})}{(1-\gamma)^{2}}$ and $L_{h} \coloneqq \frac{RM_{g}^{3}(1+\gamma)}{(1-\gamma)^{3}} + \frac{RM_{g}M_{h}}{(1-\gamma)^{2}} + \mathcal{O}\big((1-\gamma)^{-1}\big)$.
\end{lemma}

Under Assumption \ref{assum:policy}, \ref{assum:func_approx} and \ref{assum:semi_pos}, we utilize the result (Lemma 4.7) in \cite{ding2022} as the Lemma \ref{lemma:gradient_domination}.
\begin{lemma}
\label{lemma:gradient_domination}
(Relaxed weak gradient domination) Under Assumptions \ref{assum:policy}, \ref{assum:func_approx} and \ref{assum:semi_pos}, it holds that
\begin{equation}
\label{eq:gradient_domination}
    \|\nabla J(\theta)\| + \epsilon_{g} ~\geq~ \sqrt{2\mu}(J^{\star} - J(\theta)),
\end{equation}
where $\epsilon_{g} = \frac{\mu_{F} \sqrt{\epsilon_{{\rm bias}}}}{M_{g} (1-\gamma)}$ and $\mu = \frac{\mu_{F}^{2}}{2M_{g}^{2}}$.
\end{lemma}

Based on Lemma \ref{lemma:exp_return}, we derive our milestone (new), the ascent lemma with delayed updates, as follows:

\begin{lemma}
\label{lemma:ascent_lemma}
(Ascent Lemma with Delay) Under Assumptions \ref{assum:policy}, it holds that
\begin{equation}
\label{eq:ascent_lemma}
\begin{aligned}
    -J(\theta_{k+1}) ~\leq~ -J(\theta_{k}) - \frac{1}{3} \eta_{k} \| \nabla J(\theta_{k}) \| 
    + \frac{8}{3} \eta_{k} \|e_{k}\| + \frac{L_{g}}{2} \eta_{k}^{2},
\end{aligned}
\end{equation}
where $e_{k} \coloneqq d_{k - \delta_{k}} - \nabla J(\theta_{k})$.
\end{lemma}

This ascent lemma is specific to the asynchronous setting, which constructs the lower bound for the global increment, $J(\theta_{k+1})-J(\theta_{k})$, through the normalized policy gradients and our updating rules with delay $\delta_{k}$.

\begin{proof}
With the smoothness of the expected return $J(\theta)$ and the updating rule, we have
\begin{equation}
\label{eq:smooth_update}
    \begin{aligned}
        -J(\theta_{k+1}) ~&\leq~ -J(\theta_{k}) - \langle \nabla J(\theta_{k}), \theta_{k+1} - \theta_{k} \rangle + \frac{L_{g}}{2} \| \theta_{k+1} - \theta_{k} \|^{2} \\
        ~&=~ -J(\theta_{k}) - \eta_{k} \frac{\langle \nabla J(\theta_{k}), d_{k - \delta_{k}} \rangle}{\| d_{k - \delta_{k}} \|} + \frac{L_{g}}{2} \eta_{k}^{2}.
    \end{aligned}
\end{equation}

If $\|e_{k}\| \leq \frac{1}{2}\| \nabla J(\theta_{k}) \|$, we have
\begin{equation}
    \begin{aligned}
    -\frac{\langle \nabla J(\theta_{k}), d_{k - \delta_{k}} \rangle}{\| d_{k - \delta_{k}} \|} ~&=~ -\frac{\| \nabla J(\theta_{k}) \|^{2} + \langle \nabla J(\theta_{k}), e_{k} \rangle}{\| d_{k - \delta_{k}} \|} \\
    ~&\leq~ \frac{-\| \nabla J(\theta_{k}) \|^{2} + \| \nabla J(\theta_{k})\| \| e_{k} \|}{\| d_{k - \delta_{k}} \|} \\
    ~&\leq~ \frac{-\| \nabla J(\theta_{k}) \|^{2} + \frac{1}{2} \| \nabla J(\theta_{k})\|^{2}}{\| \nabla J(\theta_{k}) \| + \| e_{k} \|} \\
    ~&\leq~ -\frac{1}{3} \| \nabla J(\theta_{k}) \|.
    \end{aligned}
\end{equation}

If $\|e_{k}\| \geq \frac{1}{2}\| \nabla J(\theta_{k}) \|$, we have
\begin{equation}
    \begin{aligned}
    -\frac{\langle \nabla J(\theta_{k}), d_{k - \delta_{k}} \rangle}{\| d_{k - \delta_{k}} \|} ~&\leq~ \| \nabla J(\theta_{k}) \| \\
    ~&=~ -\frac{1}{3} \| \nabla J(\theta_{k}) \| + \frac{4}{3} \| \nabla J(\theta_{k}) \| \\
    ~&\leq~ -\frac{1}{3} \| \nabla J(\theta_{k}) \| + \frac{8}{3} \|e_{k}\|.
    \end{aligned}
\end{equation}

Combining these two conditions, and plugging the result into \eqref{eq:smooth_update}, the lemma can be proved as follows
\begin{equation}
    \begin{aligned}
     -J(\theta_{k+1}) ~&\leq~ -J(\theta_{k}) - \eta_{k} \frac{\langle \nabla J(\theta_{k}), d_{k - \delta_{k}} \rangle}{\| d_{k - \delta_{k}} \|} + \frac{L_{g}}{2} \eta_{k}^{2} \\
     ~&\leq~ -J(\theta_{k}) - \frac{1}{3} \eta_{k} \| \nabla J(\theta_{k}) \| + \frac{8}{3} \eta_{k} \|e_{k}\| + \frac{L_{g}}{2} \eta_{k}^{2}.
    \end{aligned}
\end{equation}
\end{proof}

Next, we construct the relationship between the average concurrency $\bar{\omega}$ and the average delay $\bar{\delta}$ in Lemma \ref{lemma:delay_concurrency}. This lemma gives the boundary (our result) of delays as a corollary of the result in \cite{NEURIPS2022_6db3ea52}.

\begin{lemma}
\label{lemma:delay_concurrency}
The average delay $\bar{\delta}$ depends on the average concurrency $\bar{\omega}$, and they can be upper bounded as
\end{lemma}
\begin{equation}
\label{eq:delay_concurrency}
\begin{aligned}
\bar{\delta} ~=~ \frac{K+1}{K-1+\|\mathcal{C}_{K}\|} \bar{\omega} ~\leq~ \bar{\omega} ~\le~ N.
\end{aligned}
\end{equation}

\begin{proof}
Recall that $\{ \delta_{k}^{i} \}_{i \in\mathcal{C}_{k} \setminus \{j_{k}\}}$ is the set of delays at the $k$-th global steps. After one global step, the number of cumulative delays over all agents increases by the current concurrency. Thus, we have the following connection
\begin{equation}
\begin{aligned}
    \sum_{i=0}^{k}\delta_{i} + \sum_{i \in\mathcal{C}_{k+1} \setminus \{j_{k+1} \}}\delta^{i}_{k+1} ~&=~ \sum_{i=0}^{k-1}\delta_{i} + \sum_{i \in\mathcal{C}_{k} \setminus \{ j_{k} \}}\delta^{i}_{k} + \omega_{k+1}.
\end{aligned}
\end{equation}

We note that there is no delay at the initial step ($0$-th iteration) of the algorithm. Therefore, we have $\delta^{i}_{0}=0$ for all agents. Unrolling the above expression, at the $K$-th step, we have
\begin{equation}
\begin{aligned}
\sum_{i=0}^{K-1}\delta_{i} + \sum_{i \in\mathcal{C}_{K} \setminus \{ j_{K} \}}\delta^{i}_{K} ~&=~ \sum_{i=0}^{K} \omega_{k+1} ~=~ (K+1)\bar{\omega}.
\end{aligned}
\end{equation}
According to \eqref{eq:avg_delay} and $\|\mathcal{C}_{K}\| \geq 2$, we achieve
\begin{equation}
\begin{aligned}
\bar{\delta} ~=~ \frac{K+1}{K-1+\|\mathcal{C}_{K}\|}\bar{\omega} ~\leq~ \bar{\omega} ~\leq~ N.
\end{aligned}
\end{equation}
\end{proof}

In practice, we need to use all the resources to speed up the training process. Thus, all agents engage in training, and $\bar{\omega} = \omega_{\max} = N$. Since the maximum concurrency is equal to the number of agents $N$, we have the upper boundary of the average delay as $\bar{\delta} \le \omega_{\max} \le N$. 

Notably, we do not make any assumption on the largest delay $\delta_{\max}$, while we achieve the upper boundary of the average delay $\bar{\delta}$.

\subsection{Proof of Theorem \ref{theorem:afedpg_rate} (Global Convergence Rate)}
\label{app:proof_theorem_1}

Under Assumption \ref{assum:policy} and Assumption \ref{assum:func_approx}, we derive the global convergence rate of the proposed AFedPG.

First, we denote the difference between the policy gradient estimation $g(\widetilde{\tau}_{k}, \widetilde{\theta}_{k})$ and the true policy gradient as
\begin{equation}
\begin{aligned}
\label{eq:xi_k}
\xi_{k} \coloneqq g(\widetilde{\tau}_{k}, \widetilde{\theta}_{k}) - \nabla J(\widetilde{\theta}_{k}),
\end{aligned}
\end{equation}
and the expectation of the norm is bounded by $\sigma_{g}$.

Second, according to the updating rules in Algorithm \ref{algo_server} and Algorithm \ref{algo_agent} , we expand the error term in Lemma \ref{lemma:ascent_lemma} as follows
\begin{equation}
\label{eq:e_k}
    \begin{aligned}
     e_{k} ~=&~ (1 - \alpha_{k - \delta_{k}}) d_{k-1 - \delta_{k-1}} - \nabla J(\theta_{k}) + \alpha_{k - \delta_{k}} g(\widetilde{\tau}_{k - \delta_{k}}, \widetilde{\theta}_{k - \delta_{k}}) \\
     ~=&~ (1 - \alpha_{k-\delta_{k}}) \big(d_{k-1-\delta_{k-1}} - \nabla J(\theta_{k-1}) \big) + (1 - \alpha_{k-\delta_{k}}) \big(\nabla J(\theta_{k-1}) - \nabla J(\theta_{k}) \big) \\
     &+ \alpha_{k-\delta_{k}} \big( g(\widetilde{\tau}_{k-\delta_{k}}, \widetilde{\theta}_{k-\delta_{k}}) - \nabla J(\theta_{k}) \big) \\
     ~=&~ \alpha_{k-\delta_{k}} \xi_{k-\delta_{k}} + (1 - \alpha_{k-\delta_{k}}) e_{k-1} + (1 - \alpha_{k-\delta_{k}}) \big(\nabla J(\theta_{k-1}) - \nabla J(\theta_{k}) \big) \\
     &+ \alpha_{k-\delta_{k}} \big( \nabla J(\widetilde{\theta}_{k-\delta_{k}}) - \nabla J(\theta_{k}) \big) \\
     ~=&~ \alpha_{k-\delta_{k}} \xi_{k-\delta_{k}} + (1 - \alpha_{k-\delta_{k}}) e_{k-1} + (1 - \alpha_{k-\delta_{k}}) \big(\nabla J(\theta_{k-1}) - \nabla J(\theta_{k}) \big) \\
     &+ \alpha_{k-\delta_{k}} \big( \nabla J(\widetilde{\theta}_{k-\delta_{k}}) - \nabla J(\widetilde{\theta}_{k}) \big) + \alpha_{k-\delta_{k}} \big( \nabla J(\widetilde{\theta}_{k}) - \nabla J(\theta_{k}) \big) \\
     ~=&~ \alpha_{k-\delta_{k}} \xi_{k-\delta_{k}} + (1 - \alpha_{k-\delta_{k}}) e_{k-1} + \alpha_{k-\delta_{k}} \big( \nabla J(\widetilde{\theta}_{k-\delta_{k}}) - \nabla J(\widetilde{\theta}_{k}) \big) \\
     &+ (1 - \alpha_{k-\delta_{k}}) \big(\nabla J(\theta_{k-1}) - \nabla J(\theta_{k}) + \nabla^{2} J(\theta_{k}) (\theta_{k-1} - \theta_{k}) \big) \\
     &+ \alpha_{k-\delta_{k}} \big( \nabla J(\widetilde{\theta}_{k}) - \nabla J(\theta_{k}) + \nabla^{2} J(\theta_{k}) (\theta_{k-1} - \theta_{k}) \big) \\
     &\textcolor{blue}{- (1 - \alpha_{k-\delta_{k}}) \nabla^{2} J(\theta_{k}) (\theta_{k-1} - \theta_{k}) - \alpha_{k-\delta_{k}} \nabla^{2} J(\theta_{k}) (\widetilde{\theta}_{k} - \theta_{k})} \\
     ~\stackrel{\eqref{eq:delay_adaptive}}{=}&~ \alpha_{k-\delta_{k}} \xi_{k-\delta_{k}} + (1 - \alpha_{k-\delta_{k}}) e_{k-1} + \alpha_{k-\delta_{k}} \big( \nabla J(\widetilde{\theta}_{k-\delta_{k}}) - \nabla J(\widetilde{\theta}_{k}) \big) \\
     &+ (1 - \alpha_{k-\delta_{k}}) \big(\nabla J(\theta_{k-1}) - \nabla J(\theta_{k}) + \nabla^{2} J(\theta_{k}) (\theta_{k-1} - \theta_{k}) \big) \\
     &+ \alpha_{k-\delta_{k}} \big( \nabla J(\widetilde{\theta}_{k}) - \nabla J(\theta_{k}) + \nabla^{2} J(\theta_{k}) (\theta_{k-1} - \theta_{k}) \big) .
    \end{aligned}
\end{equation}
Thus, the error term $e_{k}$ can be written in a recursive way (contains $e_{k-1}$) with serval terms that contain policy gradients. We aim to derive the upper boundary for each term at the next step.

\textbf{Remark}: The Hessian correction terms (in blue) are equal to $0$ according to our updating rules (delay-adaptive lookahead update) in Step 8 of Algorithm \ref{algo_server}. We design this technique to cancel out the second-order terms, and thus achieve the desired convergence rate. Without our technique, it would be hard to bound the above errors.

Next, we denote each term in \eqref{eq:e_k} separately as follows
\begin{equation}
\begin{aligned}
A_{k} ~\coloneqq&~ \nabla J(\theta_{k-1}) - \nabla J(\theta_{k}) + \nabla^{2} J(\theta_{k}) (\theta_{k-1} - \theta_{k}), \\
B_{k} ~\coloneqq&~ \nabla J(\widetilde{\theta}_{k}) - \nabla J(\theta_{k}) + \nabla^{2} J(\theta_{k}) (\theta_{k-1} - \theta_{k}), \\
C_{k} ~\coloneqq&~ \nabla J(\widetilde{\theta}_{k-\delta_{k}}) - \nabla J(\widetilde{\theta}_{k}).
\end{aligned}
\end{equation}

Now, we start to bound each term. With the smoothness of the expected discounted return function in Lemma \ref{lemma:exp_return} and the non-increasing learning rates, we have
\begin{equation}
\label{eq:ab_bound}
\begin{aligned}
\|A_{k}\| ~&\leq~ L_{h} \|\theta_{k-1} - \theta_{k}\|^{2} = L_{h} \eta_{k-1}^{2}, \\
\|B_{k}\| ~&\leq~ L_{h} \|\widetilde{\theta}_{k} - \theta_{k}\|^{2} = L_{h} \frac{(1 - \alpha_{k-\delta_{k}})^{2}}{\alpha_{k-\delta_{k}}^{2}} \eta_{k-1}^{2}, \\
\|C_{k}\| ~&\leq~ L_{g} \|\widetilde{\theta}_{k-\delta_{k}} - \widetilde{\theta}_{k}\| \\
~&=~ L_{g} \left\| \sum_{i=k-\delta_{k}}^{k-1} \widetilde{\theta}_{i+1} - \widetilde{\theta}_{i} \right\| \\
~&\leq~ L_{g} \sum_{i=k-\delta_{k}}^{k-1} \| \widetilde{\theta}_{i+1} - \widetilde{\theta}_{i} \| \\
~&=~ L_{g} \sum_{i=k-\delta_{k}}^{k-1} \left\| \frac{1}{\alpha_{i+1-\delta_{i+1}}}({\theta}_{i+1} - {\theta}_{i}) + \frac{1 - \alpha_{i-\delta_{i}}}{\alpha_{i-\delta_{i}}} ({\theta}_{i} - {\theta}_{i-1}) \right\| \\
~&\leq~ L_{g} \sum_{i=k-\delta_{k}}^{k-1} \left\| \frac{1}{\alpha_{i+1-\delta_{i+1}}}({\theta}_{i+1} - {\theta}_{i}) \right\| + \left\| \frac{1 - \alpha_{i-\delta_{i}}}{\alpha_{i-\delta_{i}}} ({\theta}_{i} - {\theta}_{i-1}) \right\| \\
~&=~ L_{g} \sum_{i=k-\delta_{k}}^{k-1} (\frac{1}{\alpha_{i+1-\delta_{i+1}}} \eta_{i} + \frac{1 - \alpha_{i-\delta_{i}}}{\alpha_{i-\delta_{i}}} \eta_{i-1}) \\
~&\leq~ L_{g} \sum_{i=k-\delta_{k}}^{k-1} (\frac{2}{\alpha_{k-1}} \eta_{k-\delta_{k}}) \\
~&=~ \delta_{k} L_{g} \frac{2\eta_{k-\delta_{k}}}{\alpha_{k-1}}, 
\end{aligned}
\end{equation}
where the equalities are simple plugins according to the updating rules in Algorithm \ref{algo_server}. The last step in \eqref{eq:ab_bound} happens, because there is no index $i$ inside the summation operation, and it sums a constant for $\delta_{k}$ times.

We denote $\beta_{k} \coloneqq \prod_{i=k+1}^{K} (1 - \alpha_{i-\delta_{i}})$ with $\beta_{K} = 1$. Next, unrolling the recursion \eqref{eq:e_k}, we have the error at the $K$-th step as follows
\begin{equation}
    \begin{aligned}
     e_{K} ~&=~ \beta_{0} e_{0} + \sum_{k=1}^{K} \alpha_{k-\delta_{k}} \beta_{k} \xi_{k-\delta_{k}} + \sum_{k=1}^{K} (1 - \alpha_{k-\delta_{k}}) \beta_{k} A_{k} + \sum_{k=1}^{K} \alpha_{k-\delta_{k}} \beta_{k} (B_{k} + C_{k}).
    \end{aligned}
\end{equation}

Choose learning rates $\alpha_{k} = (\frac{1}{k+1})^{\frac{4}{5}}$ and $\eta_{k} = \eta_{0}\frac{1}{k+1}$, where $\eta_{0}$ is a constant and we will show the value later. Using Jensen's inequality and technical lemmas, we achieve the following bound
\begin{equation}
\label{eq:error_k}
    \begin{aligned}
     \mathbb{E}[\|e_{K}\|] ~\leq&~ \beta_{0} \mathbb{E}[\|e_{0}\|] + \left( \mathbb{E}\left[\left\| \sum_{k=1}^{K} \alpha_{k-\delta_{k}} \beta_{k} \xi_{k-\delta_{k}} \right\|^{2}\right] \right)^{\frac{1}{2}} + \sum_{k=1}^{K} (1 - \alpha_{k-\delta_{k}}) \beta_{k} \mathbb{E}[\|A_{k}\|] \\
     &+ \sum_{k=1}^{K} \alpha_{k-\delta_{k}} \beta_{k} (\mathbb{E}[\|B_{k}\|] + \mathbb{E}[\|C_{k}\|]) \\
     ~\leq&~ \beta_{0} \mathbb{E}[\|e_{0}\|] + \left( \sum_{k=1}^{K} \alpha_{k-\delta_{k}}^{2} \beta_{k}^{2} \mathbb{E}\left[\left\| \xi_{k-\delta_{k}} \right\|^{2}\right] \right)^{\frac{1}{2}} + \sum_{k=1}^{K} (1 - \alpha_{k-\delta_{k}}) \beta_{k} \mathbb{E}[\|A_{k}\|] \\
     &+ \sum_{k=1}^{K} \alpha_{k-\delta_{k}} \beta_{k} (\mathbb{E}[\|B_{k}\|] + \mathbb{E}[\|C_{k}\|]) \\
     ~\stackrel{\eqref{eq:ab_bound}}{\le}&~ \beta_{0} \sigma_{g} + \left( \sum_{k=1}^{K} \alpha_{k-\delta_{k}}^{2} \beta_{k}^{2} \mathbb{E}\left[\left\| \xi_{k-\delta_{k}} \right\|^{2}\right] \right)^{\frac{1}{2}} + L_{h} \sum_{k=1}^{K} (1 - \alpha_{k-\delta_{k}}) \beta_{k} \eta_{k-1}^{2} \\
     &+ L_{h} \sum_{k=1}^{K} \beta_{k} (1 - \alpha_{k-\delta_{k}})^{2} \frac{\eta_{k-1}^{2}}{\alpha_{k-\delta_{k}}} + 2L_{g} \sum_{k=1}^{K} \alpha_{k-\delta_{k}} \beta_{k} \delta_{k} \frac{\eta_{k-\delta_{k}}^{2}}{\alpha_{k-1}} \\
     ~\leq&~ \beta_{0} \sigma_{g} + \left( \sum_{k=1}^{K} \alpha_{k-\delta_{k}}^{2} \beta_{k}^{2} \mathbb{E}\left[\left\| \xi_{k-\delta_{k}} \right\|^{2}\right] \right)^{\frac{1}{2}} + L_{h} \sum_{k=1}^{K} \beta_{k} \frac{\eta_{k-1}^{2}}{\alpha_{k-\delta_{k}}} \\
     &+ 2L_{g} \sum_{k=1}^{K} \alpha_{k-\delta_{k}} \beta_{k} \delta_{k} \frac{\eta_{k-\delta_{k}}^{2}}{\alpha_{k-1}} \\
     ~\leq&~ \beta_{0} \sigma_{g} + \left( \sum_{k=1}^{K} \alpha_{k-\delta_{k}}^{2} \beta_{k}^{2} \right)^{\frac{1}{2}} \sigma_{g} + L_{h} \sum_{k=1}^{K} \beta_{k} \frac{\eta_{k-1}^{2}}{\alpha_{k-\delta_{k}}} + 2L_{g} \sum_{k=1}^{K} \alpha_{k-\delta_{k}} \beta_{k} \delta_{k} \frac{\eta_{k-\delta_{k}}^{2}}{\alpha_{k-1}} \\
     ~\le&~ \beta_{0} \sigma_{g} + \left( \sum_{k=1}^{K} \alpha_{k-\delta_{k}}^{2} \beta_{k}^{2} \right)^{\frac{1}{2}} \sigma_{g} + \frac{9L_{h}}{4} \sum_{k=1}^{K} \beta_{k} \frac{\eta_{k}^{2}}{\alpha_{k}} + 2L_{g} \sum_{k=1}^{K} \alpha_{k-\delta_{k}} \beta_{k} \delta_{k} \frac{\eta_{k-\delta_{k}}^{2}}{\alpha_{k-1}} \\
     ~&\stackrel{\eqref{eq:bound_1}, \eqref{eq:bound_2}, \eqref{eq:bound_3}, \eqref{eq:bound_4}}{\le}~ \alpha_{K} \sigma_{g} + c_{1} \sqrt{\alpha_{K}} \sigma_{g} + \frac{9}{4} c_{2} L_{h}\frac{\eta_{K}^{2}}{\alpha_{K}^{2}} + c_{3} L_{g} \frac{\eta_{K}^{2}}{\alpha_{K}^{1.75}} \bar{\delta},
    \end{aligned}
\end{equation}
where $c_{1} \coloneqq 2\sqrt{c(\frac{4}{5},\frac{4}{5})}$, $c_{2} \coloneqq c(\frac{4}{5},\frac{6}{5})$, $c_{3} \coloneqq 8 \sqrt{c(\frac{4}{5},\frac{16}{5})}$ are constants, and the values of $c(\cdot, \cdot)$ are defined in Lemma \ref{lemma:lr_seq_bound}.

Notably, as we state in the last paragraph in Section \ref{sec:Convergence_Analysis}, $t_i$ is pre-determined by the computation resource at agent $i$ and is fixed. Thus, the delay $\delta_{k}$ is pre-determined by the system. It is not a random variable, but unknown until given the exact system setting. This allows us to derive the first inequality in \eqref{eq:error_k}. This pre-determined property is also suitable for all the derivations below.

We explain the boundary derivation details here. We first derive the first term of the boundary in \eqref{eq:error_k} as follows
\begin{equation}
\begin{aligned}
\label{eq:bound_1}
\beta_{0} ~=&~ \prod_{i=1}^{K} (1 - \alpha_{i-\delta_{i}}) \\
~\leq&~ \prod_{i=1}^{K} (1 - \alpha_{i}) \\
~=&~ \prod_{i=0}^{K-1} (1 - \alpha_{i+1}) \\
~\stackrel{\eqref{eq:lr_bound}}{\leq}&~ \frac{\alpha_{K}}{\alpha_{0}} \\
~=&~ \alpha_{K}.
\end{aligned}
\end{equation}

In the training process, at step $k$, when an agent sends the update to the server, as long as the agent communicates with the server during the last half training process, the delay $\delta_{k} \le \frac{k}{2}$. This becomes almost certain when $k$ is large and $k$ is usually much larger than the upper bound of the currency $N$. Thus, we have
\begin{equation}
\begin{aligned}
\label{eq:alpha_delay}
\alpha_{k-\delta_{k}} ~=&~ (\frac{1}{k+1 - \delta_{k}})^{\frac{4}{5}} \\
~=&~ (\frac{1}{k+1})^{\frac{4}{5}}(\frac{k+1}{k+1 - \delta_{k}})^{\frac{4}{5}} \\
~\leq&~ (\frac{1}{k+1})^{\frac{4}{5}} (\frac{k+1}{k+1 - \frac{k}{2}})^{\frac{4}{5}} \\
~\leq&~ 2 (\frac{1}{k+1})^{\frac{4}{5}} \\
~=&~ 2\alpha_{k}.
\end{aligned}
\end{equation}

We then derive the second term of the boundary in \eqref{eq:error_k} as follows
\begin{equation}
\begin{aligned}
\label{eq:bound_2}
\sum_{k=1}^{K} \alpha_{k-\delta_{k}}^{2} \beta_{k}^{2} ~\stackrel{\eqref{eq:alpha_delay}}{\leq}&~ 4\sum_{k=1}^{K} \alpha_{k}^{2} \beta_{k}^{2} \\
~=&~ 4\sum_{k=1}^{K} \alpha_{k}^{2} \prod_{i=k+1}^{K} (1 - \alpha_{i-\delta_{i}}) \prod_{i=k+1}^{K} (1 - \alpha_{i-\delta_{i}}) \\
~\leq&~ 4\sum_{k=1}^{K} \alpha_{k}^{2} \prod_{i=k+1}^{K} (1 - \alpha_{i}) \prod_{i=k+1}^{K} (1 - \alpha_{i}) \\
~\leq&~ 4\sum_{k=1}^{K} \alpha_{k}^{2} \prod_{i=k+1}^{K-1} (1 - \alpha_{i}) \prod_{i=k}^{K-1} (1 - \alpha_{i+1}) \\
~\stackrel{\eqref{eq:lr_bound}}{\leq}&~ 4\sum_{k=1}^{K} \alpha_{k}^{2} \prod_{i=k+1}^{K-1} (1 - \alpha_{i}) \frac{\alpha_{K}}{\alpha_{k}} \\
~=&~ 4\alpha_{K} \sum_{k=1}^{K} \alpha_{k} \prod_{i=k+1}^{K-1} (1 - \alpha_{i}) \\
~\leq&~ 4\alpha_{K} \sum_{k=0}^{K-1} \alpha_{k} \prod_{i=k+1}^{K-1} (1 - \alpha_{i}) \\
~\stackrel{\eqref{eq:lr_seq_bound}}{\leq}&~ 4\alpha_{K} c(\frac{4}{5}, \frac{4}{5}) \frac{\alpha_{K}}{\alpha_{K}} \\
~=&~ 4\alpha_{K} c(\frac{4}{5}, \frac{4}{5}).
\end{aligned}
\end{equation}

Next, we derive the third term of the boundary in \eqref{eq:error_k} as follows
\begin{equation}
\begin{aligned}
\label{eq:bound_3}
\sum_{k=1}^{K} \frac{\eta_{k}^{2}}{\alpha_{k}} \beta_{k} ~=&~ \eta_{0}^{2} \sum_{k=1}^{K} (\frac{1}{k+1})^{\frac{6}{5}} \beta_{k} \\
~=&~ \eta_{0}^{2} \sum_{k=1}^{K} (\frac{1}{k+1})^{\frac{6}{5}} \prod_{i=k+1}^{K} (1 - \alpha_{i-\delta_{i}}) \\
~\leq&~ \eta_{0}^{2} \sum_{k=1}^{K} (\frac{1}{k+1})^{\frac{6}{5}} \prod_{i=k+1}^{K} (1 - \alpha_{i}) \\
~\leq&~ \eta_{0}^{2} \sum_{k=1}^{K} (\frac{1}{k+1})^{\frac{6}{5}} \prod_{i=k+1}^{K-1} (1 - \alpha_{i}) \\
~\leq&~ \eta_{0}^{2} \sum_{k=0}^{K-1} (\frac{1}{k+1})^{\frac{6}{5}} \prod_{i=k+1}^{K-1} (1 - \alpha_{i}) \\
~\stackrel{\eqref{eq:lr_seq_bound}}{\leq}&~ \eta_{0}^{2} c(\frac{4}{5}, \frac{6}{5}) (\frac{1}{K+1})^{\frac{2}{5}} \\
~=&~ c(\frac{4}{5}, \frac{6}{5}) \frac{\eta_{K}^{2}}{\alpha_{K}^{2}}.
\end{aligned}
\end{equation}

At last, we derive the fourth term of the boundary in \eqref{eq:error_k}. We use Cauchy--Schwarz inequality and the fact that the second norm of a vector is always equal or smaller than the first norm.
\begin{equation}
\begin{aligned}
\label{eq:bound_4}
\sum_{k=1}^{K} \alpha_{k-\delta_{k}} \beta_{k} \delta_{k} \frac{\eta_{k-\delta_{k}}^{2}}{\alpha_{k-1}} ~\leq&~ \sum_{k=1}^{K} \eta_{k-\delta_{k}}^{2} \beta_{k} \delta_{k} \\
~=&~ \sum_{k=1}^{K} \eta_{k}^{2} \Big( \frac{k+1}{k+1 - \frac{k}{2}} \Big)^{2} \beta_{k} \delta_{k} \\
~\leq&~ 4\eta_{0}^{2}\sum_{k=1}^{K} \Big( \frac{1}{k+1} \Big)^{2} \beta_{k} \delta_{k} \\
~\leq&~ 4\eta_{0}^{2} \left( \sum_{k=1}^{K} \Big( \frac{1}{k+1} \Big)^{4} \beta_{k}^{2} \cdot \sum_{k=1}^{K} \delta_{k}^{2} \right)^{\frac{1}{2}} \\
~\leq&~ 4\eta_{0}^{2} \left( \sum_{k=1}^{K} \Big( \frac{1}{k+1} \Big)^{4} \beta_{k}^{2} \right)^{\frac{1}{2}} \sum_{k=1}^{K} \delta_{k}  \\
~\leq&~ 4\eta_{0}^{2} K \bar{\delta} \left( \sum_{k=1}^{K} \Big( \frac{1}{k+1} \Big)^{4} \beta_{k}^{2} \right)^{\frac{1}{2}} \\
~\leq&~ 4\eta_{0}^{2} K \bar{\delta} \left( \sum_{k=1}^{K} \Big( \frac{1}{k+1} \Big)^{4} \beta_{k} \prod_{i=k+1}^{K} (1 - \alpha_{i-\delta_{i}}) \right)^{\frac{1}{2}} \nonumber
\end{aligned}
\end{equation}
\begin{equation}
\begin{aligned}
~\leq&~ 4\eta_{0}^{2} K \bar{\delta} \left( \sum_{k=1}^{K} \Big( \frac{1}{k+1} \Big)^{4} \beta_{k} \frac{\alpha_{K}}{\alpha_{k}} \right)^{\frac{1}{2}} \\
~=&~ 4\eta_{0}^{2} K \bar{\delta} \left( \alpha_{K} \sum_{k=1}^{K} \Big( \frac{1}{k+1} \Big)^{\frac{16}{5}} \beta_{k} \right)^{\frac{1}{2}} \\
~=&~ 4\eta_{0}^{2} K \bar{\delta} \left( \alpha_{K} \sum_{k=1}^{K} \Big( \frac{1}{k+1} \Big)^{\frac{16}{5}} \prod_{i=k+1}^{K} (1 - \alpha_{i-\delta_{i}}) \right)^{\frac{1}{2}} \\
~\leq&~ 4\eta_{0}^{2} K \bar{\delta} \left( \alpha_{K} \sum_{k=1}^{K} \Big( \frac{1}{k+1} \Big)^{\frac{16}{5}} \prod_{i=k+1}^{K} (1 - \alpha_{i}) \right)^{\frac{1}{2}} \\
~\leq&~ 4\eta_{0}^{2} K \bar{\delta} \left( \alpha_{K} \sum_{k=1}^{K} \Big( \frac{1}{k+1} \Big)^{\frac{16}{5}} \prod_{i=k+1}^{K-1} (1 - \alpha_{i}) \right)^{\frac{1}{2}} \\
~\stackrel{\eqref{eq:lr_seq_bound}}{\leq}&~ 4\eta_{0}^{2} K \bar{\delta} \left( \alpha_{K} c(\frac{4}{5}, \frac{16}{5}) (\frac{1}{K+1})^{\frac{12}{5}} \right)^{\frac{1}{2}} \\
~\leq&~ 4\eta_{0}^{2} K \bar{\delta} \sqrt{c(\frac{4}{5}, \frac{16}{5})} (\frac{1}{K+1})^{\frac{8}{5}} \\
~\leq&~ 4\eta_{0}^{2} \bar{\delta} \sqrt{c(\frac{4}{5}, \frac{16}{5})} (\frac{1}{K+1})^{\frac{3}{5}} \\
~=&~ 4 \bar{\delta} \sqrt{c(\frac{4}{5}, \frac{16}{5})} \frac{\eta_{K}^{2}}{\alpha_{K}^{1.75}}.
\end{aligned}
\end{equation}

Now, we derive the final convergence rate. After plugging the result into Lemma \ref{lemma:ascent_lemma} and using Lemma \ref{lemma:gradient_domination} (Ascent Lemma with Delay), we achieve the following inequality
\begin{equation}
    \begin{aligned}
    J^{\star} - \mathbb{E}[J(\theta_{k+1})] ~\leq&~ (1 - \frac{\sqrt{2\mu}\eta_{k}}{3}) \big( J^{\star} - \mathbb{E}[J(\theta_{k})] \big) + \frac{\eta_{k}}{3} \epsilon_{g} + \frac{8}{3} \eta_{k} \mathbb{E}[\|e_{k}\|] + \frac{L_{g}}{2} \eta_{k}^{2} \\
    ~\stackrel{\eqref{eq:error_k}}{\le}&~ (1 - \frac{\sqrt{2\mu}\eta_{k}}{3}) \big( J^{\star} - \mathbb{E}[J(\theta_{k})] \big) + \frac{\eta_{k}}{3} \epsilon_{g} + \frac{L_{g}}{2} \eta_{k}^{2} \\
    &+ \frac{8}{3} \eta_{k} \big(\alpha_{k} \sigma_{g} + c_{1} \sqrt{\alpha_{k}} \sigma_{g} + \frac{9}{4} c_{2} L_{h}\frac{\eta_{k}^{2}}{\alpha_{k}^{2}} + c_{3}L_{g} \frac{\eta_{k}^{2}}{\alpha_{k}^{1.75}} \bar{\delta} \big).
    \end{aligned}
\end{equation}
Unrolling this recursion, we have
\begin{equation}
\label{eq:converge_rate}
    \begin{aligned}
    J^{\star} - \mathbb{E}[J(\theta_{K})] ~\leq&~ \frac{\eta_{0}}{3} \epsilon_{g} + \frac{J^{\star} - \mathbb{E}[J(\theta_{0})]}{(K+1)^{2}} + c_{3} \frac{8\eta_{0}^{3}}{3} \frac{L_{g}}{(K+1)^{\frac{3}{5}}} \bar{\delta} + \frac{\eta_{0}^{2}}{2} \frac{L_{g}}{K+1} \\
    &+ \frac{8\eta_{0}}{3} \frac{\sigma_{g}}{(K+1)^{\frac{4}{5}}} + c_{1} \frac{8\eta_{0}}{3} \frac{\sigma_{g}}{(K+1)^{\frac{2}{5}}} + 6 c_{2} \eta_{0}^{3} \frac{L_{h}}{(K+1)^{\frac{2}{5}}}.
    \end{aligned}
\end{equation}

Note that, introduced in Lemma \ref{lemma:exp_return}, the discount factor $\gamma$ is contained in $L_{g} = \mathcal{O}\big((1-\gamma)^{-2}\big)$ and $L_{h} = \mathcal{O}\big((1-\gamma)^{-3}\big)$. Comparing with the definition of $\epsilon_{g}$ in Lemma \ref{lemma:gradient_domination}, we choose $\eta_{0} = \frac{3\mu_{F}}{M_{g}}$ for the criteria. Thus, to satisfy the global convergence criterion $J^{\star} - \mathbb{E}[J(\theta_{K})] \leq \epsilon + \frac{\sqrt{\epsilon_{{\rm bias}}}}{1 - \gamma}$, we have the iteration complexity $K = \mathcal{O}({\epsilon}^{-2.5})$. In our algorithms, the sample complexity is equal to the iteration complexity, which is also $\mathcal{O}({\epsilon}^{-2.5})$.

\subsection{Proof of Theorem \ref{theorem:afedpg_rate_FOSP} (FOSP Convergence Rate)}
\label{app:proof_theorem_2}

Under Assumption \ref{assum:policy} and Assumption \ref{assum:semi_pos}, we derive the first-order stationary convergence rate of AFedPG. Notably, the FOSP convergence does not require Assumption \ref{assum:func_approx}, which makes assumptions on the neural network approximation error.

First, we denote the average of gradient expectations as 
\begin{equation}
\begin{aligned}
\mathbb{E} \| \nabla J(\bar{\theta}_{K}) \| \coloneqq \frac{\sum_{k=1}^{K} \eta_{k} \mathbb{E} \| \nabla J(\theta_{k}) \|}{\sum_{k=1}^{K} \eta_{k}}.
\end{aligned}
\end{equation}

Next, rearranging the terms in Lemma \ref{lemma:ascent_lemma}
(Ascent Lemma with Delay) and summing up the inequality, we achieve the inequality below
\begin{equation}
\begin{aligned}
\mathbb{E} \| \nabla J(\bar{\theta}_{K}) \| ~\leq~& \frac{3}{\sum_{k=1}^{K} \eta_{k}} \Big( J^{\star} - \mathbb{E}[J(\theta_{0})] \Big)
    + \frac{8}{\sum_{k=1}^{K} \eta_{k}} \sum_{k=1}^{K} \eta_{k} \mathbb{E}[\|e_{k}\|] \\
    &+ \frac{3 L_{g}}{2 \sum_{k=1}^{K} \eta_{k}} \sum_{k=1}^{K} \eta_{k}^{2}.
\end{aligned}
\end{equation}

Choose learning rates $\alpha_{k} = (\frac{1}{k+1})^{\frac{4}{7}}$ and $\eta_{k} = \eta_{0}(\frac{1}{k+1})^{\frac{5}{7}}$. Plug in the result into \eqref{eq:error_k} that we have shown in Appendix \ref{app:proof_theorem_1}, we have
\begin{equation}
\begin{aligned}
\mathbb{E} \| \nabla J(\bar{\theta}_{K}) \| ~\leq~& \frac{3( J^{\star} - \mathbb{E}[J(\theta_{0})] )}{(K+1)^{\frac{2}{7}}} + 16 c_{3} {\eta_{0}^{3}} \frac{L_{g}}{(K+1)^{\frac{3}{7}}} \bar{\delta} + \frac{3\eta_{0} L_{g}}{(K+1)^{\frac{5}{7}}} \\
&+ {16\eta_{0}} \frac{\sigma_{g}}{(K+1)^{\frac{4}{7}}} + 16 c_{1} {\eta_{0}} \frac{\sigma_{g}}{(K+1)^{\frac{2}{7}}} + 36 c_{2} \eta_{0}^{3} \frac{L_{h}}{(K+1)^{\frac{2}{7}}}.
\end{aligned}
\end{equation}

Thus, to satisfy the FOSP convergence criterion $\mathbb{E} \| \nabla J(\bar{\theta}_{K}) \| \le \epsilon$, we have the iteration complexity $K = \mathcal{O}({\epsilon}^{-3.5})$. In our algorithms, the sample complexity is equal to the iteration complexity, which is also $\mathcal{O}({\epsilon}^{-3.5})$.

Notably, this result does not rely on Lemma \ref{lemma:gradient_domination} and Assumption \ref{assum:policy}. The norm of the average gradient is approaching an arbitrarily small value during the training process, regardless of the function approximation error $\epsilon_{\rm bias}$.

\newpage
\section{Further Discussions}
\label{app:discussion}

\subsection{Comparison with Previous Works}

\textbf{Comparison with} \cite{shen2023towards}. We first acknowledge the theoretical contribution of this pioneer work. However, there are several limitations and differences.

1. (Different RL Algorithm Class) Instead of PG, \cite{shen2023towards} is an actor-critic (AC) method with extra value networks, which requires much more computation and memory cost compared to the pure policy gradient (PG) method. Thus, the fine-tuning of Gemini \cite{Gemma2024} and GPT-4 \cite{openai2023gpt4} uses PG methods instead of AC methods.

2. (General Function Parameterization) \cite{shen2023towards} only has Linear Parameterization (Deep RL is not included.) for the global convergence analysis, which has limited practical meaning. With a General Function Parametrization, \textit{e.g.}, neural networks (Deep RL), there is no such a result. We consider a general and practical setting with a General Function Parameterization in our work.

3. (Convergence Performance) Even in the single-agent setting (without federated agents), the SOTA result of the AC method is ${\mathcal{O}}({\epsilon}^{-3})$ \cite{Gaur2024icml} and the previous approach is ${\mathcal{O}}({\epsilon}^{-6})$ \cite{fu2021singletimescale}, which is still worse than our ${\mathcal{O}}({\epsilon}^{-2.5})$. In the federated setting, there is no result that achieves ${\mathcal{O}}({\epsilon}^{-3})$ for AC methods. Moreover, with a general function parameterization, we compare the performances of their A3C in Figure \ref{fig:fed_time}, which is much worse.

4. (Assumptions) \cite{shen2023towards} relies on a strong and unpractical assumption, their Assumption 2. It assumes that the largest delay is bounded by a constant $K_{0}$. However, in practice, the slowest agent may not communicate with the server, and thus, has an infinite delay. In our analysis, we do not require any boundary for the largest delay, because we only contain the average delay in the convergence rate, and the average delay is naturally bounded by the number of agents in Lemma \ref{lemma:delay_concurrency} (our corollary).

\subsection{Difference between FL, FedRL, and AFedRL}

Unlike supervised FL where local datasets are fixed or pre-specified, in RL, agents collect new samples in each iteration based on the current local policies. The new data are collected with dynamic dependencies, which do not appear in all prior FL and A-FL works.

In synchronous FedRL, each agent collects samples according to the same global policy $\pi_{\theta}$. However, in AFedRL, even if all agents have an identical environment, each agent collects samples according to different policies $\tau_{k} \sim p(\cdot \| \pi_{\theta_{k}})$, because of the delay. This dynamic nature makes both the problem itself and the theoretical analysis challenging. We propose a new delay-adaptive model aggregation strategy to tackle these unique challenges of FedRL.

Moreover, data collected by each agent are naturally (non-manually controlled) heterogeneous. Despite having identical environments, agents collect data according to their own (different) policies, a fundamental difficulty that our AFedPG paper solves, as verified theoretically and empirically.

  \include{ap-appendix_mappo}
  \include{ap-appendix_contextual_integrity}
  

  \pdfbookmark{COLOPHON}{colophon}
  \ifthen{\equal{true}{\ZZshowcolophon}}
    {\include{ap-colophon}}

\end{document}